 \documentclass[final,5p,times,twocolumn,authoryear]{elsarticle}

\usepackage{amssymb}
\usepackage{graphicx} 
\usepackage{svg}
\usepackage{amsmath}
\usepackage{amsfonts}
\usepackage{hyperref}
\usepackage{multirow}
\usepackage{subcaption}
\usepackage[font=small,labelfont=bf]{caption}
\usepackage{stfloats}
\usepackage{tabularx}
\usepackage{booktabs}

\usepackage{adjustbox} 
\usepackage{hhline}
\usepackage{cancel}
\usepackage{placeins}
\usepackage{pdflscape}
\usepackage{array}
\usepackage{xcolor}
\usepackage{ulem}

\newif\ifshowrev
\showrevfalse

\ifshowrev
  \newcommand{\rev}[1]{\textcolor{blue}{#1}}
\else
  \newcommand{\rev}[1]{#1}
  \definecolor{blue}{rgb}{0,0,0}
\fi

\makeatletter
\def\ps@pprintTitle{%
   \let\@oddhead\@empty
   \let\@evenhead\@empty
   \let\@oddfoot\@empty
   \let\@evenfoot\@empty
}
\makeatother
\begin{document}
\begin{frontmatter}

\title{Super-Resolved Canopy Height Mapping from Sentinel-2 Time Series Using \rev{Airborne} LiDAR HD Reference Data across Metropolitan France}

\author[label1]{Ekaterina Kalinicheva}
\author[label2]{Florian Helen}
\author[label2]{Stéphane Mermoz}
\author[label1]{Florian Mouret}
\author[label1]{Milena Planells}

\affiliation[label1]{
    organization={CESBIO},
    addressline={18 avenue Édouard Belin},
    city={Toulouse},
    postcode={31401},
    country={France}
}

\affiliation[label2]{organization={GlobEO},
    addressline={Avenue Saint-Éxupery},
    city={Toulouse},
    postcode={31400},
    country={France}}


\begin{abstract}
Fine-scale forest monitoring is essential for understanding canopy structure and its dynamics, which are key indicators of carbon stocks, biodiversity, and forest health.
Deep learning is particularly effective for this task, as it integrates spectral, temporal, and spatial signals that jointly reflect the canopy structure. To address this need, we introduce THREASURE-Net, a novel end-to-end framework for Tree Height REgression And SUper-REsolution. The model is trained on Sentinel-2 time series using reference height metrics derived from LiDAR~HD data at multiple spatial resolutions over Metropolitan France to produce annual height maps.
We evaluate three model variants, producing tree-height predictions at 2.5 m, 5 m, and 10 m resolution. 
THREASURE-Net does not rely on any pretrained model, nor on reference very high resolution optical imagery to train its super-resolution module; instead, it learns solely from LiDAR-derived height information.
Our approach outperforms existing state-of-the-art methods based on Sentinel data and is competitive with methods based on very high resolution imagery. It can be deployed to generate high-precision annual canopy-height maps, achieving mean absolute errors of 2.63 m, 2.70 m, and 2.89 m at 2.5 m, 5 m, and 10 m resolution, respectively.
These results highlight the potential of THREASURE-Net for scalable and cost-effective structural monitoring of temperate forests using only freely available satellite data. The source code for THREASURE-Net is available at: \url{https://github.com/Global-Earth-Observation/threasure-net
}.
\end{abstract}



\begin{keyword}
forest canopy height regression \sep super-resolution \sep Sentinel-2 time series \sep LiDAR HD \sep temporal transformer



\end{keyword}

\end{frontmatter}

\section{Introduction}

Forest monitoring has always been essential for the sustainable functioning of ecosystems, managing the balance between harvesting for industry, carbon storage, and species diversity ~\citep{FAO2025}. Canopy height mapping is a key component of forest monitoring, as it provides insights into stand structure, biomass, regeneration dynamics, and disturbance impacts~\citep{lang2023high}. Since many forest processes -- such as selective logging, canopy degradation, regeneration, or subtle structural changes -- occur at fine spatial scales, methods and spatial resolutions have evolved substantially over time to improve the accuracy and spatial resolution of canopy height predictions. 

To monitor canopy evolution across large territories over time, researchers increasingly rely on satellite imagery as a practical source of information~\citep{rs_forest}. Initially, a statistical approach was used to combine satellite imagery with \textit{in situ} measurements~\citep{simard2011mapping, avitabile2016integrated, bouvet2018above, mermoz2024submonthly}. The \textit{in situ} or field measurements can be national inventories carried out by the national forest agency at regular intervals (every 5 years in France) or isolated campaigns. Nevertheless, manual annotation of forest stands at large scales to obtain reference data is impractical and costly. As a result, it has become common practice to use LiDAR (Light Detection and Ranging) data -- whether airborne or satellite-based -- to derive reference data. Among the most widely used sources are GEDI~\citep{DUBAYAH2020100002} a spaceborne LiDAR mission providing discrete vertical profiles at 25-m resolution, and various airborne LiDAR campaigns.  Today, airborne LiDAR data remains the most accurate and comprehensive source of information for forest structure monitoring and analysis \citep{COOPS2021112477}. It provides detailed 3D point clouds whose spatial resolution -- directly linked to point density -- allows the precise characterization of forest vertical and horizontal structure. Many European countries have launched national LiDAR monitoring campaigns and provide Airborne laser scanning (ALS) data free of charge~\citep{Kakoulaki2021LiDAR}. In France, the LiDAR~HD\footnote{\url{https://geoservices.ign.fr/lidarhd}} national ALS campaign aims to cover the entire metropolitan territory with a minimum density of 10 pulses per m², which constitutes a key reference dataset for forest structural studies. However, despite its extensive coverage of French territory, each area is typically observed only once, which prevents any temporal analysis using LiDAR~HD data alone. 

To overcome the lack of temporal information and spatial discontinuity, these LiDAR datasets are used in conjunction with satellite image series to produce temporally continuous maps. The methods and spatial resolutions involved have evolved substantially over time. While early approaches relied on coarse to medium resolution imagery combined with traditional machine learning models such as Random Forests~\citep{morin2023estimation,AHMED201589, WANG201624}, more recent works have shown that deep learning methods using high resolution (HR) open access satellite data from missions such as Sentinel-1 and Sentinel-2, at 10-20~m of resolution, can predict canopy structure with greater accuracy~\citep{lang2023high, essd-15-4927-2023, BECKER2023269}. These methods can produce annual maps, thereby increasing the frequency of forest monitoring. To overcome the limits of the 10~m spatial resolution, recent studies use very high-resolution (VHR) commercial imagery such as PlanetScope~\citep{doi10.1126_sciadv.adh4097}, Maxar~\citep{TOLAN2024113888}, or SPOT-6/7~\citep{11095116}. However, acquiring VHR imagery remains expensive; therefore, training models at national scales is often impractical and costly. 

In France, several models have been proposed to map tree height across the territory using LiDAR as reference data. For instance, FORMS-T~\citep{SCHWARTZ2025114959} uses yearly median composites of Sentinel-1 and Sentinel-2 imagery and GEDI LiDAR data as reference to predict canopy structure at 10~m resolution over the French metropolitan territory. In contrast, Open-Canopy~\citep{11095116} relies on a single VHR SPOT-6/7 image combined with LiDAR HD to map fine-scale canopy height at 1.5~m, also over the French metropolitan territory.

Among other recent models that could be used in France, the Global Canopy Height (GCH) Model~\citep{lang2023high} can be named. It provides global 10~m canopy height estimates using GEDI LiDAR as a reference and a single-date Sentinel-2 image, combined with coordinate encoding to identify geographical zones as an input.

In this context, two central research questions remain open: how can prediction accuracy be further improved, and how can very high-resolution canopy height estimates be produced while keeping data acquisition costs low. To address these questions, we focus in this work on improving tree height prediction while simultaneously increasing the output resolution using only freely available remote sensing data and super-resolution deep learning methods. 

Super-resolution algorithms aim to transfer high resolution (HR) features to low resolution (LR) imagery ~\citep{rs14215423}. While most studies focus solely on the super-resolution (SR)
task~\citep{rs_esrgan, Liu2024, 11010858, Muller2020}, some studies combine SR with a real-life prediction task (such as building height estimation or object segmentation) to leverage the additional spatial information from higher-resolution images~\citep{Lei2019SimultaneousSA, rs13224547, CAO2024114241}. In these approaches, the SR component is typically supervised using either VHR images or artificially downsampled data to create LR–HR training pairs, while a separate dataset is used to supervise the downstream task.

These methods focus on joint Single Image Super Resolution (SISR) analysis since their prediction tasks do not benefit from multi-temporal data. Today, several SISR models~\citep{11010858, cresson2022sr4rs} can be used to enhance Sentinel-2 satellite image time series (SITS). Therefore, the resulting VHR time series can then be directly used as input for prediction algorithms. However, processing super-resolved image series requires significantly more computational resources compared to using the original low-resolution data. Moreover, errors in image calibration and co-registration within a time series can introduce significant misalignments with the reference data, leading to substantial inconsistencies in the super-resolved input time series, potentially reducing the reliability of the prediction results. 

In contrast, Multi-Image Super-Resolution (MISR) techniques aim to jointly exploit multiple low resolution observations of the same area to reconstruct a higher resolution image. While these methods have demonstrated significant gains in domains such as video restoration or change detection~\citep{9044873}, they are rarely explored in the context of Earth Observation. Furthermore, they are better suited to capture phenological variation over time, which is crucial for accurate vegetation analysis \citep{DBLP:journals/corr/abs-1811-10166}.

While we have not found any works performing MISR simultaneously with a downstream prediction task, some works focusing solely on MISR can be cited. These approaches can be generally categorized into three categories: (i) temporal-transformer models that learn shared information across time before increasing the image resolution ~\citep{BreizhSR}, (ii) diffusion-based models that reconstruct high-resolution images by iteratively denoising samples while being guided by the low-resolution inputs~\citep{DBLP:journals/corr/abs-2406-10225}, and (iii) recursive fusion models~\citep{DBLP:journals/corr/abs-2002-06460}, that progressively combine multiple low-resolution observations to generate a higher-resolution output.


Motivated by these findings, we propose a novel end-to-end framework that simultaneously performs super-resolution and forest canopy height estimation using Sentinel-2 image time series. Our model, named THREASURE-Net (Tree Height REgression And SUper REsolution), allows us to generate tree height maps at 2.5, 5 and 10~m spatial resolutions. We apply our method to metropolitan France since forest monitoring in France still relies largely on field measurements and photointerpretation of high-resolution imagery, compared with Scandinavian countries, where LiDAR and satellite imagery are already integrated into national forest inventory systems. The objective of this study is therefore to provide new tools that can complement existing practices in France. To assess their relevance, we compare our approach with several state-of-the-art algorithms commonly used for forest variable estimation from satellite data.

Our main contributions can be summarized as follows:
\begin{itemize}
    \item We propose a fully end-to-end model that does not rely on any pretrained super-resolution networks or any VHR reference optical image data.
    \item THREASURE-Net, trained solely on Sentinel-2 data, achieves an accuracy comparable to models trained on VHR satellite data, while also providing coherent fine-scale spatial details that are not visible in predictions at the native resolution.
    \item During training, model predictions are conditioned by LiDAR's acquisition date, explicitly accounting for the seasonal variability of reference data. 
    \item The model is able to handle irregular Sentinel-2 time series without explicitly requiring cloud-free observations.
    \item Our approach simultaneously predicts both canopy height and a tree/non-tree classification mask, whereas existing methods restrict height prediction to pre-defined forest areas using an external mask.
\end{itemize} 

 \begin{figure}[t]
    \centering
    \includegraphics[width=\linewidth]{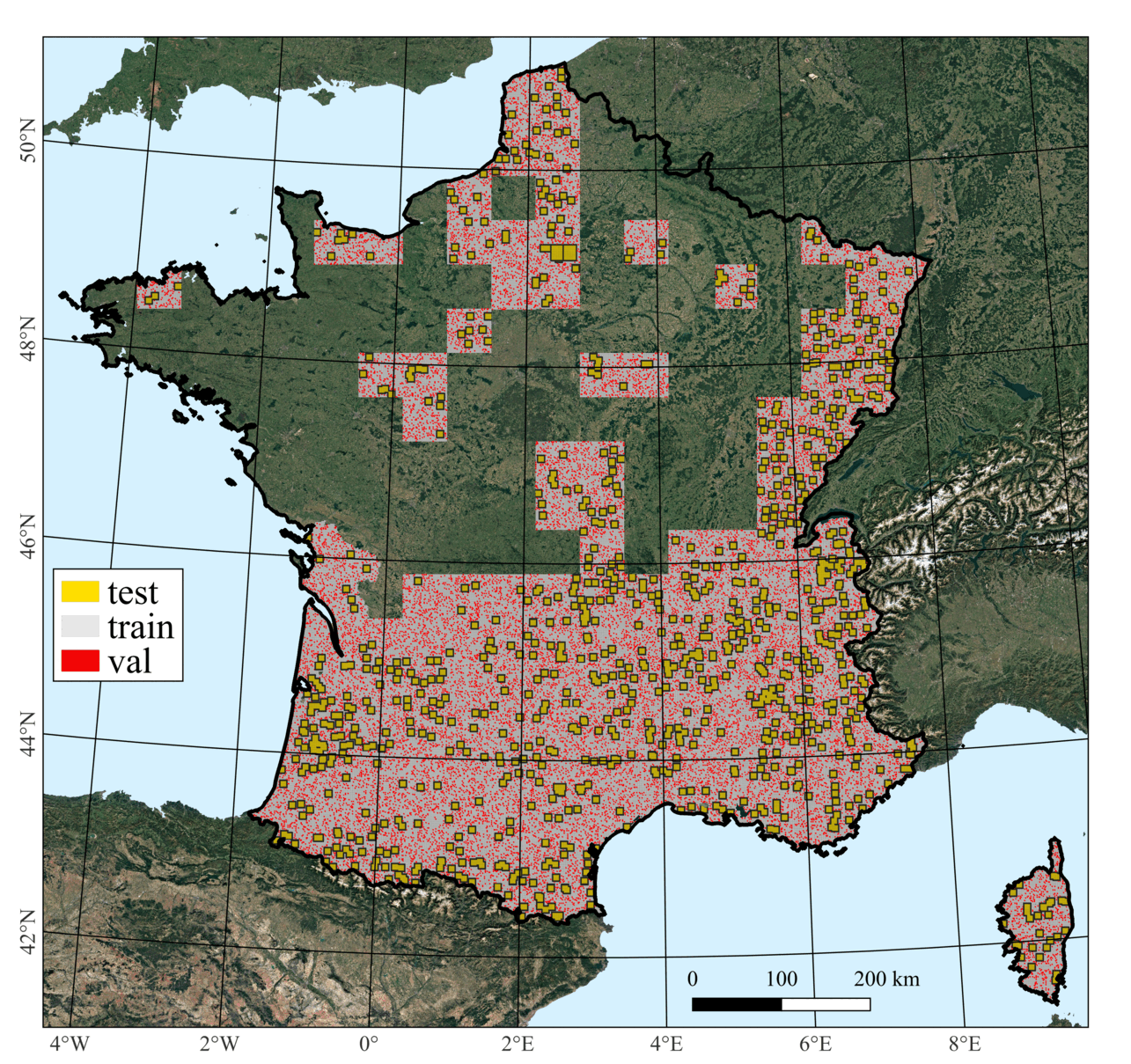}
    \caption{Distribution of train, test and validation patches. Buffers of 1 km are set around test patches to ensure reliable evaluation of model results.}
    \label{fig:dataset}
\end{figure}

\section{Data}
\label{sec:data}

\subsection{Study area}

Our area of study is mainland France (including Corsica), which spans 543,940 km². French forests cover 17.5 million hectares, accounting for nearly one-third of the territory. This makes them the fourth largest in Europe by surface area, after Sweden, Finland, and Spain. Three-quarters of these forests are privately owned. Public forests are mainly concentrated in the southern and eastern parts of the country. 
France's geographical diversity is notable, featuring five distinct climates (oceanic, transitional oceanic, semi-continental, Mediterranean, and mountain) and $190$ tree species, as well as various silvicultural techniques. This diversity makes it a prime example of temperate European forests. The National Forest Inventory has divided the metropolitan territory into $91$ sections called \textit{sylvoécorégions} (SER) -- forest ecoregions -- taking into account the bio-geographical factors that determine forest production and the distribution of major forest habitats (e.g., altitude, climate, and soil characteristics). \rev{See Figure~\ref{fig:SER} and~\ref{apx:SER}}.

\newcolumntype{C}[1]{>{\centering\arraybackslash}m{#1}}

{
\renewcommand{\arraystretch}{1.5}
\begin{table*}[!htbp]
\centering
\caption{Overview of datasets used in this study.}
\label{tab:data_all}

\begin{tabular}{C{3cm} C{1.7cm} C{1.7cm} m{5.2cm} C{2.8cm} C{1.7cm}}

\toprule
Data & Time span & Resolution & \multicolumn{1}{c}{Description} & Source & Usage \\
\midrule

Sentinel-2 SITS & 2018--2024 & 10 m & 
Multispectral satellite imagery providing dense time series used to capture spectral and temporal information related to canopy structure. &
\href{https://sentinels.copernicus.eu/web/sentinel/sentinel-data-access}{Copernicus} &
Training / Evaluation \\

LiDAR HD & 2021--2024 & 10 pts/m$^2$ &
Airborne LiDAR dataset providing high-resolution canopy height metrics used as reference data for model training and evaluation. &
\href{https://geoservices.ign.fr/lidarhd}{IGN} &
Training / Evaluation \\

Land Parcel Identification System (RPG) & 2024 & -- &
Agricultural parcel dataset used to identify and mask non-forest areas within the study region. &
\href{https://www.data.gouv.fr/}{RPG -- Data.gouv} &
Masking \\

\midrule

FORMS-T \cite{SCHWARTZ2025114959} & 2018--2023 & 10 m &
Canopy height prediction maps from FORMS-T algorithm for different years, used to compare with our predictions for independent ALS and disturbance detection. &
\href{https://browser.datastore-mtd.theia.data-terra.org/collections/forms-t}{FORMS-T} &
Comparison \\

Open-Canopy \cite{11095116} & 2018--2023 & 1.5 m &
Canopy height prediction maps from Open-Canopy algorithm for different years, used to compare with our predictions for independent ALS and disturbance detection. &
\href{https://browser.datastore-mtd.theia.data-terra.org/collections/FORMSpoT}{FORMSpoT} &
Comparison \\

Global Canopy Height (GCH) \cite{lang2023high} & 2020 & 10 m &
Canopy height prediction maps from GCH model used as reference for 2020 to compare with our predictions. &
\href{https://langnico.github.io/globalcanopyheight/}{GCH Model} &
Comparison \\

\midrule

Carcans-Hourtins & 10/2020 & 31 pts/m$^2$ &
ALS-derived tree height map. Coastal plain, pine forest with a few broadleaf trees in the understory. &
ONF &
Evaluation \\

Déodatie & 04/2018 & 13 pts/m$^2$ &
ALS-derived tree height map. Hilly terrain, mixed forest: beech, oak, hornbeam, spruce, and pine. &
ONF &
Evaluation \\

Lajoux-Lafresse & 06/2019 & 31 pts/m$^2$ &
ALS-derived tree height map. Mountainous terrain, spruce, pine, and beech. &
ONF &
Evaluation \\

Mouterhouse & 02/2019 & 17 pts/m$^2$ &
ALS-derived tree height map. Flat plain with small hills, beech, oak, hornbeam, and pine. &
ONF &
Evaluation \\

\bottomrule
\end{tabular}

\end{table*}
}

\subsection{Training dataset}

To train THREASURE-Net, a dataset composed of \mbox{Sentinel-2} SITS and corresponding LiDAR~HD patches was generated.

The French territory is covered by $80$ Sentinel-2 tiles following the Military Grid Reference System (MGRS). To build Sentinel-2 SITS, we choose images acquired from May to October, on the same year as the reference LiDAR~HD patch. We filter out images with cloud coverage and missing data superior to 50\% and 10\%, respectively.

Sentinel-2 image time series correspond to the Level 2A product, \rev{processed from L1C by MAJA~\footnote{\url{https://www.cesbio.cnrs.fr/maja/}. MAJA stands for MACCS-ATCOR Joint Algorithm, where MACCS refers to Multi-sensor Atmospheric Correction and Cloud Screening, and ATCOR to Atmospheric/Topographic Correction.} algorithm developed by Theia, the French land data center for Earth observation.} We select 10 spectral bands (B02, B03, B04, B05, B06, B07, B08, B8A, B11, B12), Cloud Mask, as well as acquisition angles: Sun azimuth and zenith angles - $\theta_{S_a}$ and $\theta_{S_z}$ respectively, satellite mean view azimuth $\theta_{V_a}$ and zenith angles $\theta_{V_z}$. All bands and angles are resampled at 10~m. The acquisition angles are encoded using trigonometric transformations to respect angular periodicity: 
\[ \bigl[
\cos(\theta_{S_z}), 
\hspace{0.5em} \cos(\theta_{S_a}),
\hspace{0.5em} \sin(\theta_{S_a}),
\hspace{0.5em} \cos(\theta_{V_z}),
\hspace{0.5em} \cos(\theta_{V_a}),
\hspace{0.5em} \sin(\theta_{V_a})
 \bigr].
\]

Due to the coarse resolution of the available angle data (5~km pixel size), the angle values are extracted as vectors corresponding to the center of each patch.

The reference data are derived from the Institut Geographique National (IGN)'s LiDAR~HD open dataset, with patches acquired between 2021 and 2024. This consists of a national airborne LiDAR campaign that aims to cover the entire French territory with a minimum density of 10 pts/m². Patches have already been acquired over 2021-2025 years, and the campaign is ongoing. The 3D point clouds are transformed into height rasters in a fully-automatic way: point cloud values are spatially projected onto a regular pixel grid with a chosen spatial resolution (10~m, 5~m or 2.5~m), from which the $95^{th}$ percentile of elevation values is computed for the ensemble of points falling within each pixel.
The LiDAR~HD product is provided with point-wise classification into multiple land cover categories. Tree height values are only derived for pixels corresponding to vegetation higher than 1.5~m, while the remaining pixels are set to zero. However, tall crop vegetation, such as corn, may be incorrectly classified as forest vegetation. This introduces potential bias into model training, as crop vegetation does not exhibit the same structural or phenological characteristics as trees. To solve this issue, we apply a crop mask derived from Land Parcel Identification System (\textit{Le Registre Parcellaire Graphique} (RPG)), the French national graphic crop register. 
All vegetation within crop parcels with height below 5~m is considered non-permanent, and its height values are excluded from the annotations.

To ensure a spatially robust and statistically representative dataset, we implemented a stratified sampling strategy on the LiDAR~HD tiles, taking into account forest coverage and canopy height variance in the SER regions. First, a subset of 15\% of LiDAR~HD tiles was assigned to the test dataset, providing an independent evaluation dataset. To reduce spatial autocorrelation and prevent information leakage, all LiDAR tiles adjacent to the test tiles, including diagonal neighbors, were designated as buffer zones and excluded from the training and validation sets. The remaining tiles were randomly split into training (90\%) and validation (10\%) subsets. Within this stratified dataset, samples were randomly  selected to obtain $179945$ patches for training, $19133$ for validation, and $35639$ for testing. This approach ensures the spatial independence of the test dataset while preserving regional representativeness in model development.

\subsection{Ancillary data}

To compare our canopy height estimates with independent reference ALS data, we use four height maps with 1 m resolution, derived from various high-density LiDAR campaigns conducted by the Office National des Forêts (ONF).


To compare our algorithm with the concurrent approaches, we equally use canopy height prediction maps 
provided by the authors of FORMS-T~\citep{SCHWARTZ2025114959} at 10\,m resolution, Open-Canopy~\citep{11095116} at 1.5\,m resolution for different years, and the Global Canopy Height Model~\citep{lang2023high} at 10\,m resolution for 2020 year.


A summary of all datasets used in this study is provided in Table~\ref{tab:data_all}.

\label{sec:method}

\section{Methodology}

\subsection{Model}

\begin{figure*}[t]
    \centering
    \includegraphics[width=\linewidth]{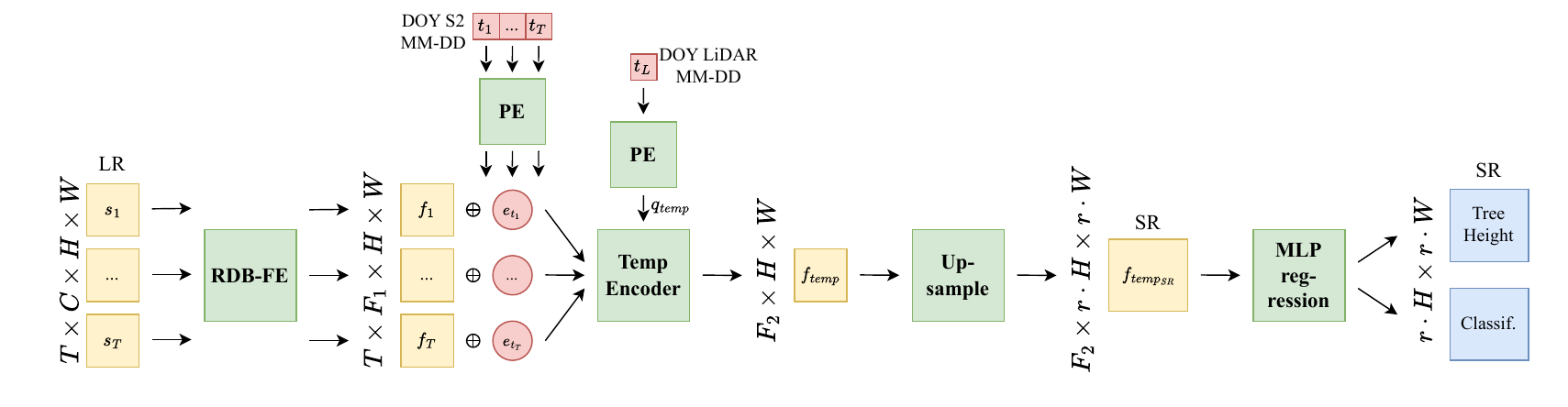}
    \caption{THREASURE-Net model architecture.}
    \label{fig:model}
\end{figure*}

In this work, we propose an end-to-end model, named THREASURE-Net, that performs super-resolution and regression tasks simultaneously. Additionally, our model produces a tree/non-tree classification mask, based on tree height regression features.
Given a Sentinel-2 SITS $s = [s_{t_1}, s_{t_2}, \dots, s_T] \in \mathbb{R}^{T \times C \times H \times W}$ acquired at dates $t = [t_1, t_2, \dots, t_T] \in \mathbb{R}^T$, the model aims to predict the tree height \(95^{\text{th}}\) percentile corresponding to the reference date $t_L$, producing an output of shape $f \cdot H \times f\cdot W $, where $f$ denotes the SR factor.
Here, $T$, $C$, $H$, and $W$ represent the number of acquisition dates, spectral channels, image height, and image width in pixels, respectively.
Note that for the sake of simplicity, we omit batch dimension in the model description.

THREASURE-Net model is composed of several blocks, as shown in Figure~\ref{fig:model}:

\textbf{Spatio-spectral (SS) feature extraction module.}
First, we extracted features independently for each SITS image with Residual Dense Blocks-based Feature Extraction module (here referred to as RDB-FE). This produces a new feature time series $f = [f_{1_{t_1}}, f_{1_{t_1}}, \dots, f_{1_T}] \in \mathbb{R}^{T \times F_1 \times H \times W}$, where the new feature dimension $F_1$ is a hyperparameter.

The RDB-FE module is extracted from Residual Dense network (RDN) \citep{DBLP:journals/corr/abs-1802-08797}, which is mainly used for natural image SISR task and originally consists of four parts:
shallow feature extraction net (SFENet), residual dense blocks (RDBs), dense feature fusion (DFF) and finally the up-sampling net (UPNet). Given the multi-temporal nature of our model, at this stage, we only deploy the feature extraction modules (SFENet, RDBs, and DFF). The up-sampling module is applied later to the temporally encoded features, rather than to each image individually.

While alternative architectures such as ResNet \citep{DBLP:journals/corr/HeZRS15} or Residual-in-Residual Dense Blocks (RRDB) \citep{DBLP:journals/corr/abs-1809-00219} could also be employed for feature extraction, we selected  RDBs as a balanced choice.
RDBs offer improved feature reuse compared to ResNet, thanks to dense connections that help capture fine-grained spatial details in Sentinel-2 imagery. At the same time, RDBs are significantly lighter than RRDBs, making them particularly well suited for our multi-temporal super-resolution task.


\textbf{Temporal feature extraction block.}
To extract a unique set of temporal features for each Sentinel-2 series, we exploit the Lightweight Temporal Self-Attention mechanism proposed by \cite{DBLP:journals/corr/abs-2007-00586}. 
The module(here referred to as Temporal Encoder) captures dependencies across the multi-temporal Sentinel-2 time series and allows the network to dynamically weight the contribution of each timestamp, by combining keys $K$, queries $Q$ and values $V$ to produce a self-attention mask.. 
The simplified version of this algorithm can be summarized as follows:

Let $\mathbf{f} = [\mathbf{f}_1, \mathbf{f}_2, \dots, \mathbf{f}_T] \in \mathbb{R}^{T \times F_1}$
be the sequence of encoded Sentinel-2 observations at times $\{t_1, \dots, t_T\}$. Each timestamp is associated with a temporal positional encoding 
$\mathbf{p}_i \in \mathbb{R}^{d}$ using a sinusoidal function from the Positional Encoding (PE) module:

\begin{equation}
\label{eq:positional_encoding}
\mathbf{p}_i^{(2k)} = \sin \!\left( \frac{t_i}{\tau^{2k/d}} \right), 
\quad 
\mathbf{p}_i^{(2k+1)} = \cos \!\left( \frac{t_i}{\tau^{2k/d}} \right),
\end{equation}

with $\tau$ a scaling constant and $k$ indexing the embedding dimensions. The positional encoding is added to the input features:

\begin{equation}
\tilde{\mathbf{f}}_i = \mathbf{f}_i + \mathbf{p}_i.
\end{equation}

The attention mechanism computes keys $\mathbf{K}$ and values $\mathbf{V}$ as linear projections of $\tilde{\mathbf{f}}$, while $\mathbf{Q}$ is a learnable parameter. The resulting aggregated temporal representation has a final size $T \times F_2 \times H \times W$,
where $F_2$ is the feature dimension after encoding and $H \times W$ are the spatial dimensions. For full details on the Lightweight Temporal Transformer, we refer the reader to \cite{DBLP:journals/corr/abs-2007-00586}.

However, since LiDAR HD reference data are acquired on different days of the year, the prediction quality may be affected by seasonal growth and on/off-leaf specificity of 3D LiDAR data. Therefore, we replace learnable queries from the original algorithm by temporal queries $\mathbf{q}_{temp}$ that condition the predicted temporal features by LiDAR acquisition day, by projecting them onto a temporally-aware latent space.

\begin{equation}
\mathbf{q}_{temp} = W_Q \, \mathbf{e}_{t_L},
\end{equation}

where $\mathbf{e}_{t_L}$ is encoded LiDAR dates, and $W_Q \in \mathbb{R}^{d \times d_a}$ are learnable weights.

The LiDAR date $t_L$ is encoded using the same sinusoidal positional encoding (eq.~\ref{eq:positional_encoding}) as the Sentinel-2 acquisition dates $t$, ensuring that both modalities share a common temporal representation space.
Finally, $f_{temp}$ serves as the temporally fused feature used for downstream regression.

As described previously, we encode both the Sentinel-2 acquisition dates $t$ and the reference LiDAR acquisition date $t_L$.
The dates are expressed as day-of-year (DOY) values,
$t = [t_1, \dots, t_i, \dots, t_T] \in \mathbb{R}^T$ and $t_L \in \mathbb{R}$,
in ``DD–MM'' format, with the year omitted to ensure generalization across unseen years.
Before applying the sinusoidal positional encoding, the DOY values are normalized as follows:

\begin{equation}
\label{eq:doy_s2}
t_i = t_i - t_{\mathrm{ref,S2}}, 
\qquad 
\text{with } t_{\mathrm{ref,S2}} = \text{January 1st}.
\end{equation}

\begin{equation}
\label{eq:doy_lidar}
t_L = t_L - t_{\mathrm{ref,L}}, 
\qquad 
\text{with } t_{\mathrm{ref,L}} = \text{July 1st}.
\end{equation}

As shown in Equations~\ref{eq:doy_s2} and~\ref{eq:doy_lidar},
the resulting DOYs represent relative time offsets (in days) from the beginning and the middle of the year, respectively.
This preprocessing highlights the seasonal relationship between the LiDAR reference date and the Sentinel-2 time series,
allowing the model to better capture phenological phase alignment.
During inference, predictions are by default conditioned on July 1st, whereas during training the DOY representation preserves temporal offsets around this date, enabling the model to encode whether an observation occurs earlier or later in the seasonal cycle.

Moreover, we use time series padding along the temporal dimension within each batch, which allows us to process SITS of various length.

\textbf{Upsampling block.} The upsampling step is applied after the temporal encoding rather than before, in order to reduce memory consumption—since a single shared temporally encoded feature map is upsampled instead of multiple individual ones—and to take advantage of the common temporal representation rather than processing separate temporal features independently.

\begin{figure}[ht]
    \centering
    \includegraphics[width=\linewidth]{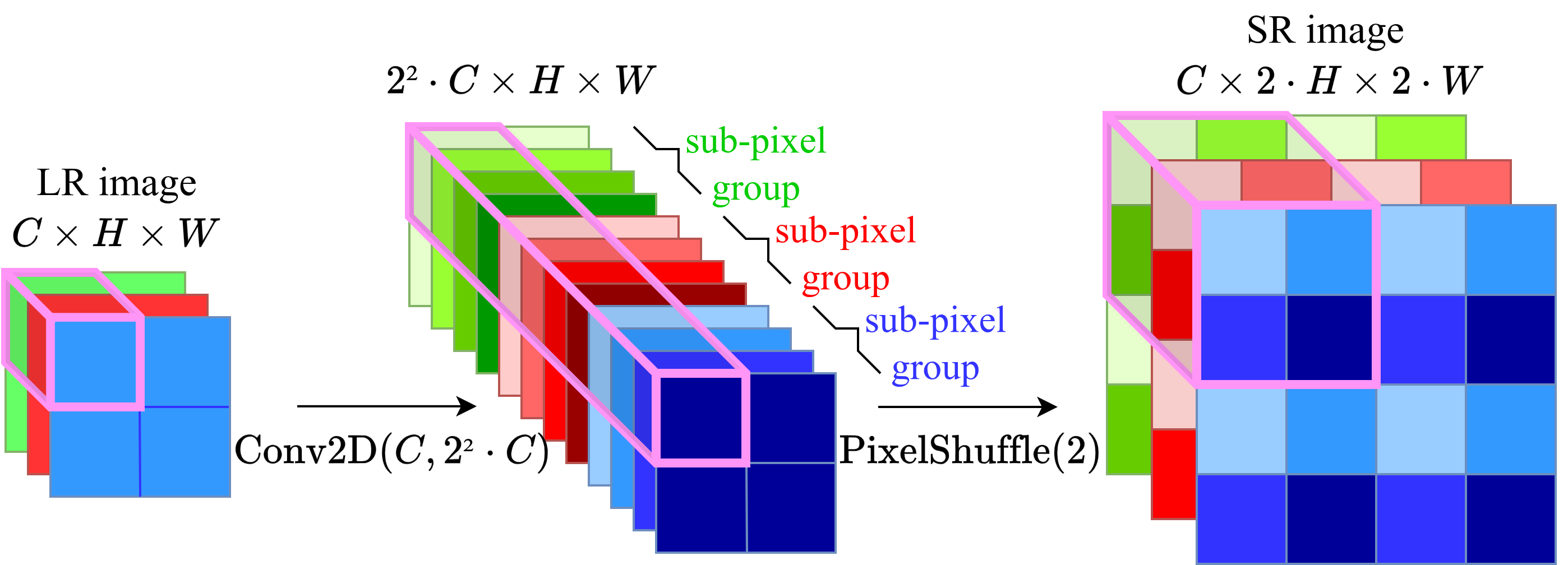}
    \caption{Sub-pixel convolution. Example of factor 2 upsampling. A LR image with $C$ feature channels and $H \times W$ spatial dimensions is passed through 2D convolution layer to obtain $2^2 \cdot C$ feature channels where each group of $2^2$ consecutive channels form a sub-pixel group that is reorganized by pixel shuffle operation to a SR image with $C$ channels and $2 \cdot H \times 2 \cdot W$ spatial dimensions.}
    \label{fig:pixelshuffle}
\end{figure}

The upsampling module employs a sub-pixel convolution (pixel shuffle) method~\citep{shuffle}, which reconstructs high-resolution outputs by rearranging low-resolution feature channels into spatial dimensions (see Figure~\ref{fig:pixelshuffle}). The SR is performed progressively by increasing the feature map resolution by a factor of 2 until the desired resolution is achieved.
Therefore, the super-resolution hyperparameter $r$ is constrained to be a power of $2$. In each SR block with $r=2$, the input feature map's feature dimension $F_2$ is increased by factor 2² with a 2D convolution operation, then the resulted feature map is rearranged by a pixel shuffle operation, producing a new feature map of size $F_2 \times 2H \times 2W$:
$Conv2D(F_2, 4\cdot F_2) \rightarrow PixelShufle(2)$.
The sequence of those blocks is followed by another 2D convolution operation with ReLU activation, without changing the feature map size.

However, the sub-pixel convolutions are known to produce checkerboard artifacts \citep{aitken2017checkerboardartifactfreesubpixel}, due to random weight initialization across sub-pixel groups. To mitigate this issue, the weights are initialized following the approach presented by \cite{aitken2017checkerboardartifactfreesubpixel}, where identical initialization is applied within each sub-pixel group to ensure smooth and artifact-free textures. A small amount of random noise is then added to break the symmetry and improve optimization dynamics.

\textbf{Tree height regression and classification.}
The final tree-height predictions and tree masks are generated by a Multi-Layer Perceptron (MLP) block (referred to as MLP regression). The MLP takes the super-resolved features as input and processes them through a set of shared layers. The final layer is task-specific, with two parallel heads. Each head performs a distinct task: the regression head predicts tree height, while the classification head outputs a tree/non-tree map at the chosen resolution. The classification head is optimized independently using detached features, preventing its gradients from affecting the shared representation.




\subsection{Loss functions}
To optimize the model, we employ the following loss functions: the L1 loss (Mean Absolute Error, MAE) for tree height prediction, Binary Cross-Entropy (BCE) loss for the tree/non-tree classification mask, and the Weighted Gradient Difference Loss (WGDL) to enhance the texture and edges of the super-resolved features. The total loss is defined as:

\begin{equation}
\mathcal{L}_{total}
 = \omega_{height} \cdot \mathcal{L}_{height}
 + \omega_{classif} \cdot \mathcal{L}_{classif}
 + \omega_{wgdl} \cdot \mathcal{L}_{wgdl}.
\end{equation}

where $\omega_{height}$, $\omega_{classif}$, and $\omega_{wgdl}$ are weighting coefficients controlling the relative contribution of each loss term.






\textbf{Tree height regression Loss.} By design choice, the tree classification mask extraction is based on the height regression features. For this reason, during training, the height prediction loss is computed both for tree and non-tree pixels. However, the height of non-tree pixels is set to zero. Therefore, we apply weights to balance the contribution of tree and non-tree pixels, preventing the loss from being dominated by non-trees.

In this work, we assume that most patches exhibit low variability in pixel values due to the homogeneous forest structure, which may lead to imbalance within a batch.
To solve this issue, instead of directly averaging pixel-wise loss values across the batch, we first average them within each patch and then across patches in the batch while applying tree/non-tree weights $w_\text{tree}$ and $w_\text{non-tree}$ loss as follows:


\begin{equation}
\begin{split}
\mathcal{L}_{\text{batch}} &=
\frac{\sum_{i=1}^{N_b} w_i \, \ell(p_{b,i})}
     {\sum_{i=1}^{N_b} w_i}, \\
\text{where } w_i &=
\begin{cases}
w_\text{tree}=1, & \text{if pixel } i \text{ is tree},\\
w_\text{non-tree}, & \text{otherwise}.
\end{cases}
\end{split}
\end{equation}

$B$ is the batch size, $N_b$ is the number of valid pixels within a patch and $\ell(p_{b,i})$ are pixel-wise loss values. For simplicity, we set $w_\text{tree}$=1 for all the experiments.

\textbf{Gradient difference Loss.}
To improve the quality of the super-resolution, we employ the WGDL, a variant of Gradient Difference Loss (GDL)~\citep{8310638}, which offers an effective balance between memory consumption and prediction accuracy.
GDL loss is computed as L1 or L2 loss between the predicted and target image gradients for adjacent pixels in horizontal and vertical planes. 

We define horizontal and vertical gradients $\nabla_x$ and $\nabla_y$ for predicted and reference images $X$ and $Y$, respectively, as follows:

\begin{equation}
\begin{split}
\nabla_x X_{i,j} = X_{i,j} - X_{i,j-1}, \quad \nabla_y X_{i,j} = X_{i,j} - X_{i-1,j}, \\
\nabla_x Y_{i,j} = Y_{i,j} - Y_{i,j-1}, \quad \nabla_y Y_{i,j} = Y_{i,j} - Y_{i-1,j}.
\end{split}
\end{equation}
where $(i,j)$ are indexed pixel coordinates.

Then the GDL loss is computed as follows:
\begin{equation}
\begin{split}
\mathcal{L}_{\text{GDL}}(X, Y) = 
(&
    \sum_{\substack{i=1 \\ j=1}}^{H, W} \Big|\, |\nabla_x X_{i,j}| - |\nabla_x Y_{i,j}| \,\Big|^\alpha \\
& + \sum_{\substack{i=1 \\ j=1}}^{H, W} \Big|\, |\nabla_y X_{i,j}| - |\nabla_y Y_{i,j}| \,\Big|^\alpha ) / 2 ,
\end{split}
\end{equation}
where $\alpha$ is the chosen loss function. In our work, L2-loss function is chosen.
However, due to the homogeneous forest structure, the latter pixels will contribute more to the loss, while the textured/edge regions will be underrepresented. To address this issue, we propose the context-adapted pixel weighting strategy: the pixel weights are based on the target gradients normalized by the maximum gradient values $max(|\nabla_x Y|)$  and $max(|\nabla_y Y|)$ within each patch. Moreover, to prevent excluding homogeneous regions, we introduce the $\lambda_{min}$ regularization term, to ensure that the weights values are comprised in $[\lambda_{min}, 1]$:

\begin{equation}
\begin{split}
W_{\nabla_x}(i,j) = \lambda_{min} + (1-\lambda_{min})\frac{|\nabla_x Y_{i,j}|}{max(|\nabla_x Y|)}, \\
W_{\nabla_y}(i,j) = \lambda_{min} + (1-\lambda_{min})\frac{|\nabla_y Y_{i,j}|}{max(|\nabla_yY|)} 
\end{split}
\end{equation}

Finally, the Weighted Gradient Difference Loss (WGDL) can be expressed as follows:
\begin{equation}
\begin{split}
\mathcal{L}_{\text{WGDL}}(X, Y) = (&
\sum_{\substack{i=1 \\ j=1}}^{H, W} W_{\nabla_x}(i,j) \cdot \Big|\, |\nabla_x X_{i,j}| - |\nabla_x Y_{i,j}| \,\Big|^\alpha \\
&+ \sum_{\substack{i=1 \\ j=1}}^{H, W}  W_{\nabla_y}(i,j) \cdot  \Big|\, |\nabla_y X_{i,j}| - |\nabla_y Y_{i,j}| \,\Big|^\alpha
) / 2
\end{split}
\end{equation}

Moreover, for WGDL, we apply $w_\text{non-tree}$ weight to non-tree pixels following the same principle as for L1 loss.

{\color{blue}

\subsection{Experimental Setup}
\subsubsection{Evaluation Metrics}
To evaluate the  performance of the model for tree height prediction, we report four complementary metrics: the coefficient of determination (R²), the MAE, the root mean squared error (RMSE), and the relative mean absolute error (rMAE). We additionally  report Intersection over Union (IoU) to assess the quality of the tree classification mask.

R² measures how well the predicted values follow the variability of the true heights, with higher values indicating better explanatory power.
MAE reflects the average magnitude of the prediction errors, while RMSE emphasizes larger errors more strongly and, therefore, highlights occasional large deviations.
The rMAE expresses the average error relative to the true values, enabling comparison across height ranges with different scales. Finally, IoU quantifies the spatial overlap between predicted and reference tree masks, measuring the ratio between their intersection and union, and thus evaluates how accurately the model delineates tree-covered areas.

Moreover, super-resolution predictions may not be perfectly aligned with the reference data, which is a well-known issue in SR evaluation~\cite{11010858}. In particular, lower-resolution Sentinel-2 imagery may exhibit geolocation misalignment of several meters. While such shifts have limited impact at the native resolution, they can lead to pixel-level misalignment between SR predictions and higher-resolution reference data. Therefore, when evaluating our approach against independent LiDAR-derived height maps, we employ shifted evaluation metrics for 5 and 2.5~m models. These metrics are computed by evaluating several spatial offsets within a local neighborhood and selecting the alignment yielding the lowest MAE. The remaining evaluation metrics are subsequently computed using this optimal shift.


\subsubsection{Super-resolution quality assessment}
To evaluate the quality of super-resolution (SR) frequency restoration, we use the Frequency Attenuation Profile (FAP) metric based on Discrete Fourier Transform (DFT)~\citep{11010858}, as it is not sensitive to cross-dataset distortions and provides a robust frequency-domain comparison. FAP indicates how strongly each spatial frequency component is preserved or attenuated in an image. Low frequencies correspond to large-scale structures, whereas high frequencies capture fine textures and edges. By comparing the attenuation curves of the HR reference, the SR prediction, and the lower resolution input upsampled with bicubic interpolation, we can assess how effectively the model restores or preserves spatial details.

\subsubsection{Additional experiments}
To assess the generalizability of the proposed model, we compare its predictions against independent LiDAR-derived reference data and state-of-the-art methods.

We also conduct an ablation study to evaluate the influence of different loss functions -- specifically patch-wise averaging, the WGDL loss, and the weight assigned to non-tree pixels ($w_{\text{non-tree}}$) -- on model performance.

\subsubsection{Model Configuration}

We evaluate THREASURE-NET at three spatial resolutions: 10 m, 5 m, and 2.5 m. 
For training we use both tree and non-tree pixels, considering the pixels with values superior to 1.5~m as trees (permanent vegetation). 
However, the tree height regression validation metrics  are exclusively computed over tree pixels derived from reference data, whereas the IoU metric is evaluated separately to assess the accuracy of tree/non-tree classification.

All models share the same high-level architecture, which includes an RDB feature extraction block, a temporal transformer for multi-temporal fusion, a super-resolution block, and an MLP prediction head.
For the 10~m model, the super-resolution block is omitted, as no upscaling is required. Hyperparameters were chosen based on preliminary experiments and practical constraints associated with the extensive training time, with an additional objective of minimizing prediction error.

The RDB-FE block uses 5 RBD blocks with 5 layers each in the 5~m and 2.5~m models, and 4 dense blocks with 4 layers in the 10~m model. The channel growth rate and output feature dimension are fixed at 24 and 64 across all models. The 10~m model uses a smaller RDB configuration since no super-resolution step is applied, while increasing the RDB size for the 2.5~m model did not yield performance improvements during training.

The temporal transformer employs 4 attention heads and a hidden dimension of 16 to process the multi-temporal inputs. For positional encoding, a scaling constant of $\tau= 365$ is used for both LiDAR and Sentinel-2 acquisition dates.

The super-resolution block applies upscaling factors of r = 2 and r = 4 for the 5~m and 2.5~m models, respectively, with the convolutional filter size increasing from 64 to 256 for each upscaling step.

Feature sizes $F_1$ and $F_2$ are both set to 64. Temporal transformer parameters are set to 4 heads with feature size of 16 for each head.

Finally, the prediction head consists of an MLP with layer sizes (64, 128, 64) followed by two task-specific output heads: one producing a single-channel tree-height regression map and the other generating a binary classification mask.

The maximum time-series length is set to 12, and all series shorter than 5 observations are discarded. During training, a random sampling strategy is applied when the time-series length exceeds 12, whereas an equal-range sampling strategy is used during validation to ensure the consistency of the validation data.

The input Sentinel-2 patches contain $64 \times 64$ pixels ($640\times640$~m²), augmented with border margins that are passed through the model but discarded from the output before loss computation and inference, in order to avoid border effects due to 2D convolution operations.
During training, we apply a random sampling strategy within the borders of each LiDAR HD patch ($1 \times 1$ km²), whereas during validation, the central region of the patch is systematically selected. For this reason, the initial Sentinel-2 patches were constructed with 250~m margins on each side relative to the LiDAR patches.

As mentioned in Section~\ref{sec:method}, patch-weighted L1 loss, BCE and WGDL losses are used for model optimization. The loss weights were set as follow: $\omega_{height}=1$, $\omega_{classif}=1$, and $\omega_{wgdl}=0.33$. The weight for non-tree pixels is set to $w_\text{non-tree}=0.1$ for both patch-weighted L1 and wGDL losses.

All input features were standardized to ensure stable training.

Model training and evaluation were performed on the CNES TREX HPC cluster, equipped with NVIDIA A100 and H100 GPUs. The implementation used PyTorch~2.5.1 with CUDA~11.8. Experiments were executed in mixed-precision mode to improve computational efficiency.

The network was trained using the Adam optimizer with a cosine annealing with warm restarts scheduler \citep{warmrestart}, where the learning rate decays at the beginning of each new cycle by a factor of 0.25 with the initial learning rate set to $10^{-3}$. 
We used a batch size of 32 and accumulated gradients over 4 steps, resulting in an effective batch size of 128.

The training took roughly 3 days for each model.

}

\section{Results and Discussion}

\subsection{Results}
\subsubsection{Quantitative Results}

\begin{table*}[!htbp]
    \centering
    \caption{Quantitative results for our models of 10, 5 and 2.5~m resolutions at the test set. We also report the performance of two other recent state-of-the-art algorithms developed for french temperate forests: FORMS-T~\citep{SCHWARTZ2025114959} at 10m and Open-Canopy~\citep{11095116} at 1.5m resolution. N/R stands for ``Not Reported''. Here we report the performance of concurrent methods for reference purposes only.}
    \label{tab:results}
    \begin{tabular}{|l|c|c|c|c|c|c|}
    \hline
        Model & Res.,~m & MAE,~m & MAE, \% & $R^2$       & RMSE,~m       & IoU, \% \\ \hline
        \multirow{3}*{THREASURE-Net}
            & 10  & 2.63 & \textbf{22.3} & \textbf{0.75} & \textbf{3.55} & 88.5 \\ \cline{2-7}
            & 5   & 2.70 & 25.3          & 0.72          & 3.66          & 86.7 \\ \cline{2-7}
            & 2.5 & 2.89 & 29.7          & 0.67          & 3.92          & 84.7 \\ \hhline{|=|=|=|=|=|=|=|}
        Open-Canopy
            & 1.5 & \textbf{2.52} & 22.9 & N/R & 4.02 & 90.5$^{*}$ \\ \hline
        FORMS-T
            & 10  & 3.06 & N/R & 0.68 & N/R & 76.8$^{*}$ \\ \hline
    \end{tabular}
\\$^{*}$The authors report tree classification within a forest mask, while we perform tree/non-tree classification for the whole test dataset.
\end{table*}

\begin{figure*}[!htbp]
    \centering
    \includegraphics[width=\linewidth]{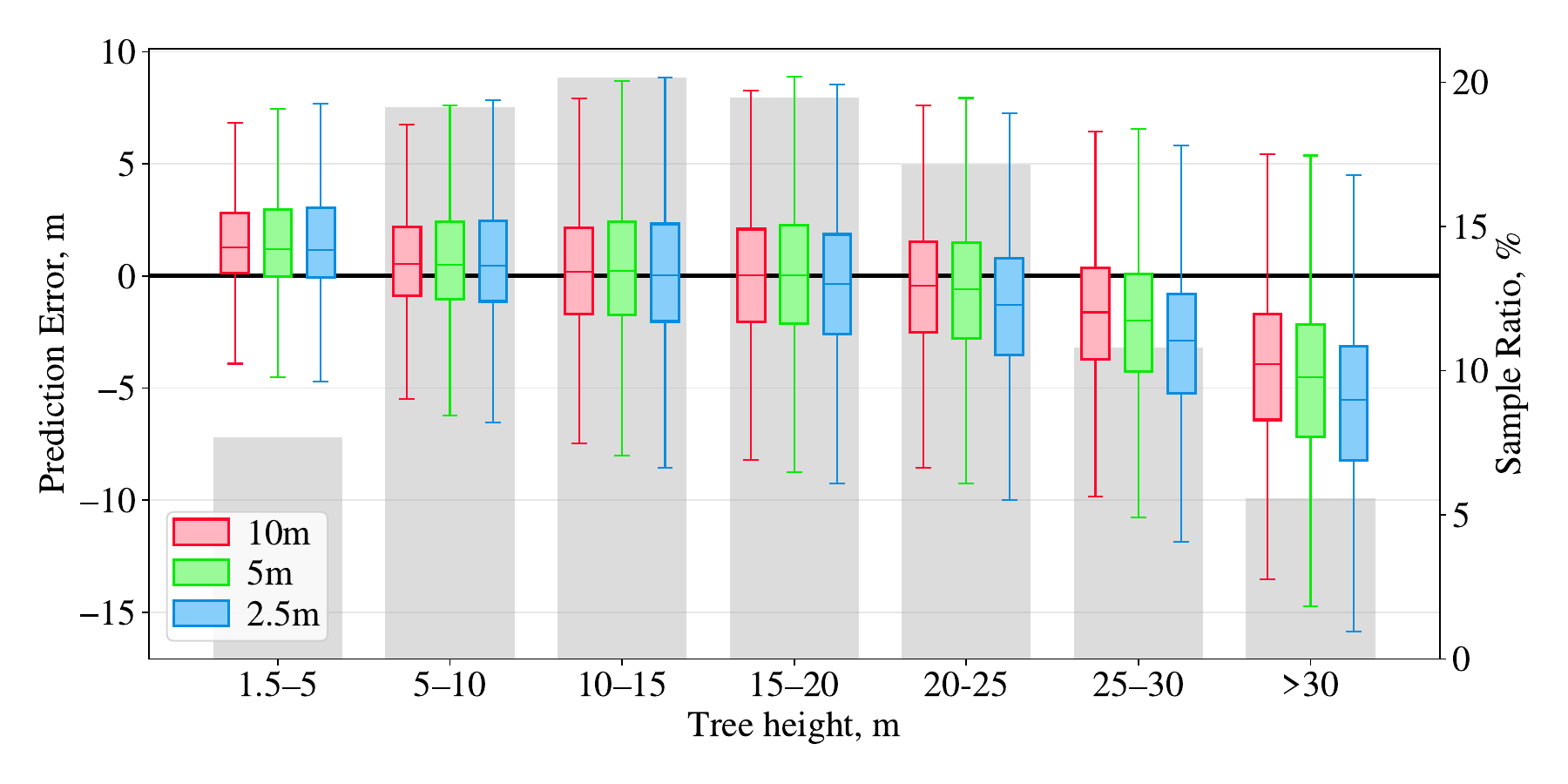}
    \caption{Error distribution across canopy-height bins (1.5-5, 5-10, 10-15, 15-20, 20-25, 25-30, $>$30 m) for our models at different output resolutions in a form of box-and-whisker plots. A boxplot shows the median and the interquartile range (IQR) (from the 1st to the 3rd quartile), with whiskers extending to the most extreme values within 1.5$\times$IQR. Gray bars indicate the proportion of samples falling into each bin relative to the entire dataset.}
    \label{fig:error_bins}
\end{figure*}

\begin{figure*}[!htbp]
    \centering
    
    \begin{subfigure}[t]{0.32\textwidth}
        \centering
        \includegraphics[width=\linewidth]{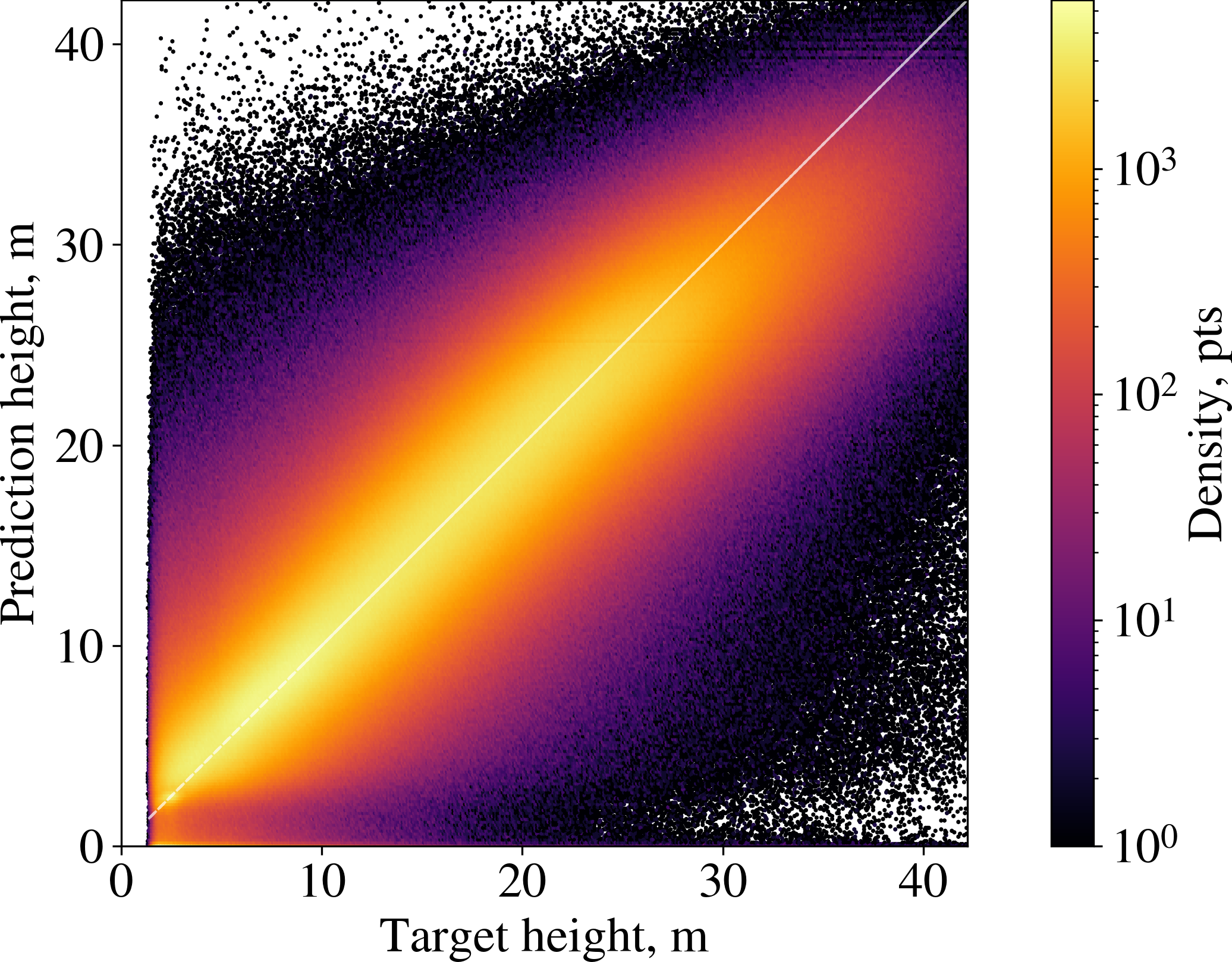}
        \caption{10 m}
        \label{fig:scatter_10m}
    \end{subfigure}
    \hfill
    \begin{subfigure}[t]{0.32\textwidth}
        \centering
        \includegraphics[width=\linewidth]{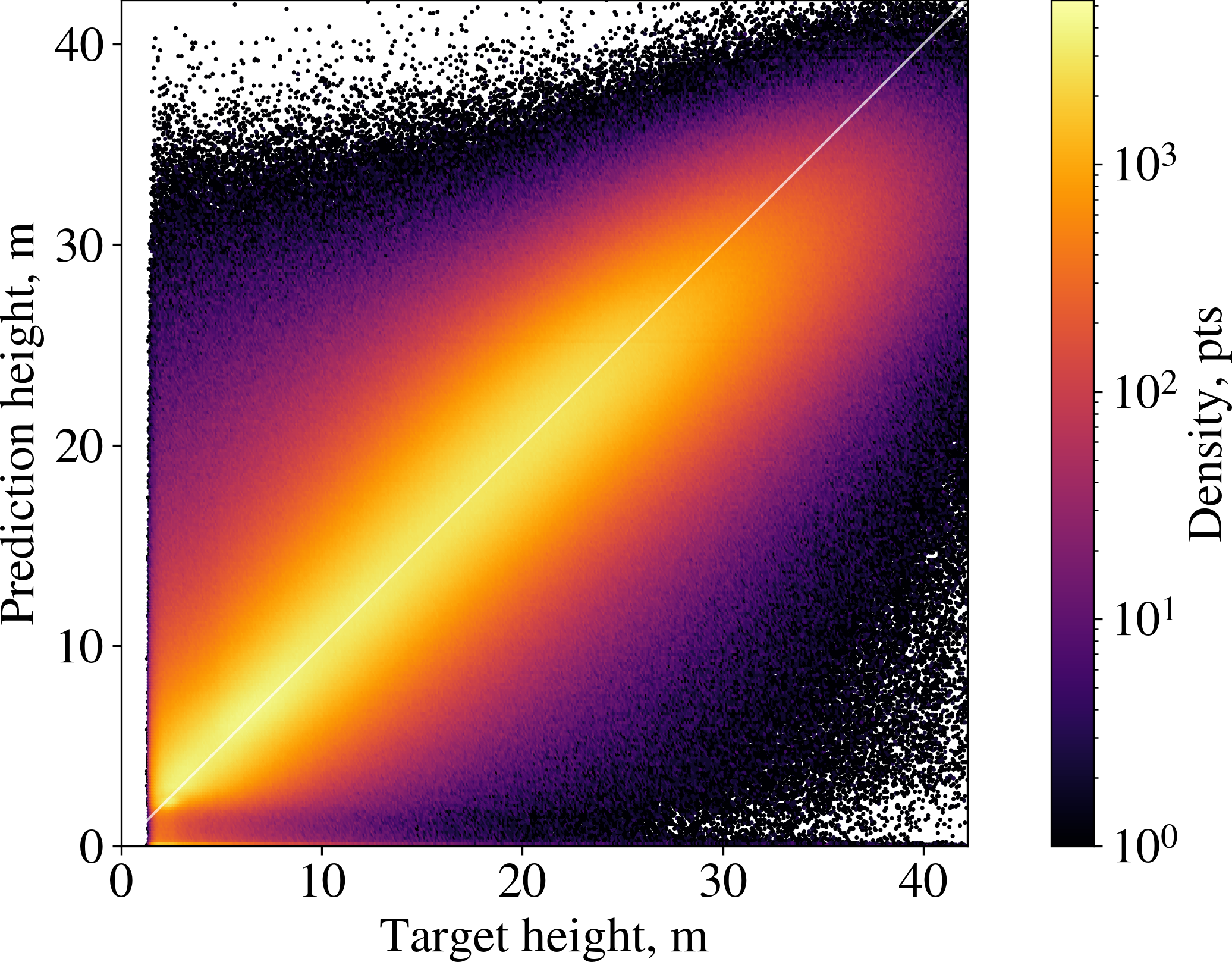}
        \caption{5 m}
        \label{fig:scatter_5m}
    \end{subfigure}
    \hfill
    \begin{subfigure}[t]{0.32\textwidth}
        \centering
        \includegraphics[width=\linewidth]{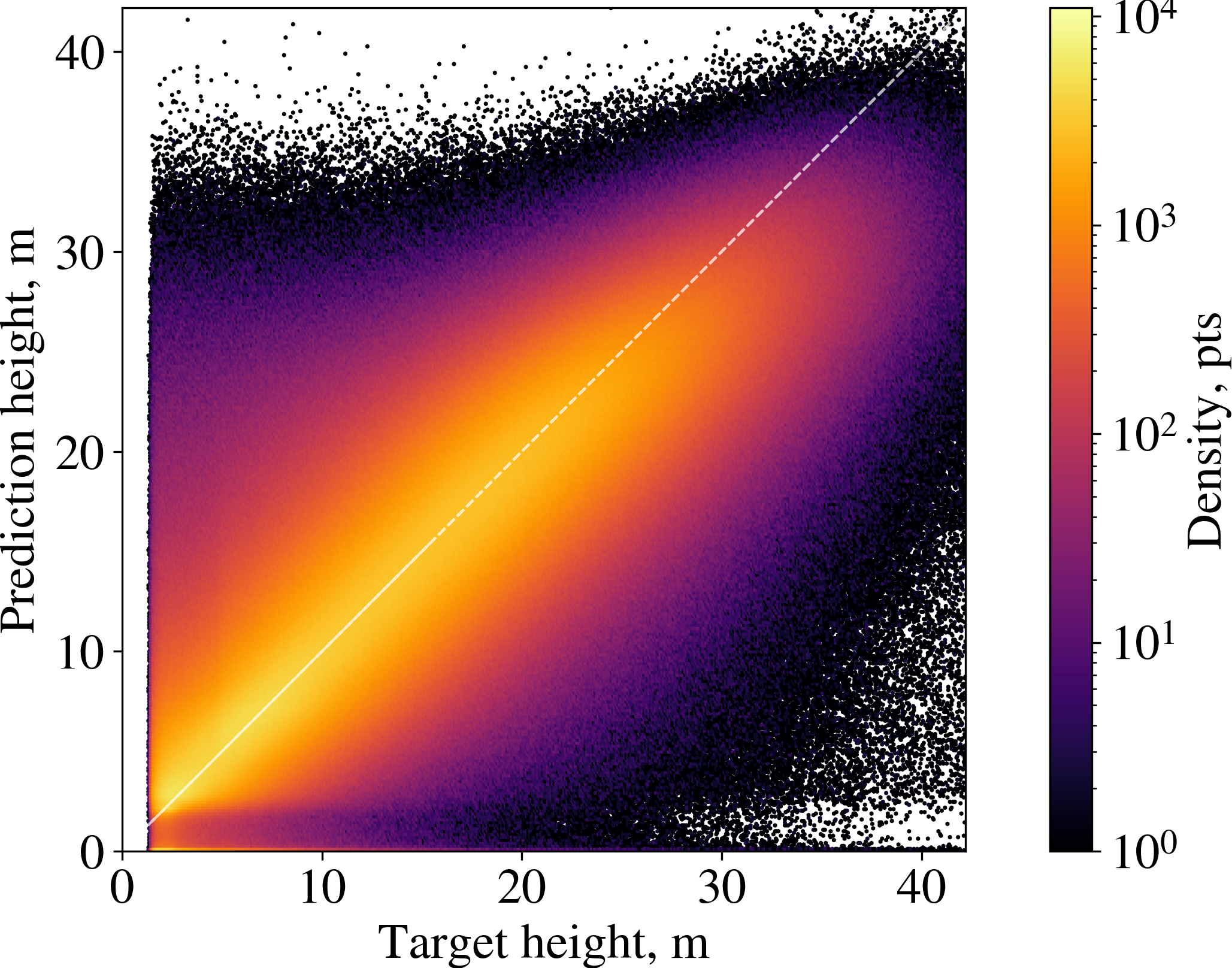}
        \caption{2.5 m}
        \label{fig:scatter_2_5m}
    \end{subfigure}

    \caption{Prediction–target scatterplots for our 10 m, 5 m, and 2.5~m models for tree pixels. For visualization purposes, we randomly sampled an equal number of points ($\approx$96 million) from each model, corresponding to the total number of points in the 10~m dataset. The identity line is shown in white. Note that although the minimum target values in the dataset are set to 1.5 m, we report the model predictions without applying any clipping.}
    \label{fig:scatter_all}
\end{figure*}

In this subsection, we report the quantitative results of our model performance at different resolutions for the test set. The different metrics are reported in Table~\ref{tab:results}. 
For reference, we report the performance of two recent state-of-the-art algorithms developed for French temperate forests, FORMS-T~\citep{SCHWARTZ2025114959} and Open-Canopy~\citep{11095116}, using results reported in their original publications. A direct comparison with our method on independent LiDAR data is provided in Section~\ref{subsec:comparison}.

Our model achieves mean absolute errors of 2.63~m, 2.70~m, and 2.89~m at 2.5~m, 5~m, and 10~m resolution, respectively.
Although Open-Canopy obtains the lowest MAE overall, its RMSE is higher than the RMSE achieved by THREASURE-Net at each of the three resolutions. This indicates that Open-Canopy performs well on average values but struggles more with the extreme ones, tending to regress predictions toward the mean.
In our case, quantitative performance decreases as the target resolution becomes finer. Nevertheless, the 2.5~m version of our model remains competitive with Open-Canopy, despite relying exclusively on freely available 10~m Sentinel-2 imagery. By contrast, Open-Canopy is trained on 1.5~m SPOT-6/7 data, which are commercial products or accessible only through specific research agreements. This highlights an important advantage of our approach: comparable performance can be achieved without relying on costly high-resolution remote sensing input data.
Moreover, all variants of THREASURE-Net reached better results than 10~m resolution FORMS-T, despite relying solely on Sentinel-2 data, whereas FORMS-T uses Sentinel-1 and Sentinel-2 data, which requires considerably heavier data processing.

Figure~\ref{fig:error_bins} shows the error distribution across tree-height bins (1.5–5, 5–10, 10–15, 15–20, 20–25, 25–30, and $>$30 m) using box-and-whisker plots. Overall, all three resolutions exhibit similar behavior, with a slight degradation in performance for the smallest and tallest trees. In contrast, for mid-range heights (10-15 and 15-20 m), the models perform nearly identically, with median errors close to zero.
Moreover, all models tend to overestimate low height values, while for higher values they exhibit the opposite behavior.
The most noticeable performance differences between the models occur above 25 m, where their accuracy drops more sharply than in the mid-range bins. This may be explained by the uneven distribution of tree heights in the dataset or by the limitations of Sentinel-2 in distinguishing fine spectral differences in short and tall forest stands -- an issue that could be addressed in future work.

Figure~\ref{fig:scatter_all} presents the prediction–target scatterplots for our models. The 10~m model produces predictions that cluster closely around the identity line, indicating strong agreement with the target values. For the 5~m and 2.5~m models, the point dispersion becomes increasingly noticeable.
Interestingly, although the 2.5~m model exhibits the largest overall spread around the identity line, it produces fewer extreme errors than the 10~m and 5~m models. 

The superior performance of Open-Canopy in forest mask classification (IoU) at all resolutions suggests that the classification component of our model could be further optimized.

\rev{We equally evaluate out algorithm's performance on tree loss detection task and compare it with state-of-the-art methods FORMS-T~\citep{SCHWARTZ2025114959}, Open-Canopy~\citep{11095116}, and Global Canopy Height Model~\citep{lang2023high} in~\ref{apx:change}.}

\subsubsection{Qualitative Results}

In this subsection, we qualitatively evaluate the capacity of our algorithm to produce sharp and fine-grained textures at 5 and 2.5~m resolutions. Evaluating super-resolution can be a challenging task, as the evaluation metrics often do not take into account geometric distortions between two datasets~\citep{11010858}. Moreover, in our case, in the absence of similar algorithms, SR metrics might be difficult to interpret for our results alone. Therefore, we present visual results of our algorithm as well as Frequency Attenuation Profiles (FAP) for the SR task. 

\begin{figure*}[!htbp]
\begin{minipage}[c]{0.949\textwidth}
    
  \centering
  \begin{subfigure}{0.235\textwidth}
    \includegraphics[width=\linewidth]{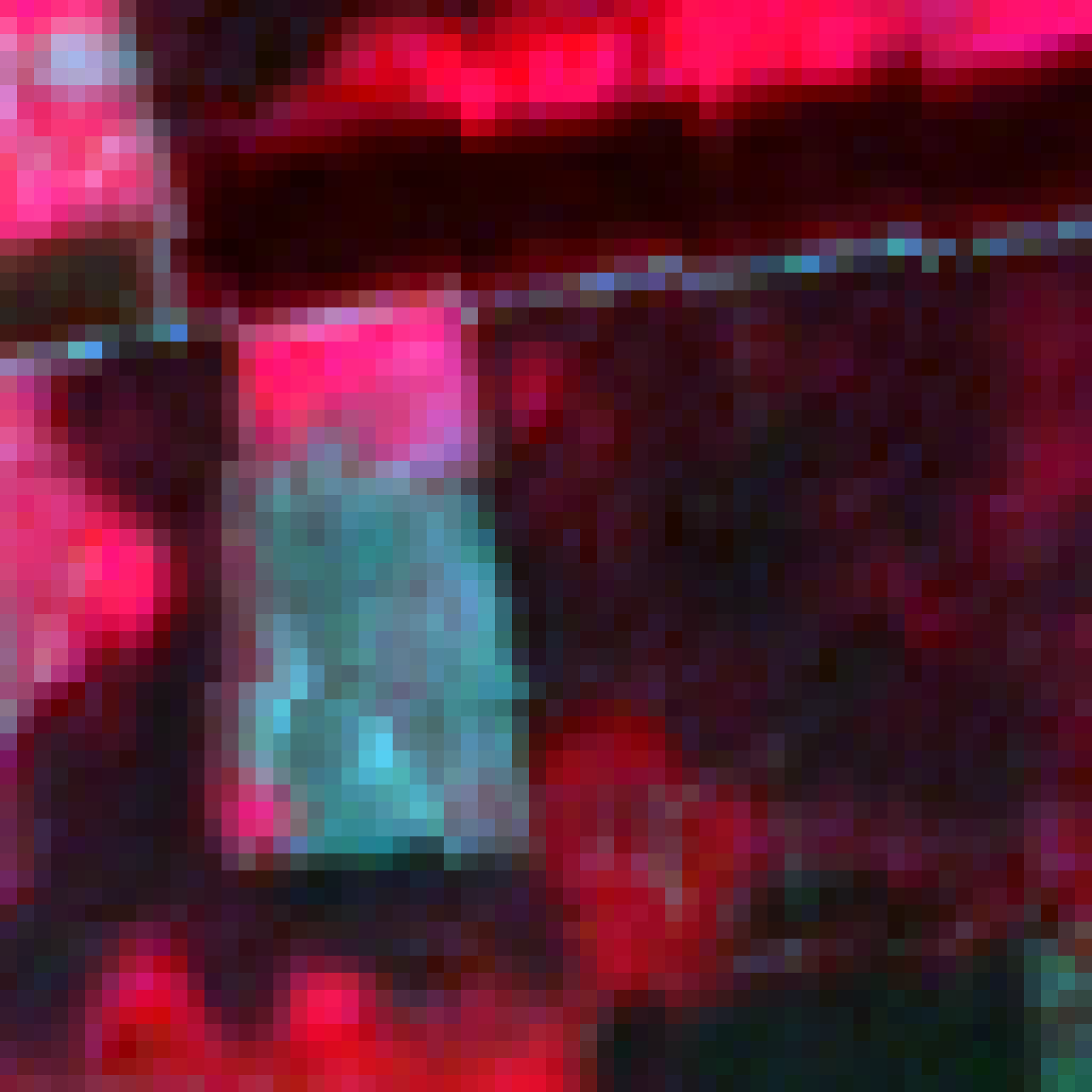}
    \caption{Sentinel-2 image}
  \end{subfigure}
  \begin{subfigure}{0.235\textwidth}
    \includegraphics[width=\linewidth]{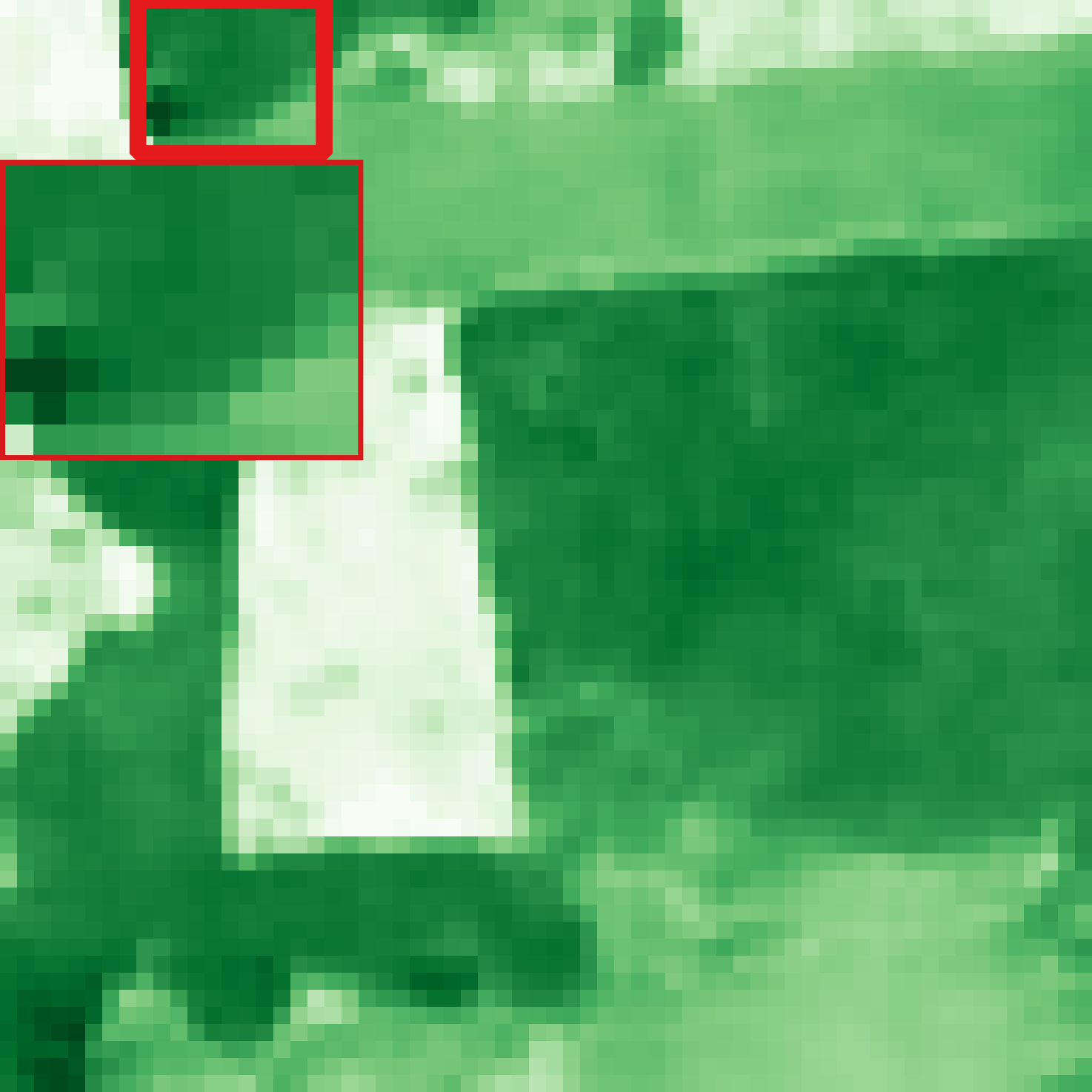}
    \caption{10m prediction}
    \label{fig:10m}
  \end{subfigure}
  \begin{subfigure}{0.235\textwidth}
    \includegraphics[width=\linewidth]{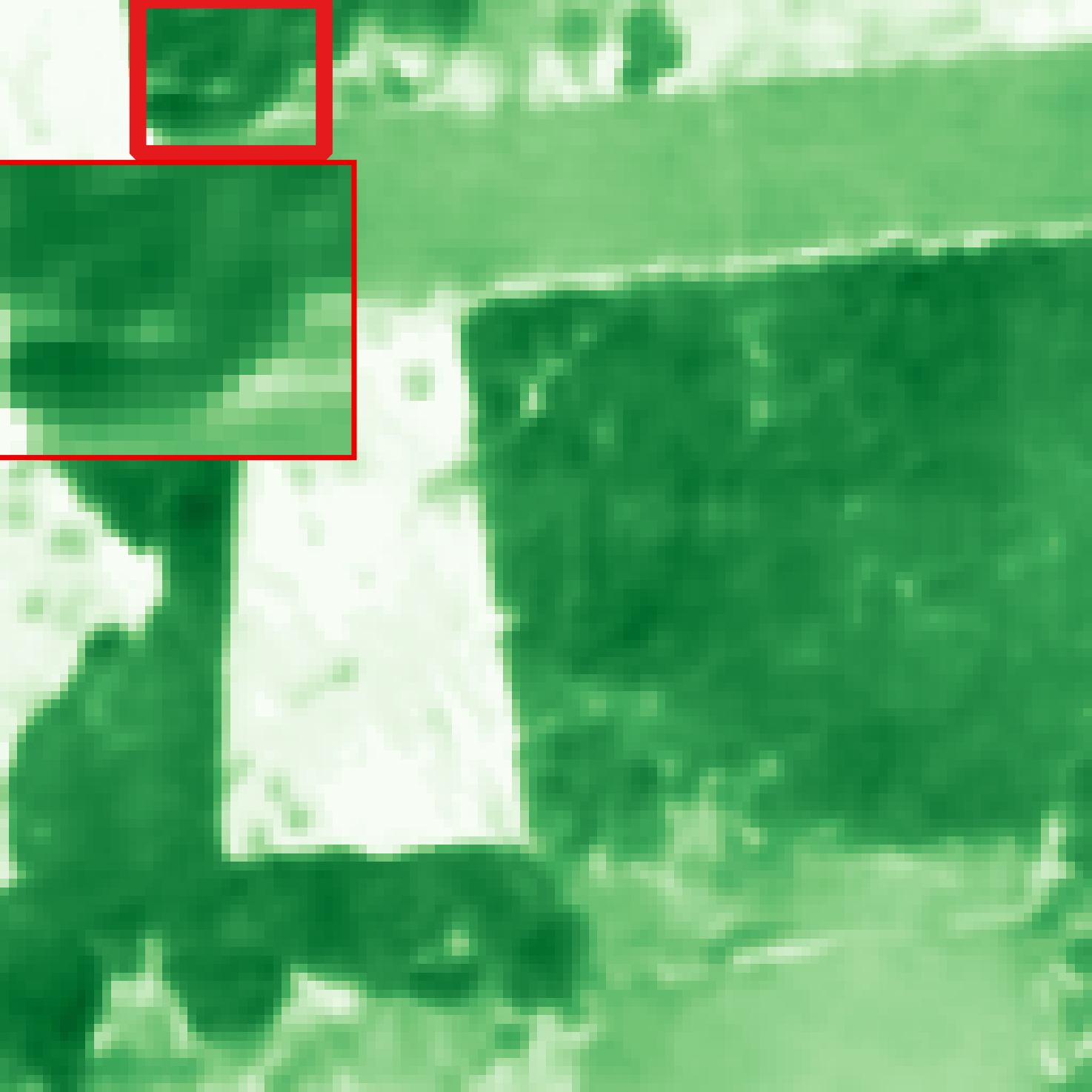}
    \caption{5m prediction}
    \label{fig:5m}
  \end{subfigure}
  \begin{subfigure}{0.235\textwidth}
    \includegraphics[width=\linewidth]{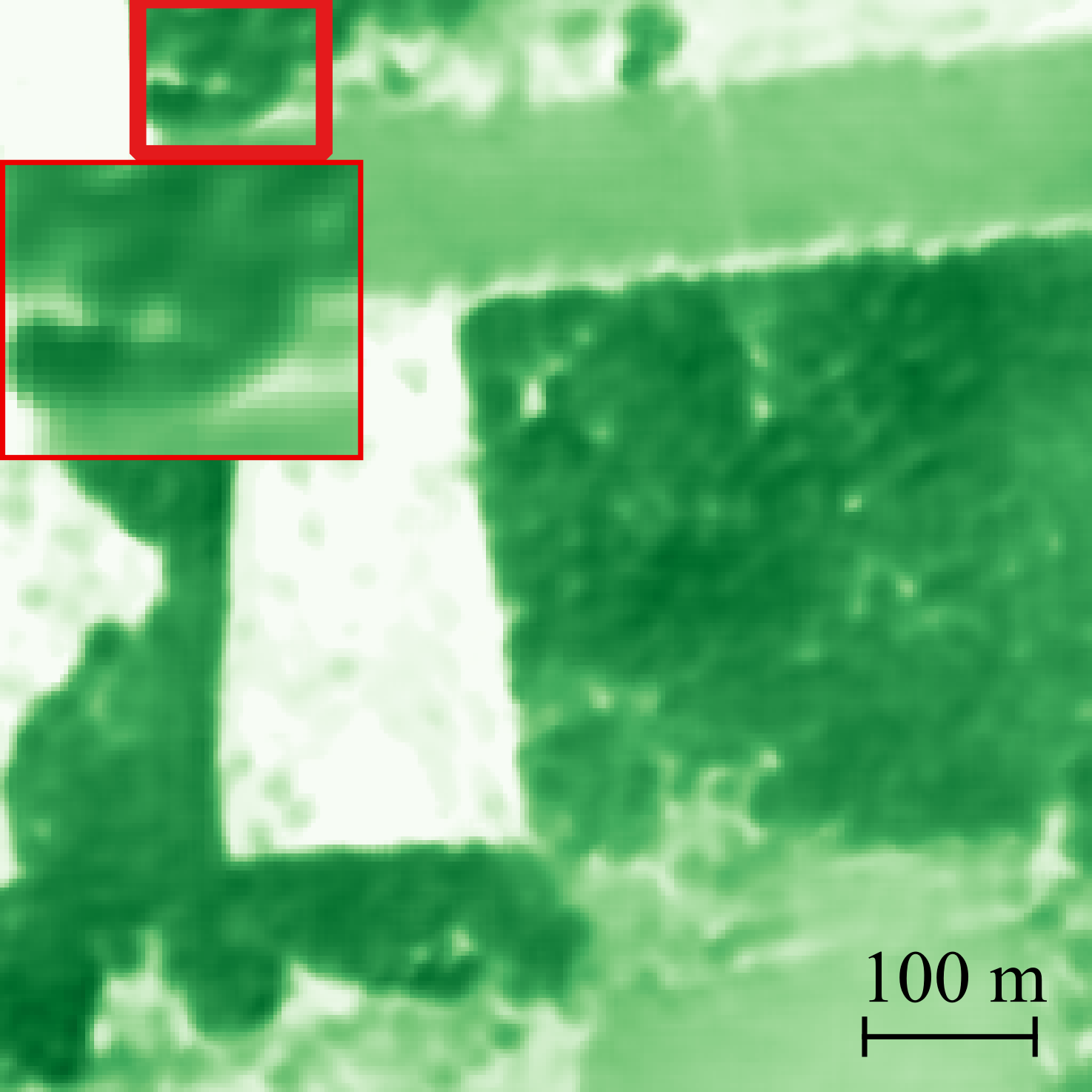}
    \caption{2.5m prediction}
    \label{fig:2.5m}
  \end{subfigure}
  
  \vspace{1.5em} 

  \begin{subfigure}{0.235\textwidth}
    \includegraphics[width=\linewidth]{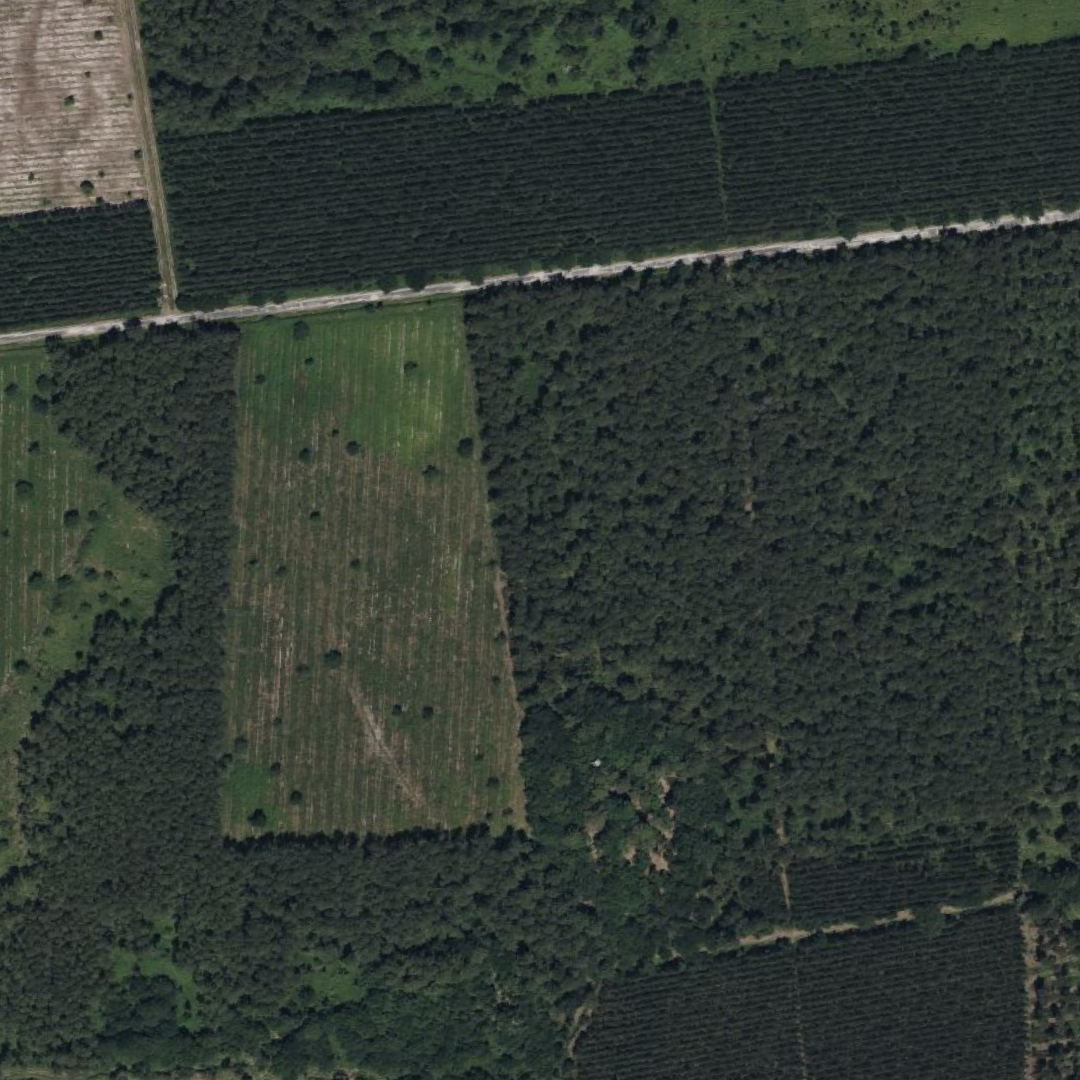}
    \caption{Bing image}
  \end{subfigure}
  \begin{subfigure}{0.235\textwidth}
    \includegraphics[width=\linewidth]{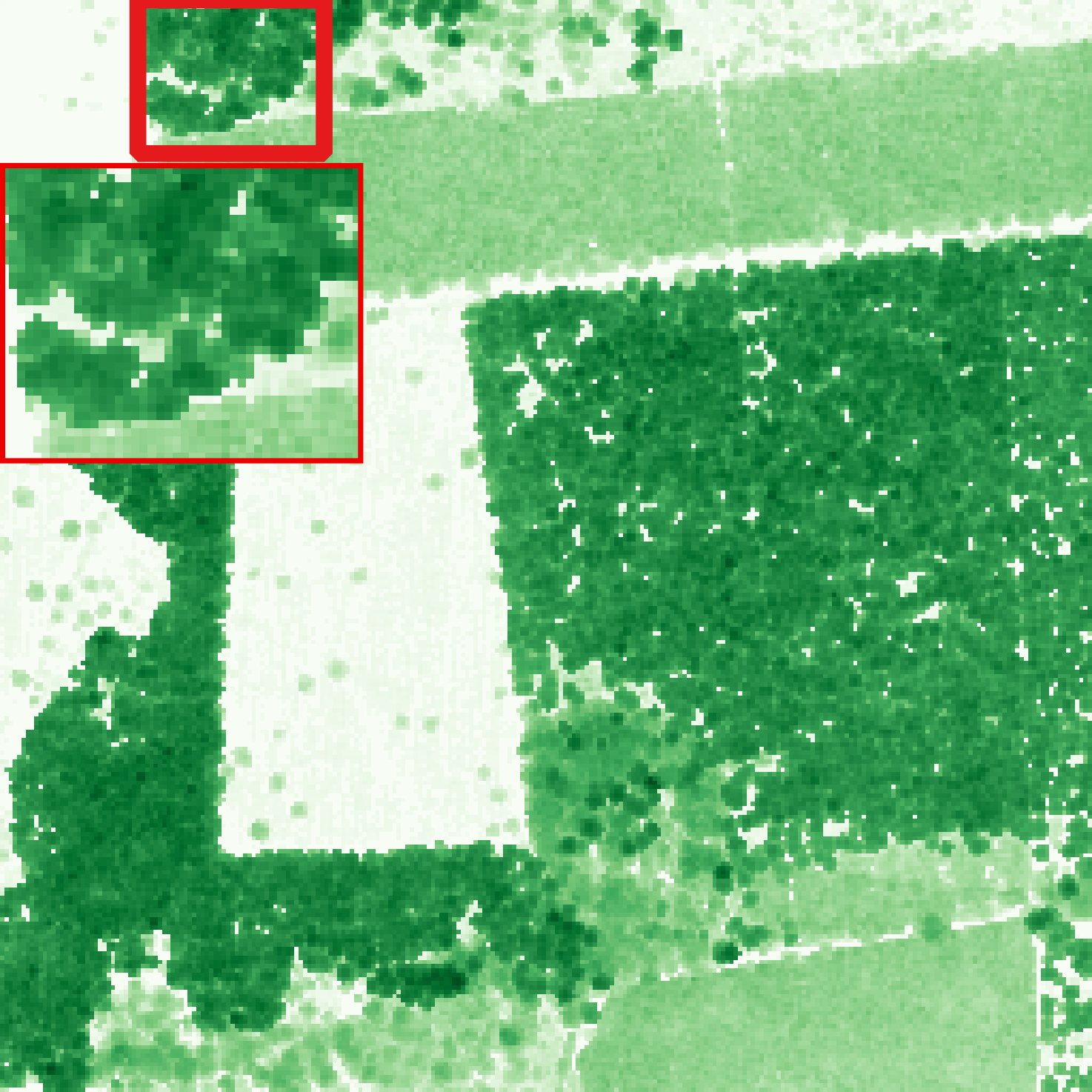}
    \caption{2.5m reference}
  \end{subfigure}
  \begin{subfigure}{0.235\textwidth}
    \includegraphics[width=\linewidth]{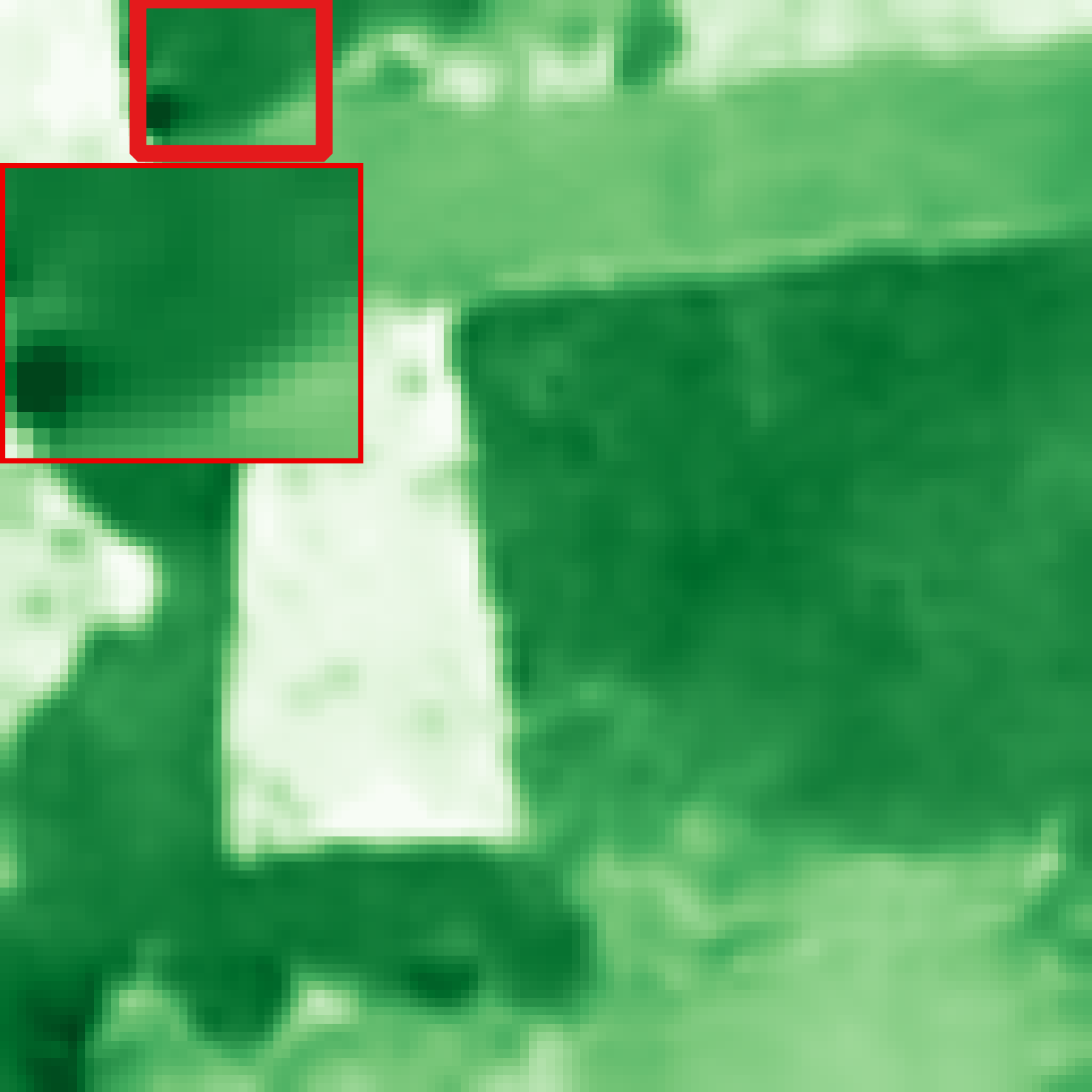}
    \caption{10m pred. upsampled to 5m}
    \label{fig:up5m}
  \end{subfigure}
  \begin{subfigure}{0.235\textwidth}
    \includegraphics[width=\linewidth]{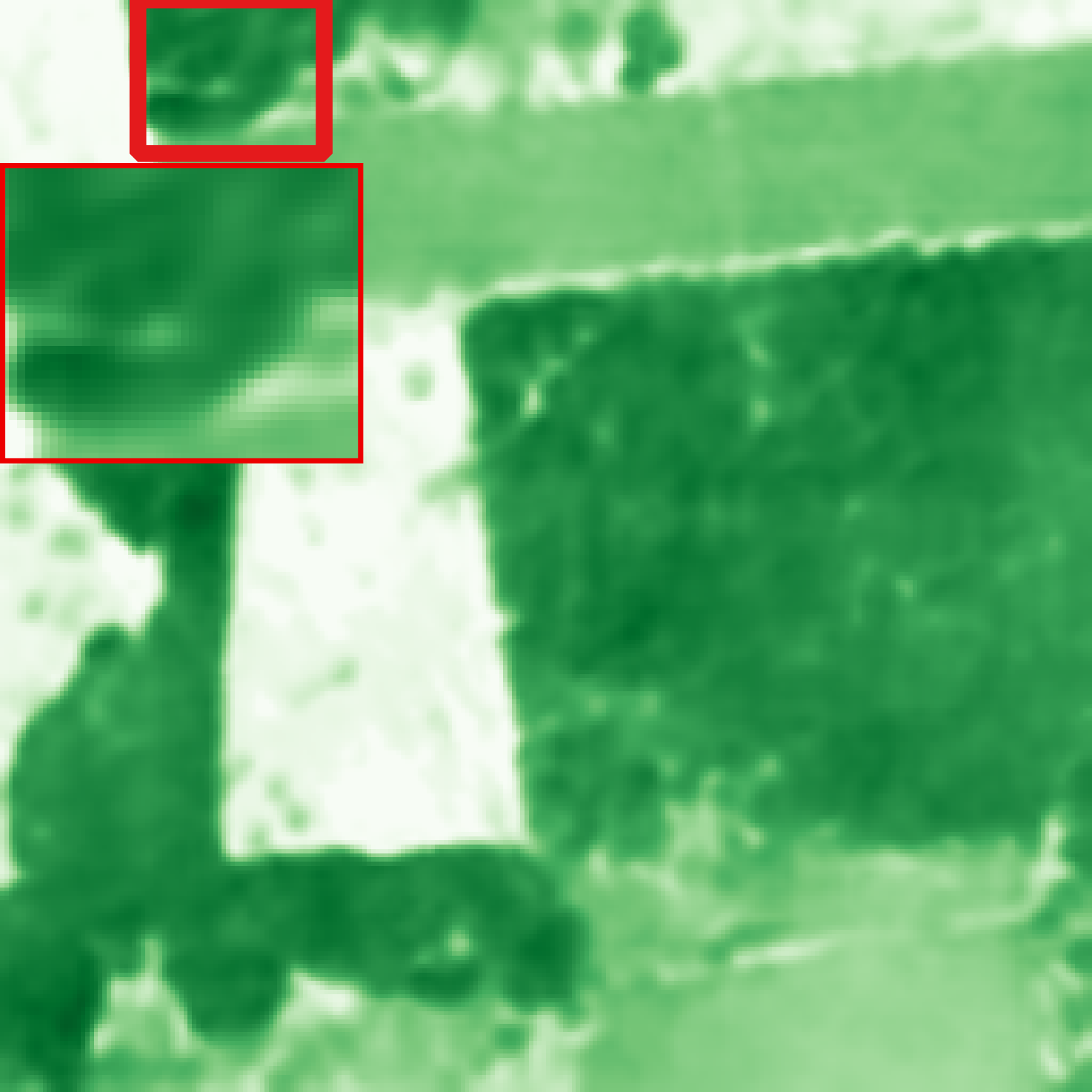}
    \caption{5m pred. upsampled to 2.5m}
    \label{fig:up2.5m}
  \end{subfigure}
    \end{minipage}%
    \hspace{0.00001\textwidth}%
    \begin{minipage}[c]{0.05\textwidth}
      \raggedleft
      \includegraphics[width=\linewidth]{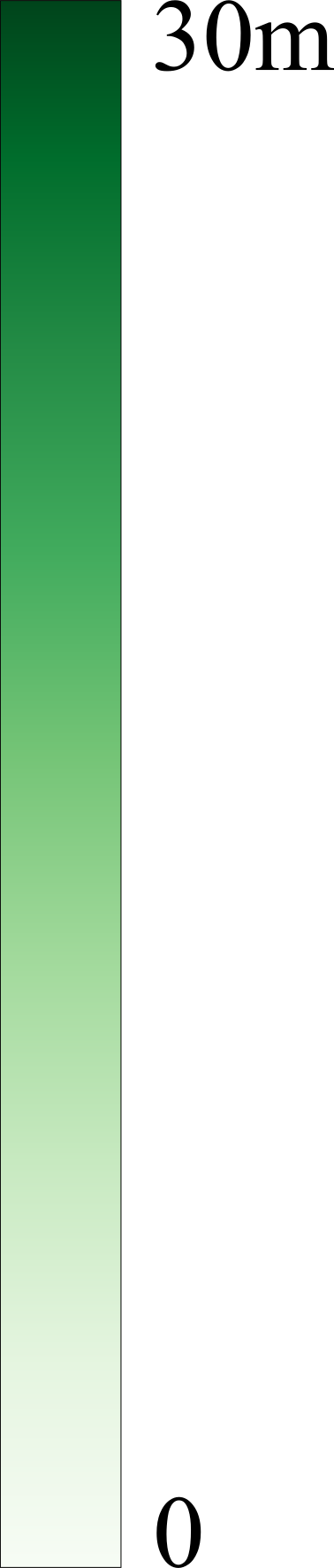}
    \end{minipage}

  \caption{
    $640 \times 640$~m tree height prediction patches centred on 44.893°N and 0.997°W.
    Subfigure (a) shows a Sentinel-2 image from the input time series acquired on 7 July 2023, and (b) provides a VHR Bing image for visual comparison.
    Subfigures (c)–(e) present the prediction results of our model at 10 m, 5 m, and 2.5~m resolutions, respectively.
    Subfigures (f) and (g) display the 10 m and 5~m predictions upsampled to 5 m and 2.5 m, respectively, using bicubic interpolation.
    Finally, (h) depicts the reference LiDAR-derived height map at 2.5~m resolution.
    Zoomed-in views (top-left corner) highlight differences in fine details between predictions. No masking is applied in these examples.
    }
  \label{fig:superregression}
\end{figure*}

Figure~\ref{fig:superregression} illustrates the predictions obtained for a $640\times640$~m forest patch located in the south-west of France. A progressive gain in spatial detail can be observed when moving from the 10~m prediction (Subfigure~\ref{fig:10m}) to the 5~m (Subfigure~\ref{fig:5m}) and finally to the 2.5~m prediction (Subfigure~\ref{fig:2.5m}). For comparison, we additionally provide bicubic upsampling of the 10~m and 5~m models outputs to 5~m (Subfigure~\ref{fig:up5m}) and 2.5~m (Subfigure~\ref{fig:up2.5m}), respectively. This allows us to highlight the benefit of deep-learning-based super-resolution compared with a classical interpolation method.

Although the super-resolved predictions do not reach the full level of detail present in the reference image, it is nevertheless possible to distinguish tree crown shapes and fine-scale structures such as forest gaps. The improvement from 10~m to 5~m resolution is particularly noticeable, whereas the enhancement from 5~m to 2.5~m is more subtle. When comparing the bicubic 5$\rightarrow$2.5~m upsampling with the 2.5~m super-resolved output, small gaps appear sharper in the latter, but it introduces little new structural information. This behavior is consistent with the limited amount of spatial information available in Sentinel-2 imagery, which constrains the degree of recoverable fine-scale detail.

\begin{figure*}[!htbp]
    \centering
    
    \begin{subfigure}{0.495\textwidth}
        \centering
        \includegraphics[width=\linewidth]{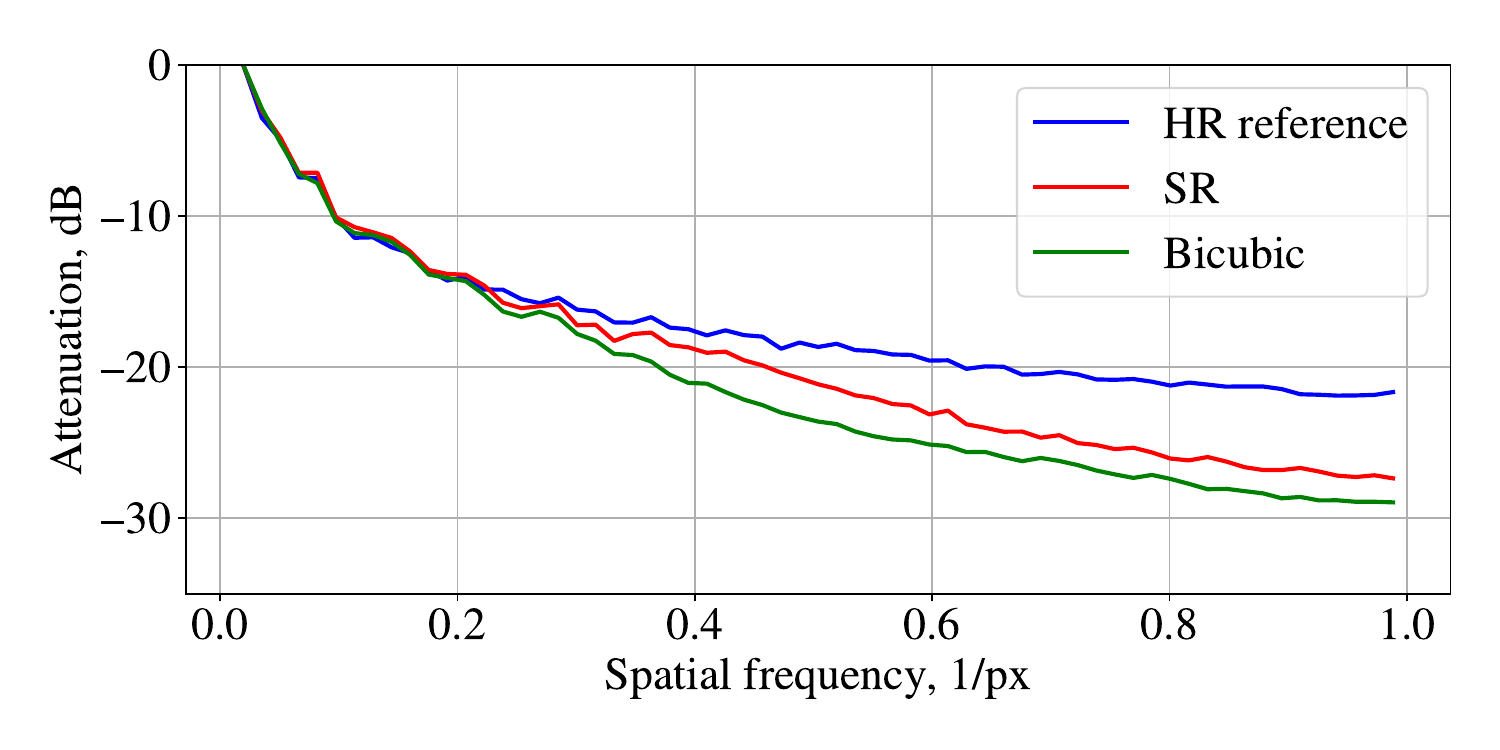}
        \caption{5 m}
        \label{fig:fda_5m}
    \end{subfigure}
    \hfill
    \begin{subfigure}{0.495\textwidth}
        \centering
        \includegraphics[width=\linewidth]{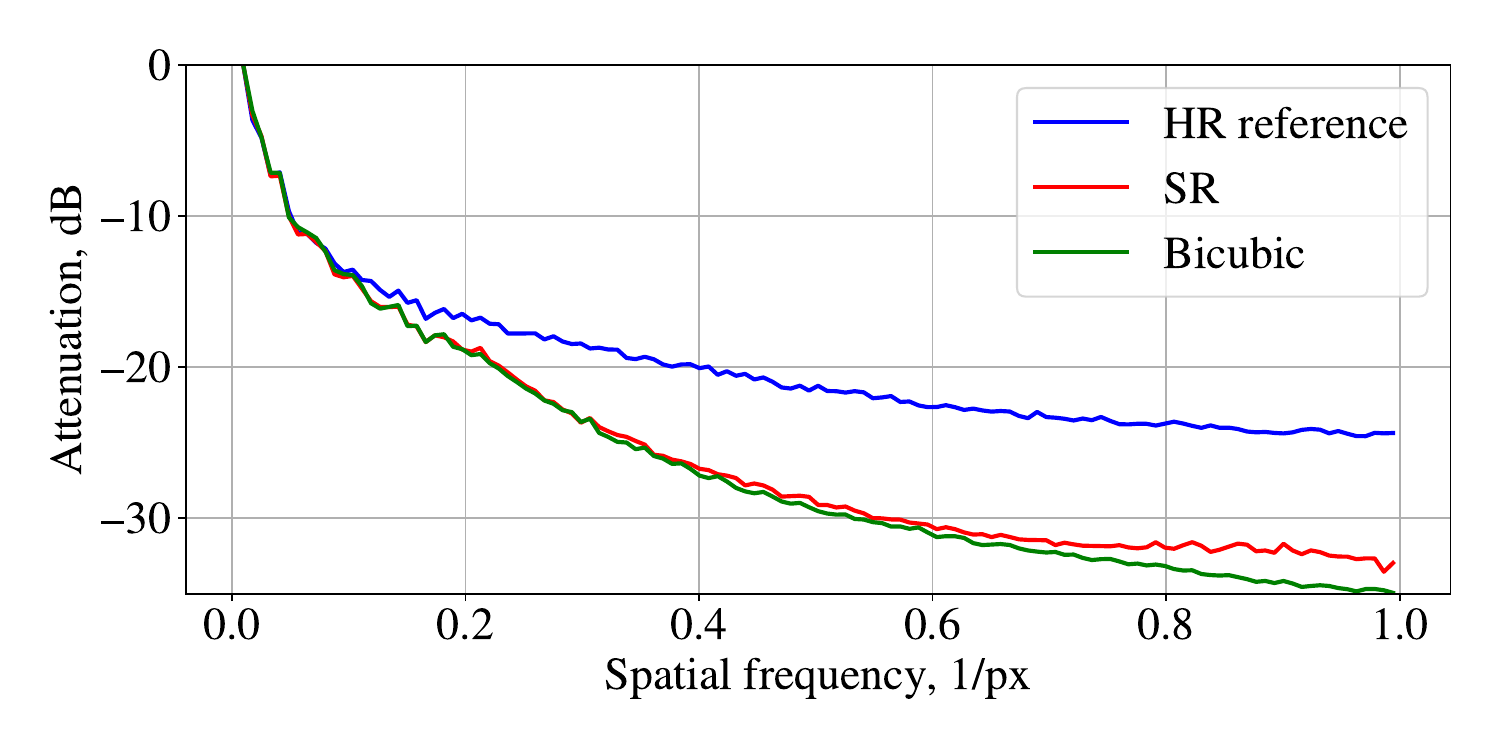}
        \caption{2.5 m}
        \label{fig:fda_2.5m}
    \end{subfigure}

    \caption{Normalized log-Frequency Attenuation Profile FAP for HR reference patch, SR results obtained with THREASURE-Net and bicubic interpolation for 5 m (a) and 2.5 m (b) predictions for patch presented in Figure~\ref{fig:superregression}. Bicubic interpolation profiles correspond to Subfigures~\ref{fig:up5m} and \ref{fig:up2.5m} respectively. To maintain consistency, we used HR reference data with the same native resolution for both predictions. Spatial frequencies in x axis correspond to normalized spatial frequencies $f = \frac{f_n}{f_N}$, 
        where $f_N$ is the maximum spatial frequency that can be represented at a given image resolution.}
    \label{fig:fda}
\end{figure*}

Figure \ref{fig:fda} presents the normalized log-Frequency Attenuation Profiles for the 5 m and 2.5 m SR predictions, the corresponding HR reference data, and the 10 m and 5 m THREASURE-Net predictions upsampled to 5 m and 2.5 m, respectively, using bicubic interpolation. The results are presented for the same patch as in Figure~\ref{fig:superregression}.
For both resolutions, SR predictions exhibit a progressive attenuation as frequency increases, reflecting the loss or smoothing of fine spatial details relative to the HR reference, with stronger degradation at 2.5 m. At 5 m, the SR model consistently outperforms bicubic upsampling across most of the frequency spectrum. At 2.5 m, although improvements over bicubic interpolation are modest in the mid-frequency range, the SR output still shows a clear advantage at higher frequencies.

\rev{One potential application of the improved resolution of THREASURE-Net is the analysis of multi-year tree growth. This idea is explored further in~\ref{apx:tree_regrowth}. Without multi-year validation data, the results cannot be evaluated. However, the annual tree height time series shows a consistent growth trend, suggesting the potential for monitoring tree growth.}




\subsection{Ablation Study}

To justify the benefits of the selected loss functions, we assess their individual contributions relative to a baseline L1 loss. We conduct an ablation study on the 5~m resolution model considering four variants: (i) training with the standard L1 (MAE) loss only; (ii) training with the patch-average L1 loss only (here referred to as patch-weighted MAE, pwMAE); (iii) the proposed configuration combining wGDL with pwMAE; and (iv) the proposed configuration combining wGDL with pwMAE using an increased non-tree pixel weight of \( w_{\text{non-tree}} = 0.5 \). Note that in both models (i) and (ii) we apply a non-tree weight of \( w_{\text{non-tree}} = 0.1 \), as in our proposed configuration (iii).

\begin{table*}[!htbp]
\centering
\caption{Ablation study of different loss functions.}
\label{tab:ablation}
    \begin{tabular}{|l|c|c|c|c|c|}
    \hline
    Model Variant           & MAE & rMSE & MAE (\%) & $R^2$ & IoU \\
    \hline
    MAE                     & 2.76 & 3.74 & 26.2 & 0.71 & 85.6 \\
    \hline
    pwMAE                   & 2.76 & 3.73 & 25.9 & 0.71 & 86.4 \\
    \hline
    wGDL + pwMAE (Our)      & \textbf{2.70} & \textbf{3.66} & \textbf{25.3} & \textbf{0.72} & 86.7 \\
    \hline
    wGDL + pwMAE $w_\text{non-tree}=0.5$ & 2.90 & 3.93 & 27.4 & 0.67 & \textbf{87.3} \\
    \hline
    \end{tabular}
\end{table*}

\begin{figure}[!htbp]
    \centering
    \includegraphics[width=\linewidth]{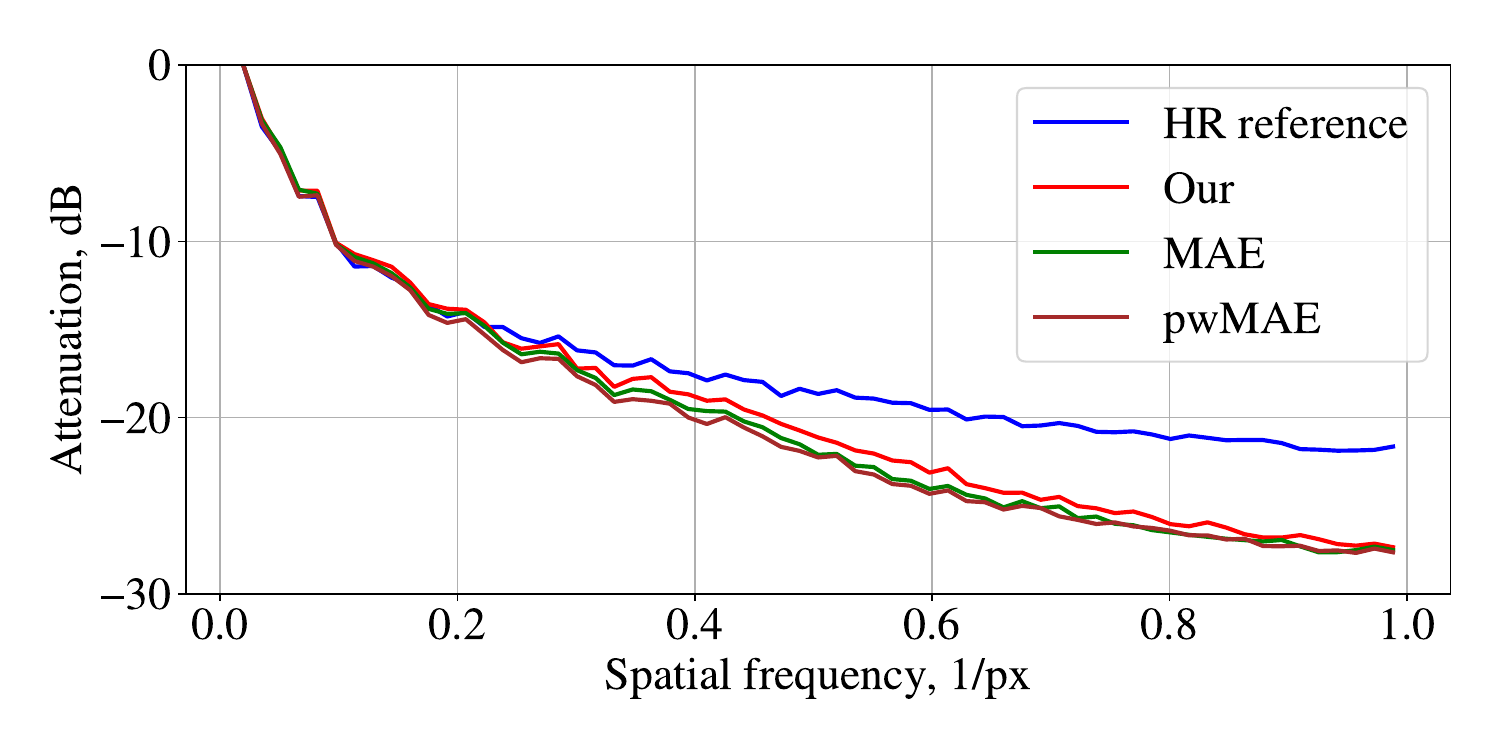}
    \caption{FAP profile for the target image, SR image produced by our model at 5~m resolution with the chosen hyperparameters and two variation of our model: the one using only MAE loss, for training, and the other using only pwMAE for training, without wGDL loss.}
    \label{fig:ablation}
\end{figure}

The quantitative results are presented in Table~\ref{tab:ablation}. We observe a slight improvement of pwMAE over MAE. Moreover, the improvement of our full model compared to the configuration without the wGDL loss is non-negligible. Finally, increasing the weight of non-tree pixels to 0.5 improves the classification mask but reduces regression performance.

Figure~\ref{fig:ablation} presents FAP profiles for configurations (i), (ii), and (iii), where we observe improved frequency reconstruction when using the wGDL loss compared to the variants without it. Those observations prove the efficacy of the proposed model configuration.

\subsection{Comparison with other methods using independent LiDAR reference map}
\label{subsec:comparison}
We evaluate our model at different spatial resolutions and compare it with the three canopy height mapping approaches already cited in the previous sections: FORMS-T~\citep{SCHWARTZ2025114959} at 10\,m resolution, Open-Canopy~\citep{11095116} at 1.5\,m resolution, and the Global Canopy Height (GCH) Model~\citep{lang2023high} at 10\,m resolution. For the competing methods, we use canopy height maps provided directly by the authors of the corresponding models.

The evaluation presented in this subsection is conducted over the Landes region in southwestern France, where we have an independent LiDAR dataset. The reference LiDAR data cover an area of 580~ha. \rev{The site has a mean tree height of $12\pm5$ m.} The evaluation is also conducted for other three areas in other years for the FORMS-T and Open-Canopy maps and presented in \ref{app:extended_comparison}.

Evaluation is performed at the native resolution of each canopy height product by reprojecting and aggregating the reference LiDAR height data to the corresponding grid.

\begin{figure*}[!htbp]
\centering
\begin{subfigure}{0.48\textwidth}
    \centering
    \includegraphics[width=\linewidth]{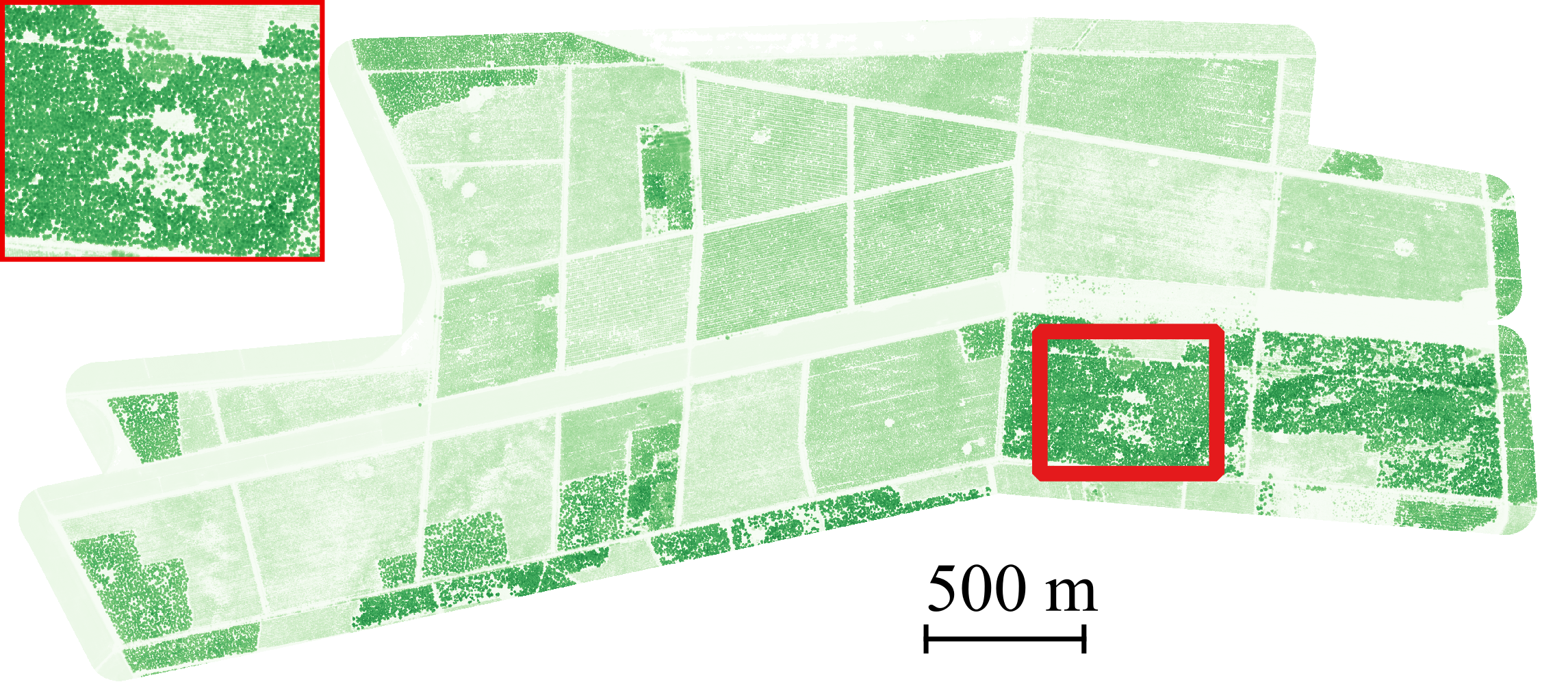}
    \caption{Reference Height, 1~m}
\end{subfigure}
\hfill
\begin{subfigure}{0.48\textwidth}
    \centering
    \includegraphics[width=\linewidth]{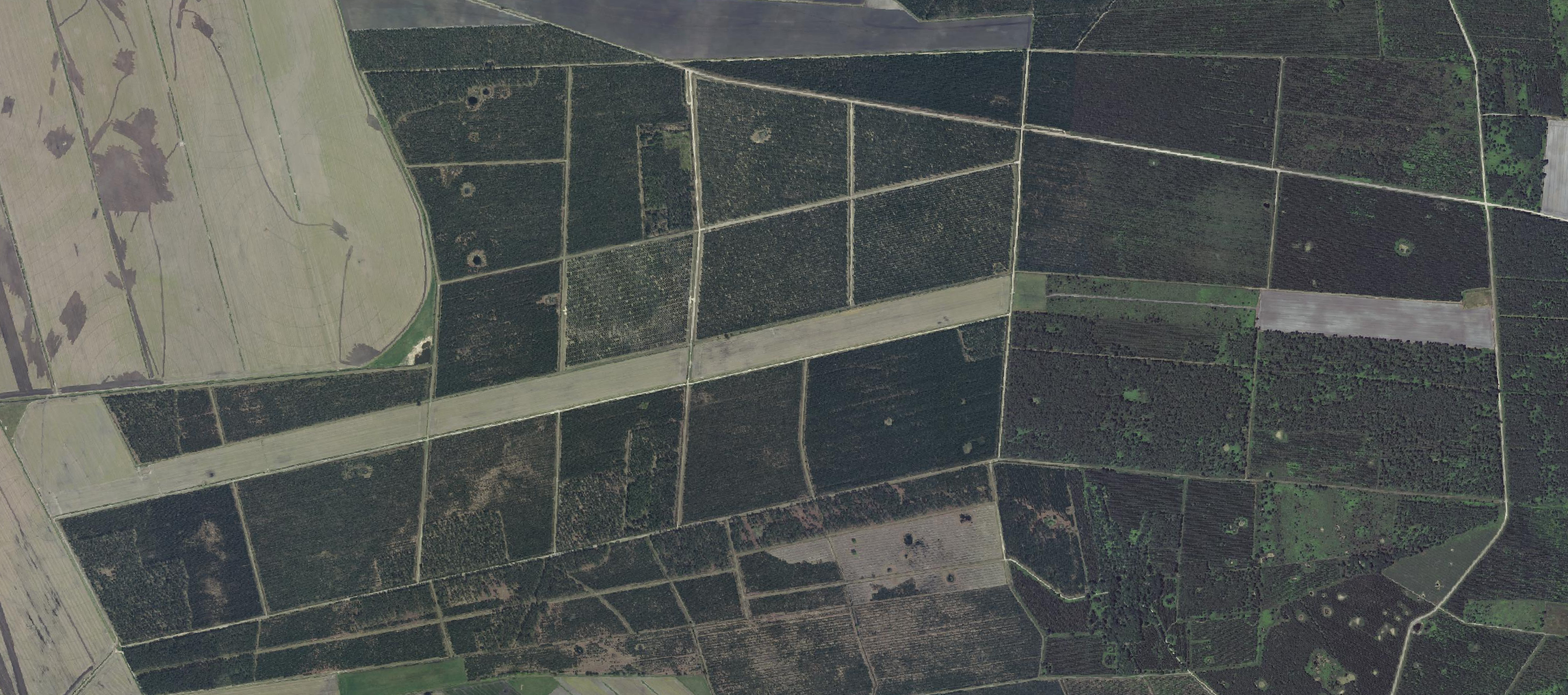}
    \caption{VHR Image}
\end{subfigure}

\vspace{4pt}

\begin{subfigure}{0.48\textwidth}
    \centering
    \includegraphics[width=\linewidth]{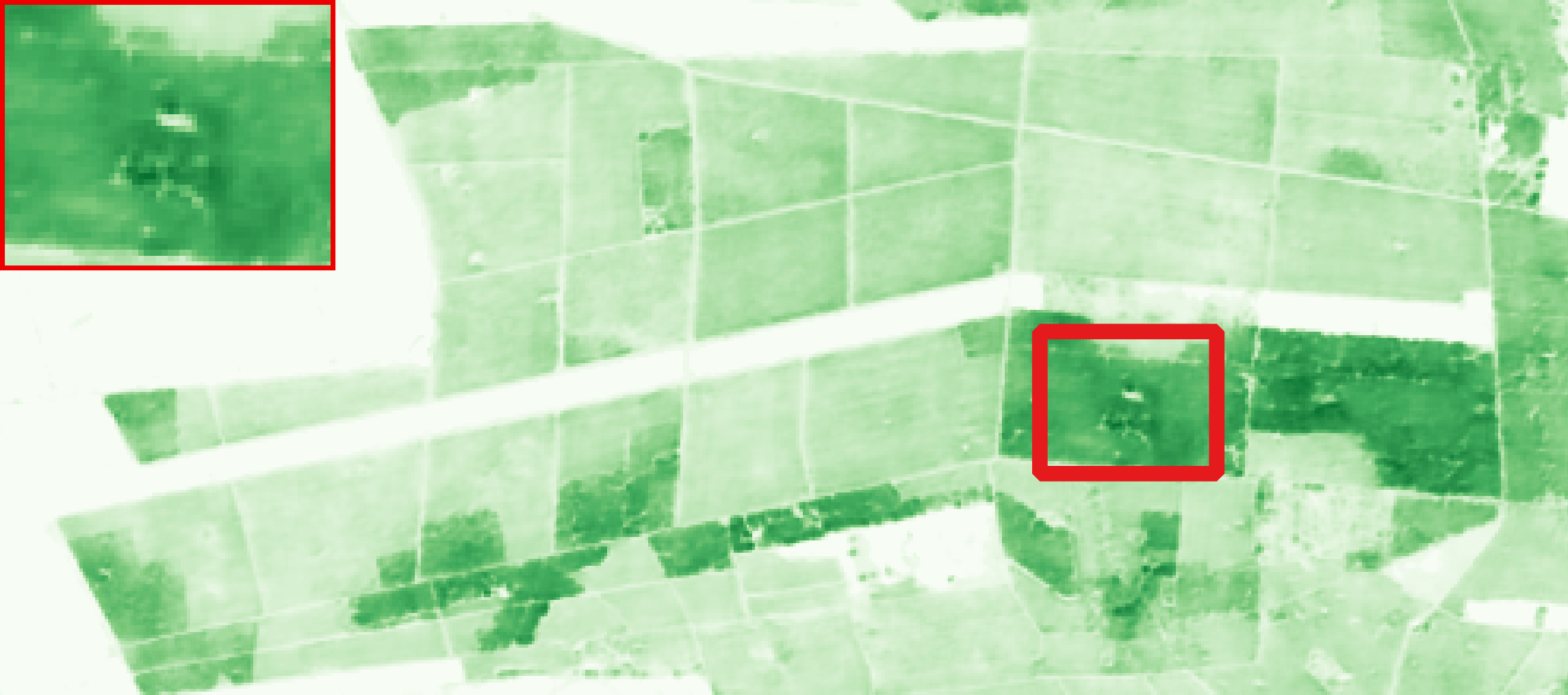}
    \caption{Our, 10~m}
\end{subfigure}
\hfill
\begin{subfigure}{0.48\textwidth}
    \centering
    \includegraphics[width=\linewidth]{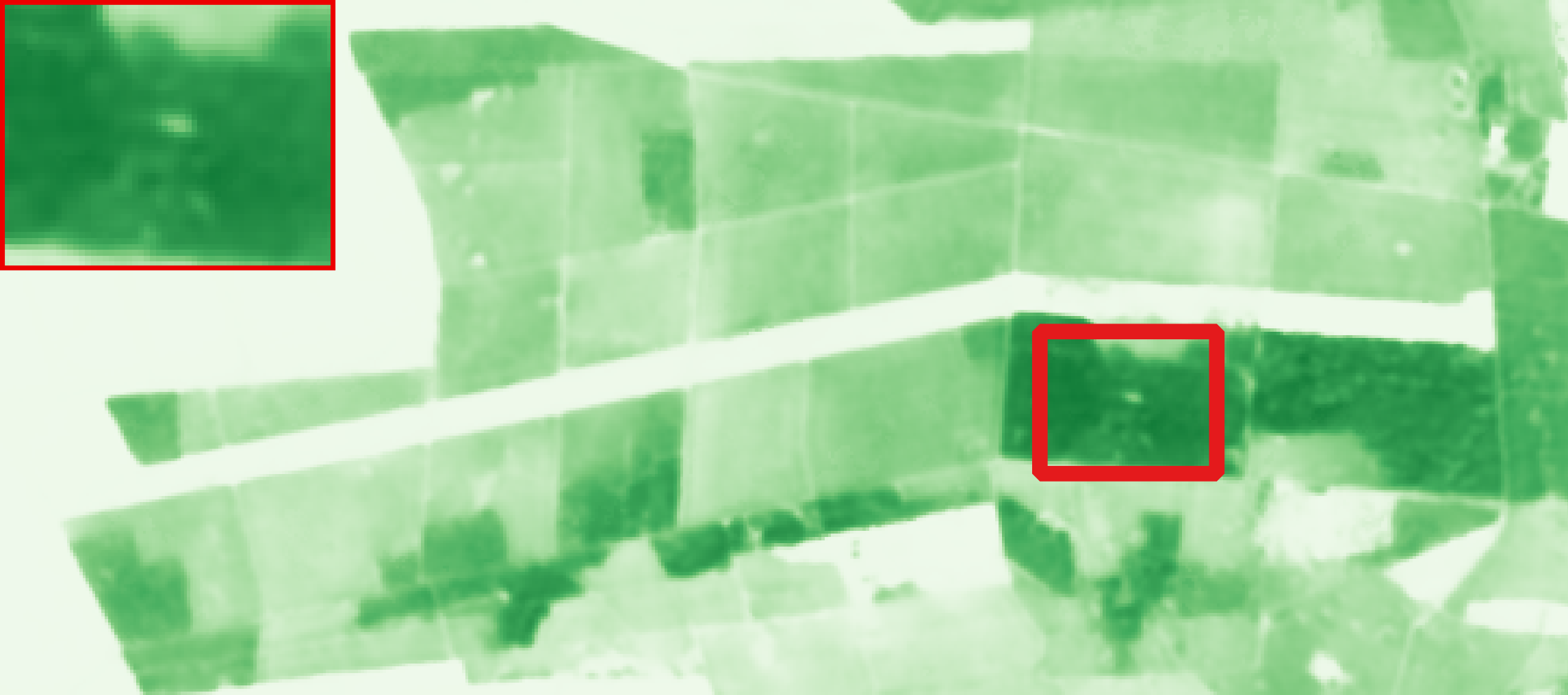}
    \caption{FORMS-T, 10~m}
\end{subfigure}

\vspace{4pt}

\begin{subfigure}{0.48\textwidth}
    \centering
    \includegraphics[width=\linewidth]{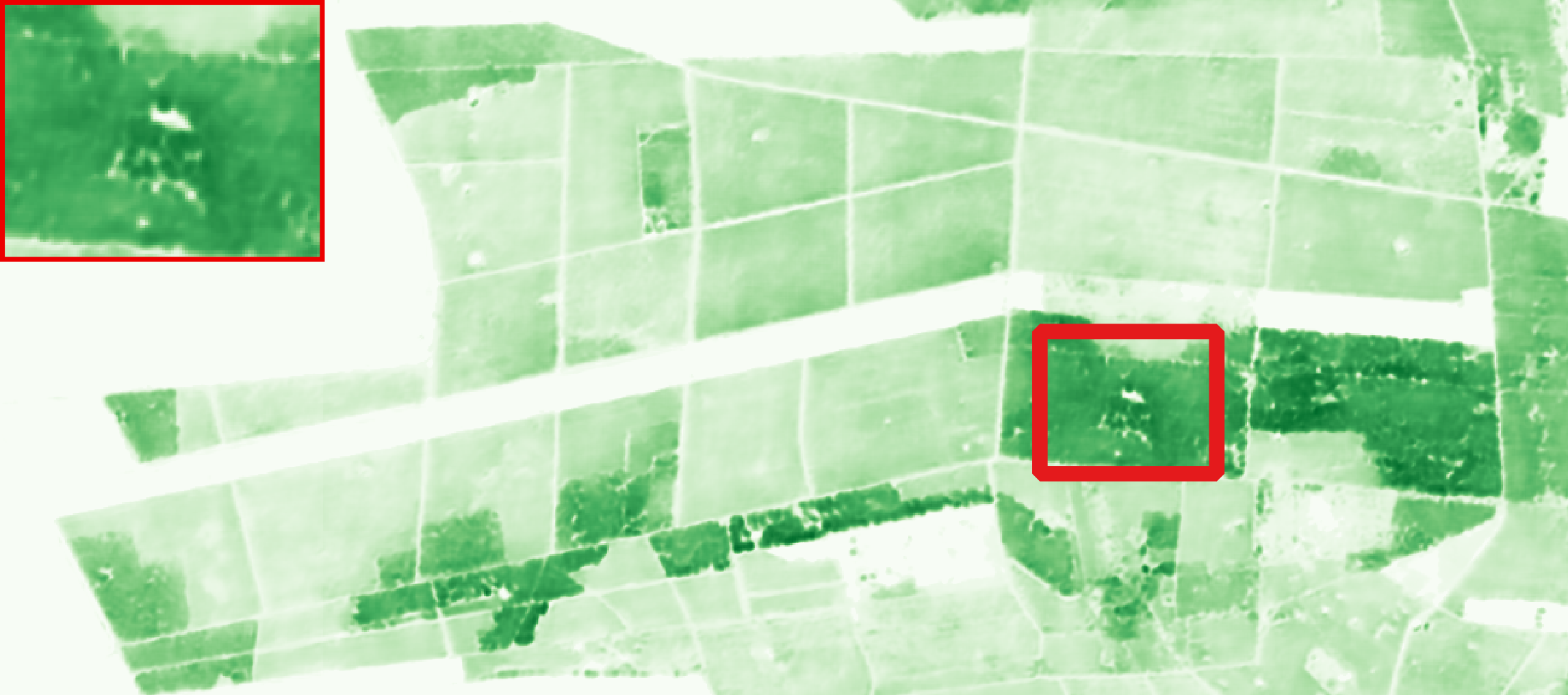}
    \caption{Our, 5~m}
\end{subfigure}
\hfill
\begin{subfigure}{0.48\textwidth}
    \centering
    \includegraphics[width=\linewidth]{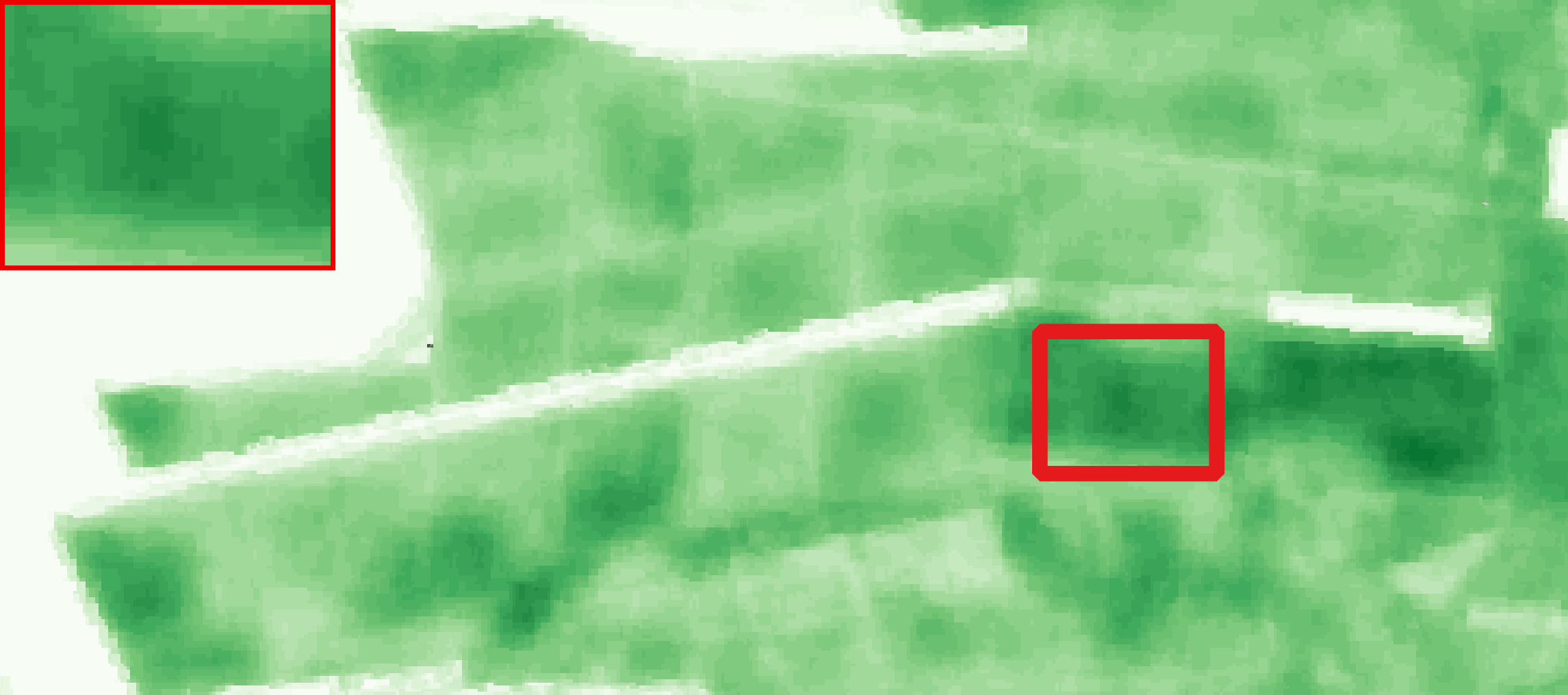}
    \caption{Lang, 10~m}
\end{subfigure}

\vspace{4pt}

\begin{subfigure}{0.48\textwidth}
    \centering
    \includegraphics[width=\linewidth]{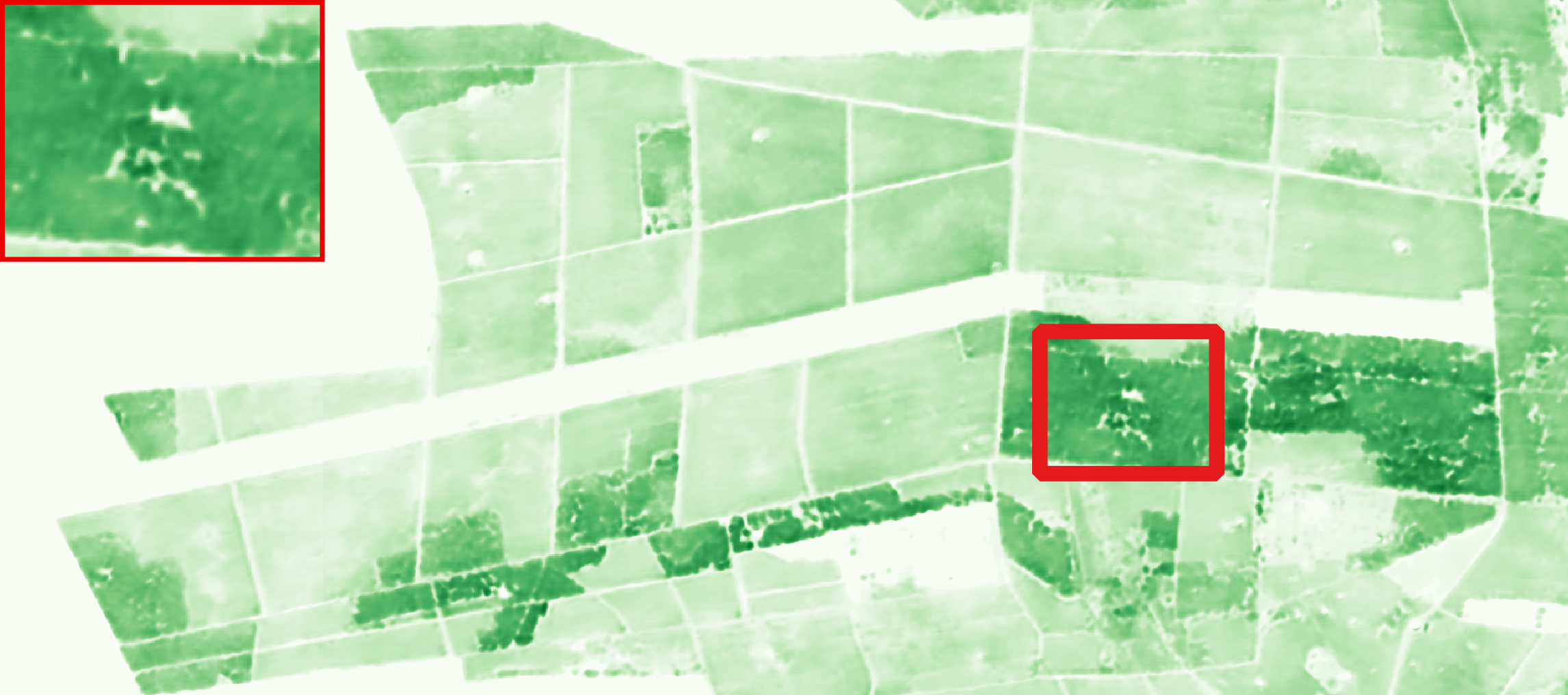}
    \caption{Our, 2.5~m}
\end{subfigure}
\hfill
\begin{subfigure}{0.48\textwidth}
    \centering
    \includegraphics[width=\linewidth]{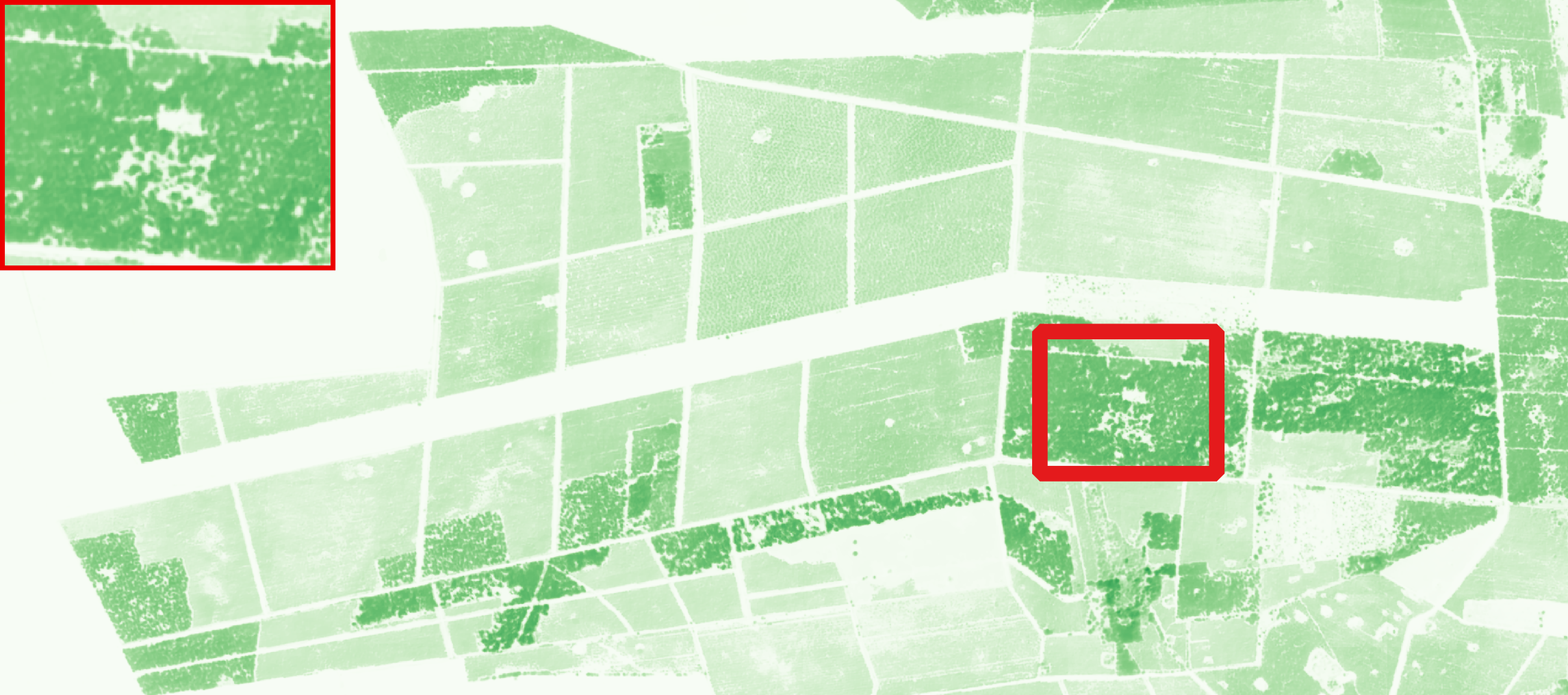}
    \caption{Open-Canopy, 1.5~m}
\end{subfigure}

\vspace{2pt}

\begin{subfigure}{0.25\textwidth}
    \centering
    \includegraphics[width=\linewidth]{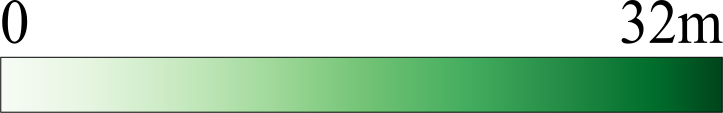}
\end{subfigure}

\caption{Tree height prediction of 580~ha area centered on 45.127°N and 0.901°W. Comparison of a) independent reference ALS-derived heights and our model at three resolutions -- b) 10 m, c) 5 m, and d) 2.5 m -- with concurrent approaches: d) FORMST-T~\citep{SCHWARTZ2025114959}, e) Global Canopy Height Model~\citep{lang2023high}, and f) Open-Canopy~\citep{11095116}. Zoomed-in views (top left) highlight differences in fine details among predictions on the eastern side of the study area.}
\label{fig:landes_all}
\end{figure*}

\begin{table*}[!htbp]
    \centering
    {\color{blue}

    \caption{Quantitative results of tree height estimation for our models of 10, 5 and 2.5~m resolutions on the independent area in South-West France (site Carcans-Hourtins). We also report the performance of three recent state-of-the-art algorithms: FORMS-T~\citep{SCHWARTZ2025114959} at 10~m, Open-Canopy~\citep{11095116} at 1.5~m resolution, and Global Canopy Height Model~\citep{lang2023high} at 10~m resolution.}
    \begin{tabular}{|l|c|c|c|c|c|c|}
    \hline
        Model & Res.,~m & MAE,~m & MAE, \% & $R^2$ & RMSE,~m & IoU, \% \\ \hline
        \multirow{3}{*}{THREASURE-Net}
        & 10
        & 2.41
        & \textbf{24.2}
        & 0.70
        & 3.08
        & \textbf{97.1} \\ \cline{2-7}

        & 5
        & 2.30
        & 27.0
        & 0.71
        & 2.98
        & 94.6 \\ \cline{2-7}

        & 2.5
        & 2.22
        & 30.6
        & 0.69
        & 3.06
        & 92.8 \\ \hhline{|=|=|=|=|=|=|=|}

        FORMS-T
        & 10
        & \textbf{2.16}
        & 28.5
        & \textbf{0.73}
        & \textbf{2.93}
        & 84.5 \\ \hline
        
        Open-Canopy
        & 1.5
        & 2.49
        & 30.4
        & 0.60
        & 3.44
        & 94.0 \\ \hline
        
        GCH
        & 10
        & 3.91
        & 54.7
        & 0.27
        & 4.82
        & 89.3 \\ \hline
       

    \end{tabular}
\label{tab:landes_all}
}
\end{table*}

Figure~\ref{fig:landes_all} shows the qualitative results for the study area, while Table~\ref{tab:landes_all} summarizes the corresponding quantitative results. For our algorithm, for tree/non-tree classification, we directly use predicted binary mask while for other algorithms, the 2~m threshold was applied to the tree height prediction to isolate ``non-tree'' class.
The Global Canopy Height model gave the worst results compared to other models, likely because it was trained with global data, contrary to the other algorithms designed specifically for the French temperate forests. 
This result suggests that region-specific models provide advantages over globally trained models for local-scale studies.
From visual observation, it can be noted that Open-Canopy underestimates high values and flattens spatial textures, providing more homogeneous values within parcels, contrary to the other algorithms. This is reflected in the higher rMSE compared to FORMS-T and our models at 5 and 10 m, confirming that the Open-Canopy model tends to predict values close to the dataset mean, which results in poorer performance for extreme values, as noted earlier.
Moreover, for 10~m resolution, our model visually gives much sharper textures than FORMS-T, which can possibly be explained by the use of wGDL loss.

High rMAE values are observed for our models at 5 and 2.5~m resolution, as well as for the concurrent approaches, despite their low absolute MAE. This can be explained by the prevalence of narrow gaps within the study area -- often smaller than the pixel resolution -- which are represented by low reference values and therefore increase the sensitivity of relative error to small absolute deviations, as the algorithms tend to smooth over these gaps at finer resolutions. As observed, our model at 10~m resolution yields the best rMSE, likely because the number of such gaps decreases at this coarser resolution.


\section{Conclusion}

In this paper, we have introduced THREASURE-Net, a novel end-to-end framework for super-resolved tree canopy height estimation. Our approach generates high-resolution annual predictions at 2.5, 5 and 10~m directly from Sentinel-2 time series with 10~m resolution, supervised with LiDAR~HD–derived canopy height data. THREASURE-Net outperforms existing Sentinel-2–based methods for forest height prediction and achieves performance comparable to approaches relying on 1.5~m SPOT-6/7 imagery. Our model achieves mean absolute errors of 2.63~m, 2.70~m, and 2.89~m at 2.5~m, 5~m, and 10~m resolution, respectively. \rev{The improved resolution of THREASURE-Net shows potential for multi-year tree loss detection and tree growth monitoring tasks. However, both tasks are complex and would require further development and refinement before achieving operational implementation.}

\rev{Among the three evaluated resolutions, the 5~m model appears to provide the best trade-off between tree height prediction accuracy and spatial detail. While the 2.5~m model improves the representation of fine-scale spatial structures, its prediction accuracy is often insufficient, particularly in more challenging sites.}

\rev{Our model achieves competitive or superior performance using Sentinel-2 data alone than state-of-the-art methods that rely on both optical and radar data, thanks to its efficient use of vegetation phenology captured in Sentinel-2 time series. Incorporating radar data could be, however, a promising direction to further improve the prediction of very small and very tall trees, as it provides complementary structural information.  Further work will focus on these improvements as well as the mitigation of the imbalance across canopy height classes. }

\rev{Furthermore, it would be interesting to investigate the model's conditioning on LiDAR acquisition in more detail, with a view to potentially extending the framework towards monthly change detection.
}

\section*{Acknowledgements}
Ekaterina Kalinicheva, Florian Mouret and Milena Planells were supported by the PC ALAMOD from the PEPR FairCarbon and received government funding managed by the Agence Nationale de la Recherche under the France 2030 program. Florian Helen and Stéphane Mermoz were supported by a ``R\&T: Estimation annuelle de paramètres bio-géophysiques des forêts françaises et leurs changements par IA'' grant from the Centre National d'Études Spatiales (CNES) (Marché n°5700013599).

We would like to express our sincere gratitude to Julien Michel and Jordi Inglada for their assistance and valuable advice on deep learning methods, without which this work would not have been possible. We also gratefully acknowledge Jerôme Bock from ONF for providing the independent ALS-derived canopy height data used in this study for forest disturbance detection task.

\appendix
{\color{blue}

\section{Extended comparison with other methods using independent LiDAR reference maps}
\label{app:extended_comparison}

\begin{table*}[!htbp]
{\color{blue}
\centering
\caption{
Quantitative results of tree height estimation for our models of 10, 5 and 2.5~m resolutions on four independent study areas in France. 
We also report the performance of recent state-of-the-art algorithms: 
FORMS-T~\citep{SCHWARTZ2025114959}, 
Open-Canopy~\citep{11095116}, 
and Global Canopy Height Model~\citep{lang2023high} (only for Carcans-Hourtins site).
}
\begin{tabular}{cccccccc}
\hline
\begin{tabular}{c}
\textbf{Site}\\\textbf{(Tree Height, m $\pm$ STD)}
\end{tabular}
& \textbf{Model}
& \textbf{Res.,~m}
& \textbf{MAE,~m}
& \textbf{MAE, \%}
& \textbf{$R^2$}
& \textbf{RMSE,~m}
& \textbf{IoU, \%} \\
\hline

\multirow{6}{*}{
\begin{tabular}{c}
Carcans-Hourtins \\
(12 $\pm$ 5 m)
\end{tabular}
}
& \multirow{3}{*}{THREASURE-Net}
& 10
& 2.41
& \textbf{24.2}
& 0.70
& 3.08
& \textbf{97.1} \\

&
& 5
& 2.30
& 27.0
& 0.71
& 2.98
& 94.6 \\

&
& 2.5
& 2.22
& 30.6
& 0.69
& 3.06
& 92.8 \\
\cline{2-8}

& FORMS-T
& 10
& \textbf{2.16}
& 28.5
& \textbf{0.73}
& \textbf{2.93}
& 84.5 \\

& Open-Canopy
& 1.5
& 2.49
& 30.4
& 0.60
& 3.44
& 94.0 \\

& GCH
& 10
& 3.91
& 54.7
& 0.27
& 4.82
& 89.3 \\
\hline

\multirow{5}{*}{
\begin{tabular}{c}
Déodatie\\(19 $\pm$ 10 m)
\end{tabular}
}
& \multirow{3}{*}{THREASURE-Net}
& 10
& 4.04
& \textbf{27.9}
& 0.73
& 5.34
& 84.3 \\

&
& 5
& 4.13
& 33.7
& 0.72
& 5.38
& 90.3 \\

&
& 2.5
& 4.65
& 38.7
& 0.63
& 6.03
& 87.2 \\
\cline{2-8}

& FORMS-T
& 10
& 4.05
& 30.6
& \textbf{0.72}
& 5.39
& \textbf{95.3} \\

& Open-Canopy
& 1.5
& \textbf{3.77}
& 28.1
& 0.74
& \textbf{5.01}
& 90.5 \\
\hline

\multirow{5}{*}{
\begin{tabular}{c}
Lajoux-Fresse\\(22 $\pm$ 9 m)
\end{tabular}
}
& \multirow{3}{*}{THREASURE-Net}
& 10
& 4.10
& \textbf{21.3}
& 0.65
& 5.35
& 94.4 \\

&
& 5
& 4.50
& 28.5
& 0.59
& 5.92
& \textbf{94.6} \\

&
& 2.5
& 5.33
& 38.2
& 0.46
& 6.83
& 93.5 \\
\cline{2-8}

& FORMS-T
& 10
& \textbf{3.96}
& 22.0
& \textbf{0.66}
& \textbf{5.32}
& 93.7 \\

& Open-Canopy
& 1.5
& 4.26
& 33.9
& 0.61
& 5.78
& 92.5 \\
\hline

\multirow{5}{*}{
\begin{tabular}{c}
Mouterhouse\\(22 $\pm$ 8 m)\end{tabular}
}
& \multirow{3}{*}{THREASURE-Net}
& 10
& 5.79
& 29.5
& 0.17
& 7.14
& 95.4 \\

&
& 5
& 5.54
& 31.2
& 0.23
& 6.87
& \textbf{95.7} \\

&
& 2.5
& 6.38
& 35.3
& -0.02
& 7.84
& 93.7 \\
\cline{2-8}

& FORMS-T
& 10
& \textbf{3.84}
& \textbf{25.5}
& \textbf{0.61}
& \textbf{5.02}
& 94.1 \\

& Open-Canopy
& 1.5
& 5.11
& 28.4
& 0.34
& 6.33
& 92.3 \\
\hline

\end{tabular}

\label{tab:all_sites_quantitative}
}
\end{table*}

In this section, we present extended results from the comparison with independent ALS-derived tree height maps at 1~m resolution. The evaluation is conducted over four study areas: Carcans-Hourtins, Déodatie, Lajoux-Fresse and Mouterhouse (see Table~\ref{tab:data_all} for site descriptions). The maps are provided by the ONF (Office National de Forêts) corresponding to one acquisition year. The four study sites represent different forest compositions and topographic contexts. Carcans-Hourtins, located in the Landes region, is characterized by flat terrain and a maritime pine (\textit{Pinus pinaster}) dominated stand, with ONF ALS acquisition in October 2020. Mouterhouse is also located in a predominantly flat area, although it is characterized by numerous small forested hills with short but sometimes very steep slopes (over 50\%) The site consists of a mixed stand dominated by oak, hornbeam, beech, and pine. The ONF ALS data were acquired in February 2019. The Deodatie site is located in a hilly landscape with a mixed forest composition of deciduous and conifers, with ONF ALS data acquired in April 2018. Finally, Lajoux-Fresse is situated in a mountainous environment, a forest dominated by spruce, Norway spruce, and beech, with ONF ALS acquisition in June 2019. For each study area, we randomly sampled a test zone of 580, 960, 520 and 520~ha respectively.

We report the performance of our method at three spatial resolutions (10, 5, and 2.5~m), and compare it against Open-Canopy~\citep{11095116} at 1.5~m resolution and FORMS-T~\citep{SCHWARTZ2025114959} at 10~m resolution. For the Carcans-Hourtins site, we additionally include results from the Global Canopy Height Model~\citep{lang2023high}. To ensure a consistent comparison across resolutions, all reference ALS height maps are resampled to match the spatial resolution of the corresponding predictions.

The quantitative results are summarized in Table~\ref{tab:all_sites_quantitative}. Overall, for Déodatie, Lajoux-Fresse, and Carcans-Hourtins, our method at 10~m resolution performs comparably to FORMS-T, while achieving consistently lower relative MAE, despite higher MAE, indicating improved robustness with respect to site-level height variability.

We observe that performance of the 2.5~m THREASURE-Net model decreases with increasing mean canopy height across sites, a trend already reported in Figure~\ref{fig:error_bins}. This effect is likely related to an imbalance in the training distribution and could be mitigated through more balanced sampling strategies, which we leave for future work.

The 5~m model shows performance comparable to the 10~m model, suggesting that 5~m resolution provides a good trade-off between spatial detail and prediction stability. In contrast, Open-Canopy does not consistently achieve the best performance across sites, despite being trained on VHR data.

Overall, our method achieves the strongest performance in tree/non-tree classification, as it is explicitly trained for this task across the entire French territory, rather than being restricted to forest-only regions as in other approaches. 

Finally, the Mouterhouse site presents a notable performance drop for all our models and Open-Canopy. This can be attributed to several factors. First, the site is located in a hilly area with predominantly tall forest stands, where strong topographic shading in Sentinel-2 time series may degrade the spectral signal. Second, the ALS acquisition was conducted at a higher incidence angle compared to other sites (45° versus approximately 40°), which may introduce additional geometric inconsistencies. In this challenging setting, Open-Canopy also shows reduced performance, likely due to its reliance on optical data affected by shadowing. In contrast, FORMS-T achieves the best results, potentially due to its use of radar data, which is not affected by shadowing and atmospheric conditions. Future work could explore the integration of radar data into our framework to improve robustness in such challenging environments.

Beyond the methodological considerations inherent to deep learning, the methods compared in this study differ in terms of input data (optical only vs. optical and SAR), input spatial resolution (10 m Sentinel-2 vs. 1.5 m SPOT-6/7), and geographic scope of training (national vs. global). THREASURE-Net is specifically designed to maximize the use of freely available (globally consistent Sentinel-2 time series), and its results demonstrate that this choice does not fundamentally limit predictive accuracy for the temperate forest types considered here, while offering clear advantages in terms of accessibility and operational scalability.

The competitive performance of THREASURE-Net (relying solely on Sentinel-2 imagery) against FORMS-T (based on Sentinel-1 and -2 imagery), can be attributed to the rich phenological information captured in dense annual time series. Seasonal variations in canopy reflectance serve as an indirect proxy for vertical forest structure. These temporal variations partially compensate for the absence of radar-derived structural information, particularly in temperate forests where phenological cycles are well-marked. In addition, a Sentinel-2-only approach offers a considerably simpler and more reproducible workflow, which is advantageous for large-scale or near-real-time applications.

Nevertheless, the complementary nature of SAR data should not be overlooked. Sentinel-1 backscatter is sensitive to volume scattering and canopy roughness, providing structural information whatever the meteorological conditions. This is reflected in our results, where FORMS-T outperforms our approach for example on Mouterhouse site characterized by tall trees, where SAR backscatter remains more sensitive to above-ground biomass than optical data. Indeed, several recent state-of-the-art approaches~\citep{rs16213992, rs17091536, CHEN2025104814} exploit the complementary nature of optical and SAR observations to improve canopy height estimation. Incorporating Sentinel-1 data into THREASURE-Net is therefore a promising direction for future work, particularly to improve predictions for tall and structurally complex forest stands.

}

\section{Sylvoécorégions of France}
\label{apx:SER}
\begin{figure}[h]
    \centering
    \includegraphics[width=\linewidth]{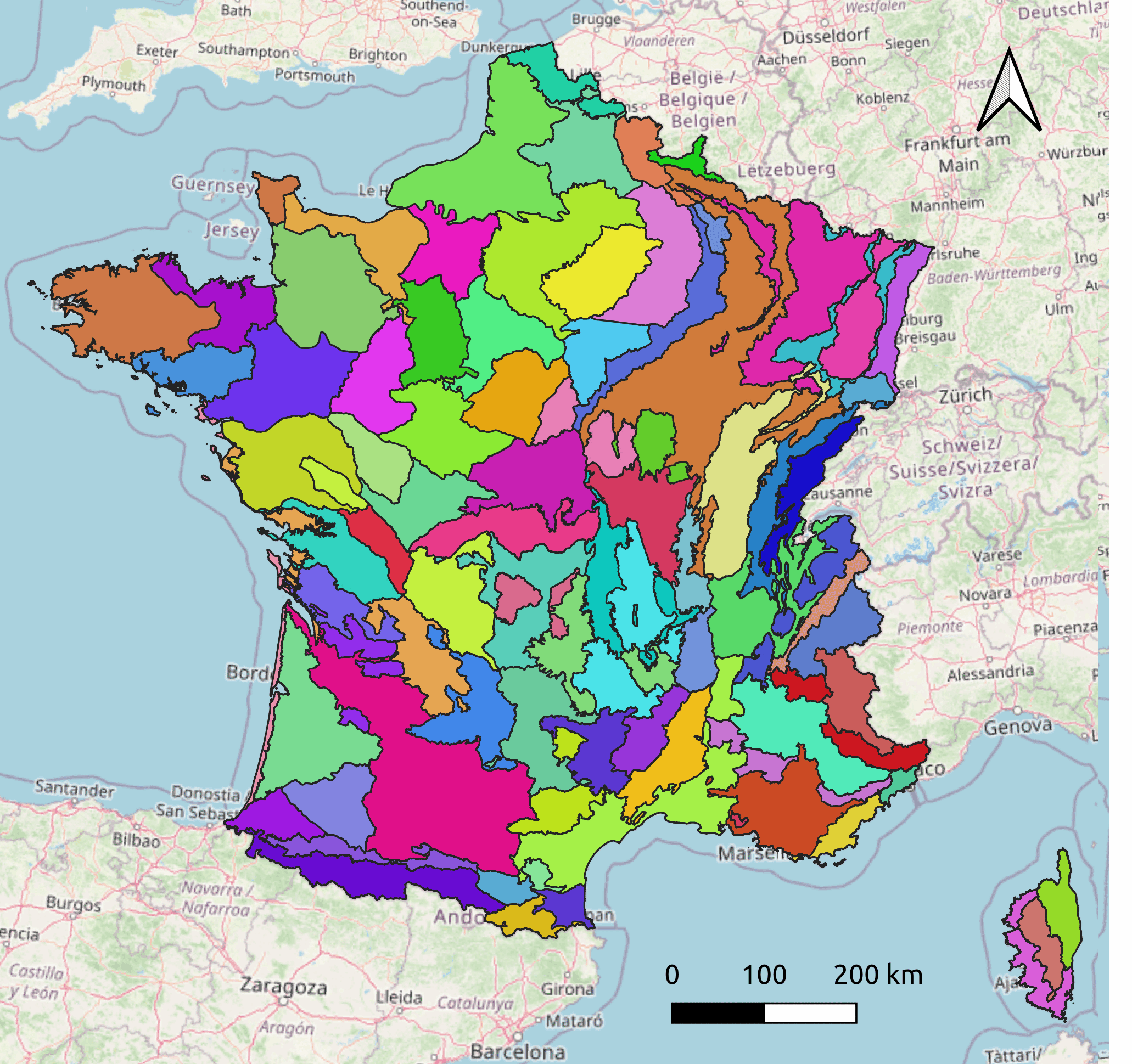}
    \caption{Sylvoecoregions of France, illustrating the ecological diversity of forest types across the country.}
    \label{fig:SER}
\end{figure}

 \rev{In mainland France, the National Forest Inventory defines $91$ sylvoecoregions (SERs), which are spatial units that are homogeneous in terms of the ecological factors influencing forest production and the distribution of forest habitats. Situated at an intermediate scale between forest massifs and major biomes, they provide a framework for the study and management of French forests. The SERs are based primarily on bioclimatic and pedological criteria. Their delineation is based on the analysis of numerous abiotic factors such as altitude, soil properties (texture, pH, hydromorphism, depth, bedrock, water availability) and climatic characteristics (temperatures, precipitation, water balance). The SER classification provides a coherent framework for describing the distribution and productivity of forest habitats across the country. It is widely used in forest management, ecological monitoring, and biodiversity assessments because it reflects the environmental gradients that shape tree species composition and forest dynamics. By grouping territories with similar ecological characteristics, sylvoecoregions also facilitate comparisons of forest functioning and responses to environmental change. The sylvoecoregions of France are shown in Figure~\ref{fig:SER}.}

\section{Tree growth evaluation}
\label{apx:tree_regrowth}

We evaluate our algorithm on the tree regrowth analysis task. For this task, we choose an entire Sentinel-2 tile 30TXQ corresponding to the Landes of Gascogne in the South-West of France, mostly populated by \textit{Pinus Pinaster}. First, we produce time series of tree height maps and associated binary tree classification masks with THREASURE-Net for 2018-2024 years for three spatial resolutions -- 10, 5, and 2.5~m. We analyze tree classification masks in a temporal context to determine the test area by selecting pixels that were classified as tree in at least three timestamps.

Second, we apply the Mann–Kendall (MK) test~\cite{mann1945mann} to the resulting time series. The Mann–Kendall test is a non-parametric statistical test used to detect monotonic trends in time series data. The test statistic varies between -1 (strong decrease) and +1 (strong increase). The MK test is applied at the parcel level using cadastral boundaries combined with the vegetation mask. Statistically insignificant results ($p_{value} \geq 0.05$) are discarded. The remaining values are grouped into three trend intervals: decreasing $[-1, -0.5)$, stable $[-0.5, 0.5)$, and increasing $[0.5, 1]$.

Within the tree classification mask, \rev{37.5\%} of pixels are statistically insignificant. This proportion is expected given the relatively short length of the time series, which may be insufficient to capture annual growth trends, especially when the mean absolute error (MAE) exceeds the typical yearly regrowth signal. These pixels may correspond to disturbed areas, height estimates with higher uncertainty, or non-vegetated surfaces.

Furthermore, the Mann–Kendall index has limitations when interpreting forest disturbances occurring at the beginning or in the middle of the time series, as such events do not necessarily produce a monotonic trend. In practice, only disturbances occurring toward the end of the time series tend to be reflected by low index values.

For the remaining statistically significant pixels, the results are consistent across all spatial resolutions. Approximately \rev{8\%} of pixels exhibit low Mann–Kendall values, while about \rev{92\%} show high values, indicating increasing trends. In contrast, the proportion of medium trends $[-0.5,0.5)$ is extremely small ($< 0.001\%$).

The near absence of medium trends suggests that almost no parcels exhibit a constant monotonic signal throughout the time series. This behavior is expected for forest vegetation, where growth dynamics are typically non-linear, and it also supports the quality of the tree classification mask.

\begin{figure*}[!htbp]
\centering
\begin{subfigure}[b]{0.32\textwidth}
    \includegraphics[width=\linewidth]{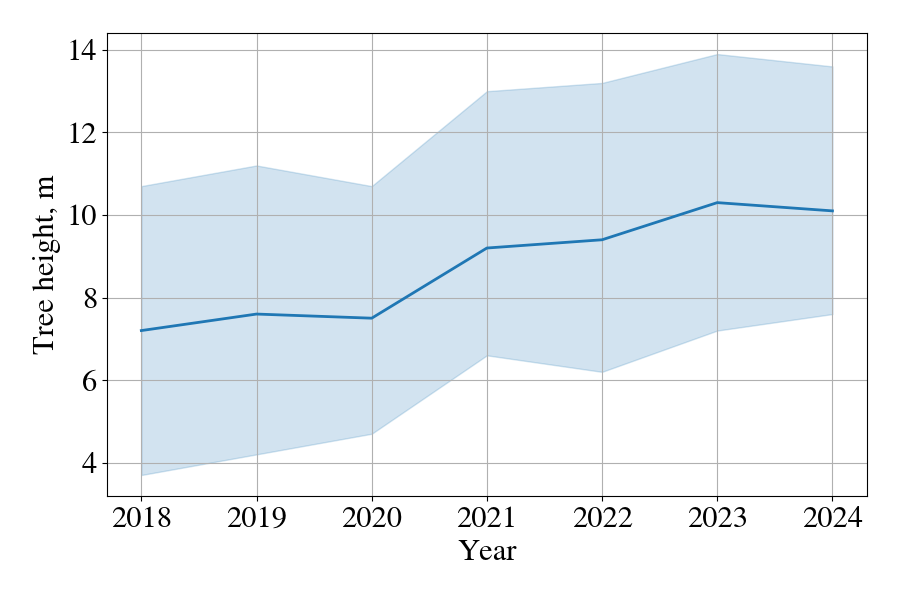}
    \caption{2.5 m}
\end{subfigure}
\hspace{0.01\textwidth}
\begin{subfigure}[b]{0.32\textwidth}
    \includegraphics[width=\linewidth]{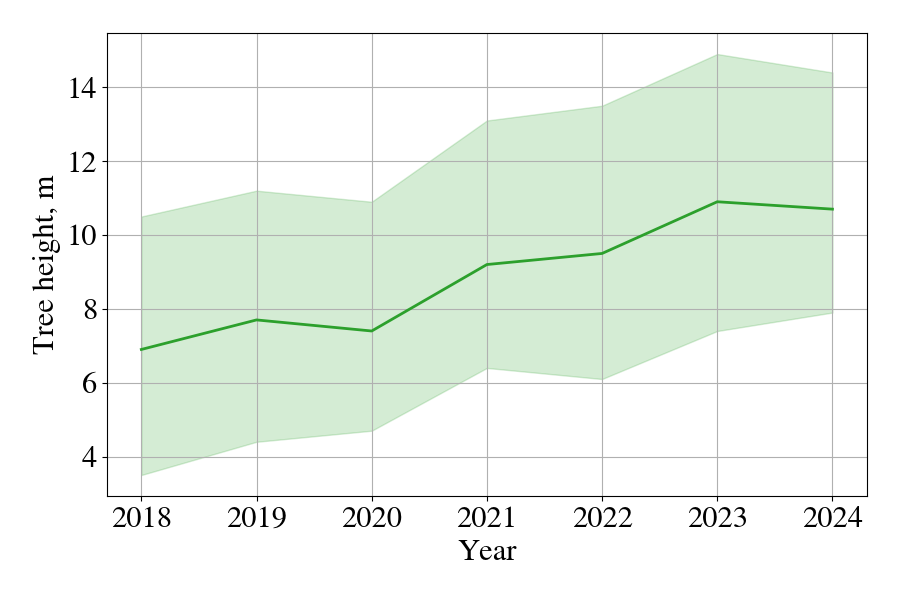}
    \caption{5 m}
\end{subfigure}
\hspace{0.01\textwidth}
\begin{subfigure}[b]{0.32\textwidth}
    \includegraphics[width=\linewidth]{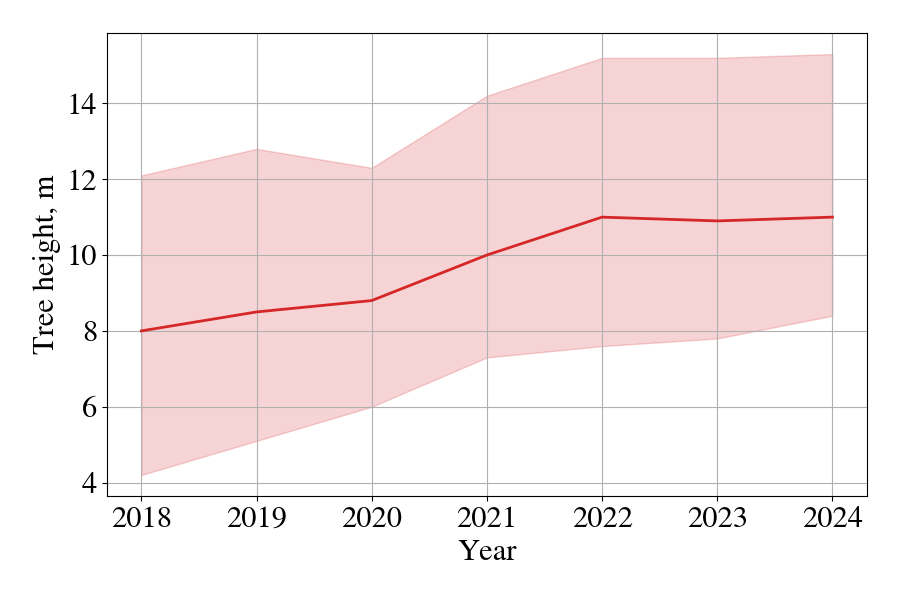}
    \caption{10 m}
\end{subfigure}

\caption{Tree height dynamics for Mann–Kendall high trend class for three spatial resolutions (2.5 m, 5 m, and 10 m). The curves show median tree height evolution over time with interquartile ranges.}
\label{fig:mk_results}
\end{figure*}

Figure~\ref{fig:mk_results} shows the tree growth of parcels exhibiting high positive trends across the three spatial resolutions. Values are reported as the median, along with the 25\textsuperscript{th} and 75\textsuperscript{th} percentiles of statistically significant pixels with high trends.
Interestingly, the years showing weaker or negative growth in the time series coincide with extreme climatic events, which may influence vegetation physiology and leaf composition, thereby affecting canopy reflectance properties that are directly related to height estimation. In particular, 2022 was marked by a severe drought in France, while 2024 was reported as one of the hottest summers recorded in the country since 1900.
Although the variations observed in the predicted heights remain within the expected error margins of the model, these temporary decreases may still reflect the impact of extreme climatic conditions on vegetation growth. Moreover, while the 10~m curve looks smoother, the 2.5 and 5~m curves exhibit identical behavior. Finally, the predicted growth curves align with the expected growth dynamics of young \textit{Pinus pinaster} in the Landes forest~\citep{Pinus}, which reaches approximately 1~m per year, depending on the climatic conditions.

Consequently, deviations from the presented expected growth dynamics may reveal disturbance events, motivating the analysis presented in the following subsection.

\section{Disturbance detection}
\label{apx:change}

We evaluate the robustness of our algorithm on the disturbance detection task. 
Unlike \citep{11095116}, who uses a fixed threshold of 15~m to detect tree loss, we employ an adaptive thresholding approach based on Otsu's method~\citep{otsu1979threshold} to perform binary tree loss classification. Otsu's method -- based on histogram analysis -- is commonly used in remote sensing and image analysis to automatically determine a data-driven threshold separating changed and unchanged areas~\citep{otsu_change}. This choice accounts for the fact that canopy changes can occur at more subtle levels than a fixed threshold would capture. For each site, a pixel-wise height difference map ($\Delta H$) is computed between two acquisitions. To reduce noise and limit the influence of single-pixel outliers, a median filter with a $3\times 3$ window is applied. Only negative $\Delta H$ values are considered for threshold computation to obtain disturbance maps.

For postprocessing, as in \citep{11095116}, we applied erosion and dilation with a 3-pixel kernel to regularize the binary masks, and removed connected components smaller than 200~m². This approach enabled robust detection of tree loss while preserving canopy structure and minimizing the influence of noise in the LiDAR data.

Note that tree loss detection is not the primary objective of the proposed algorithm. The disturbance detection workflow presented here is not intended as a final end-to-end solution; in operational settings, more advanced approaches that go beyond simple height differencing are typically required to achieve robust tree loss detection.

In this work, the threshold is computed independently for each site and for each height difference map. This strategy allows the method to be applied consistently across different areas while still producing stable results. In practical applications, some parameters may need to be adapted, such as the spatial extent over which the threshold is estimated.

Nevertheless, we consider this approach more robust than the use of a fixed threshold, as the adaptive Otsu threshold is computed separately for each product and therefore accounts for the specific characteristics of the predicted map. For instance, if a product systematically underestimates height changes compared to the reference data, the Otsu threshold will adjust accordingly.


We evaluate the proposed method on four different study areas: Mouterhouse,  Deodatie, Lajoux-Fresse, and Carcans-Hourtins. For each site, an independent ALS-based tree height maps at 1~m resolution are provided by the ONF corresponding to one acquisition year. We match those areas with available LiDAR~HD tiles to constitute an image pair for disturbance detection.
The four study sites represent different forest compositions and topographic contexts. Carcans-Hourtins, located in the Landes region, is characterized by flat terrain and a maritime pine (\textit{Pinus pinaster}) dominated stand, with ALS acquisition in October 2020 and LiDAR~HD in September 2023. Mouterhouse is also located in a predominantly flat area, although characterized by numerous small forested hills. The ONF ALS data were acquired in February 2019 and the LiDAR~HD data in February 2022. The site consists of a mixed stand dominated by oak, hornbeam, beech and pine. The Deodatie site is located in a hilly landscape with a mixed forest composition, with ONF ALS data acquired in April 2018 and LiDAR~HD in March 2022. Finally, Lajoux-Fresse is situated in a mountainous environment, with ONF ALS acquisition in June 2019  and LiDAR~HD in August 2022, and a forest dominated by spruce, Norway spruce, and beech. These sites therefore provide a range of structural conditions, forest types, and topographic settings for evaluating disturbance detection performance.

We evaluate the performance of tree loss detection algorithm based on THREASURE-Net predictions at three different spatial resolutions -- 10, 5, and 2.5~m, as well as Open-Canopy~\citep{11095116} at 1.5~m resolution and FORMS-T~\citep{SCHWARTZ2025114959} at 10~m resolution. For each resolution, the reference ALS height maps are resampled to match the resolution of the prediction in order to ensure a consistent comparison.

\begin{landscape}

\begin{figure}[p]

\centering

\setlength{\tabcolsep}{2pt}
\captionsetup[subfigure]{font=tiny,labelfont=tiny,justification=centering}
\setlength{\tabcolsep}{0.2pt}

\begin{adjustbox}{width=\linewidth}
\centering
\begin{tabular*}{\textwidth}{@{\extracolsep{\fill}}ccccccccccccccc}

\multicolumn{15}{c}{\textbf{Carcans-Hourtins}} \\[-2pt]
\setcounter{subfigure}{0}
\begin{subfigure}{.067\textwidth}\includegraphics[width=\linewidth]{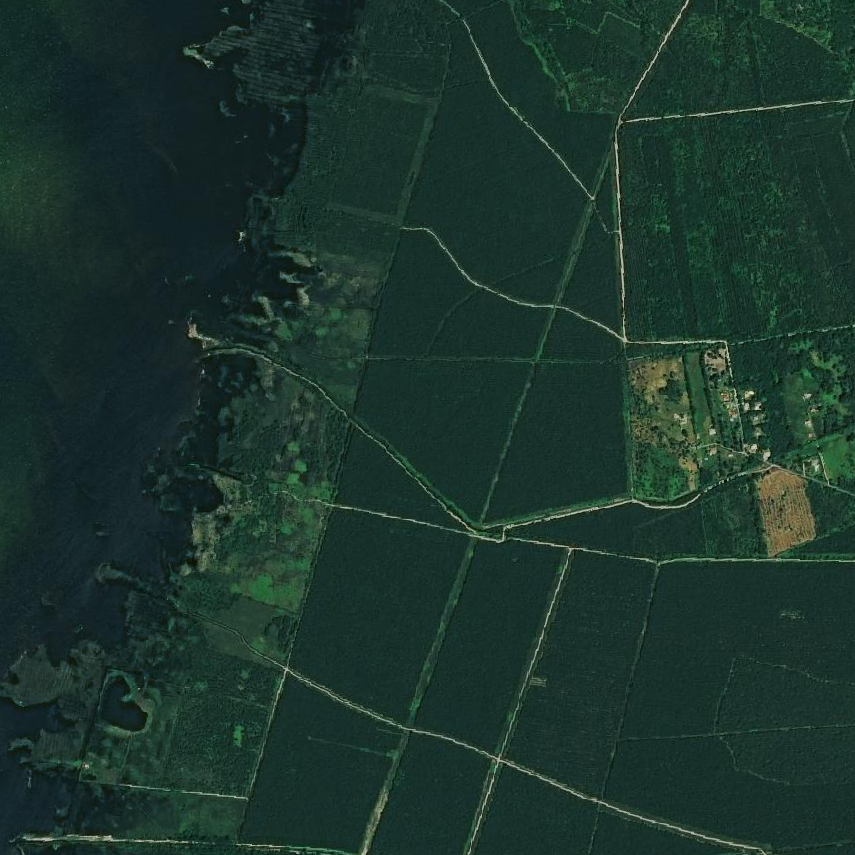}\caption{SPOT, 2019}\end{subfigure} &
\begin{subfigure}{.067\textwidth}\includegraphics[width=\linewidth]{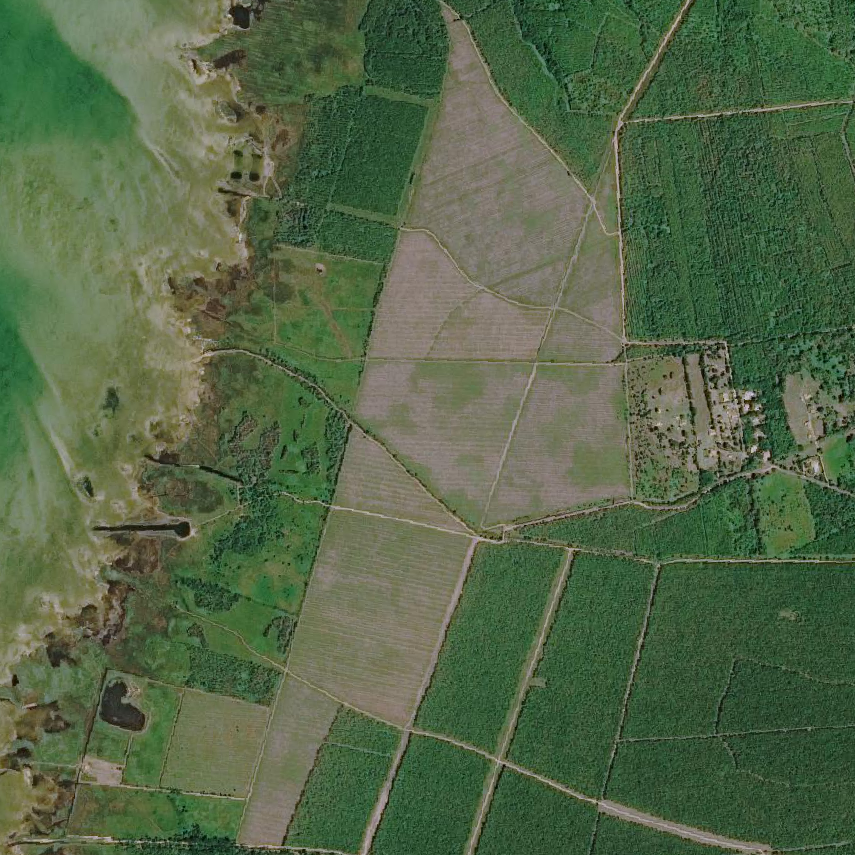}\caption{SPOT, 2023}\end{subfigure} &
\begin{subfigure}{.067\textwidth}\includegraphics[width=\linewidth]{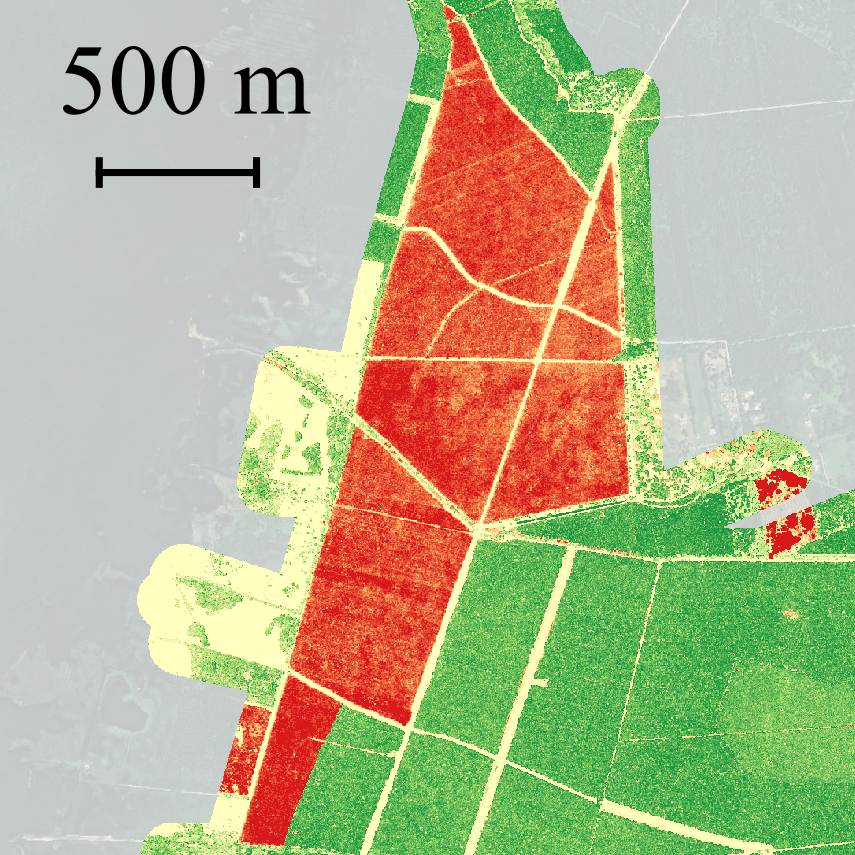}\caption{HDM ALS}\end{subfigure} &
\begin{subfigure}{.067\textwidth}\includegraphics[width=\linewidth]{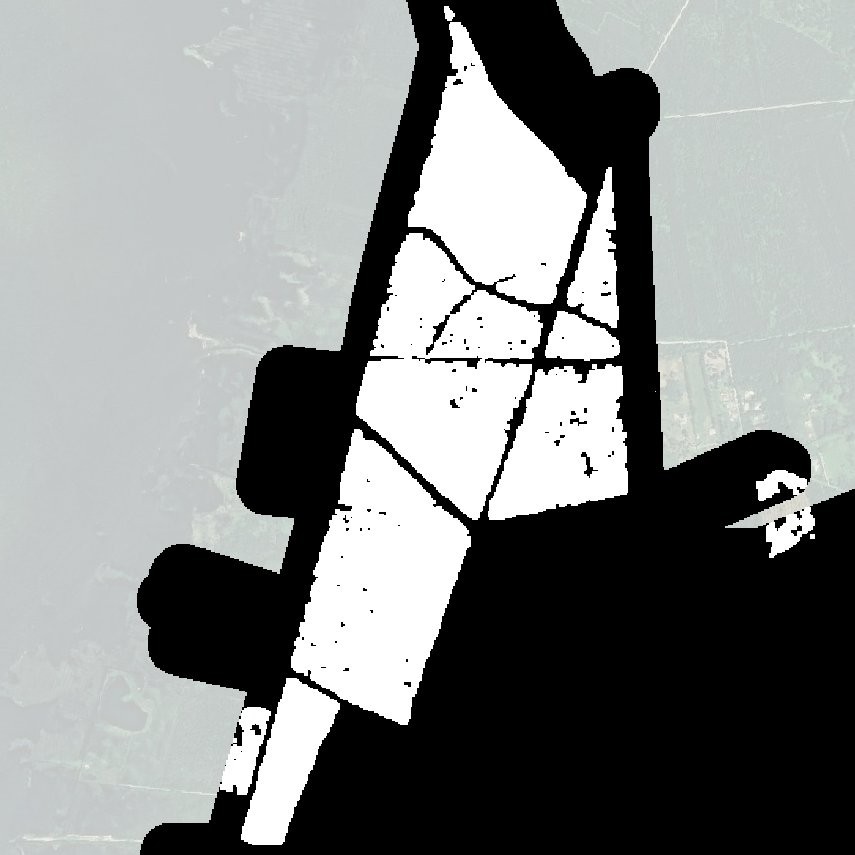}\caption{CM ALS}\end{subfigure} &
\begin{subfigure}{.067\textwidth}\includegraphics[width=\linewidth]{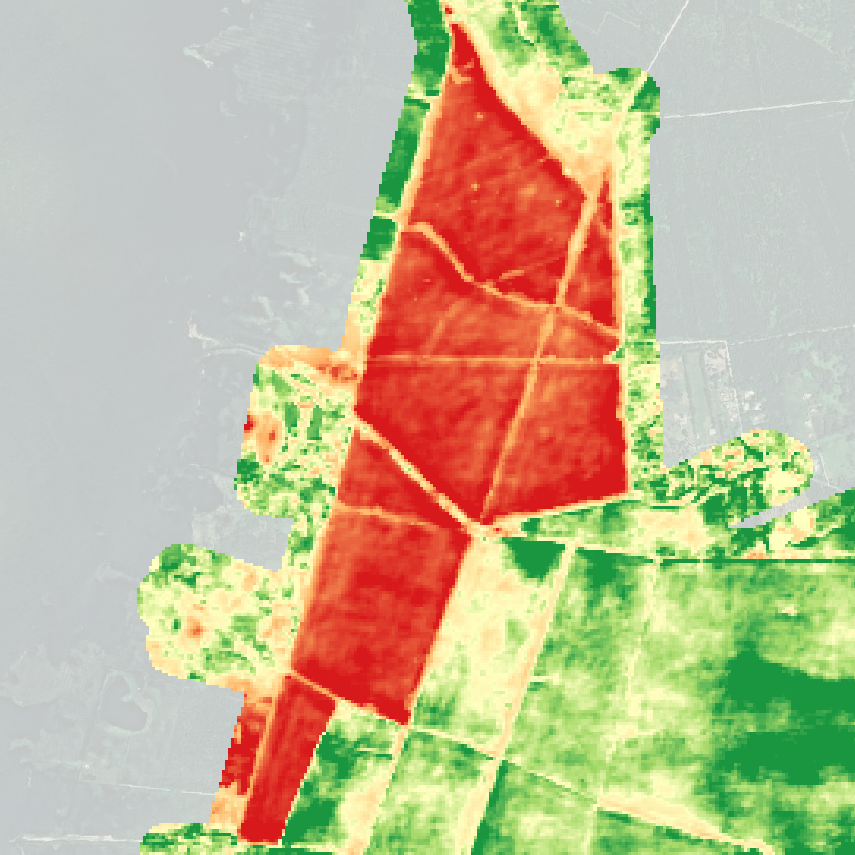}\caption{Ours 10~m}\end{subfigure} &
\begin{subfigure}{.067\textwidth}\includegraphics[width=\linewidth]{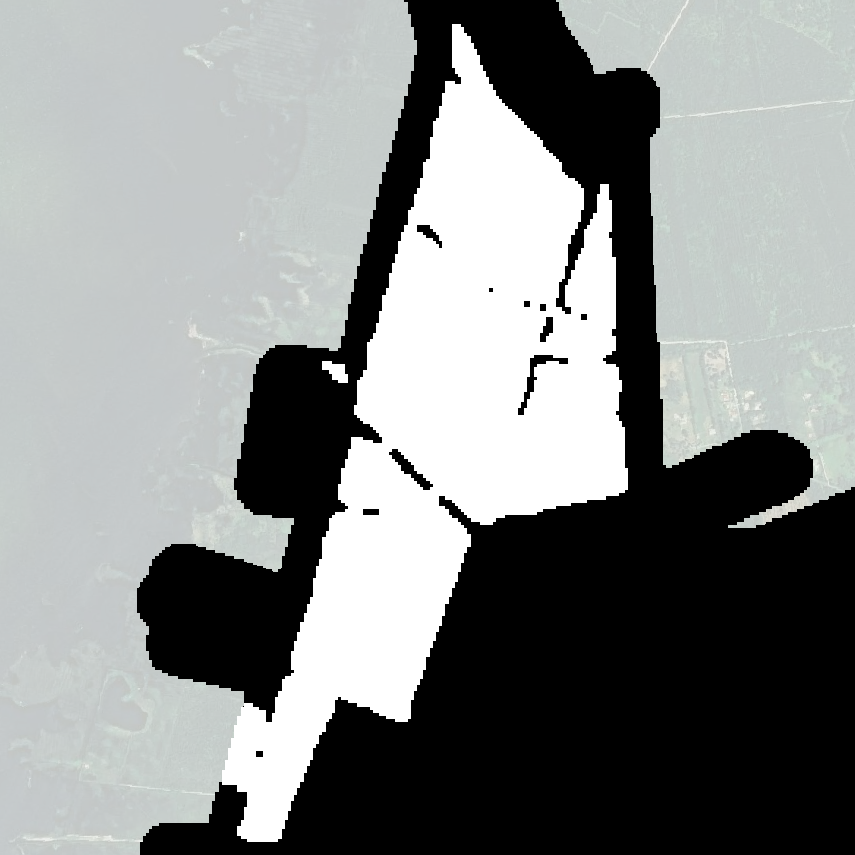}\caption{Ours 10~m}\end{subfigure} &
\begin{subfigure}{.067\textwidth}\includegraphics[width=\linewidth]{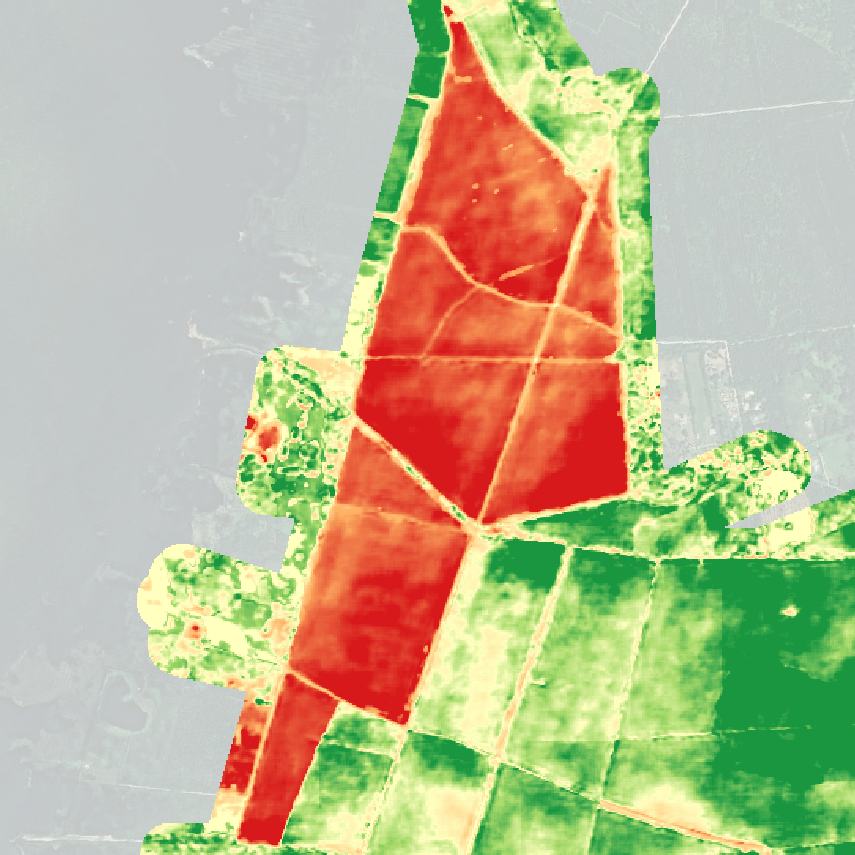}\caption{Ours 5~m}\end{subfigure} &
\begin{subfigure}{.067\textwidth}\includegraphics[width=\linewidth]{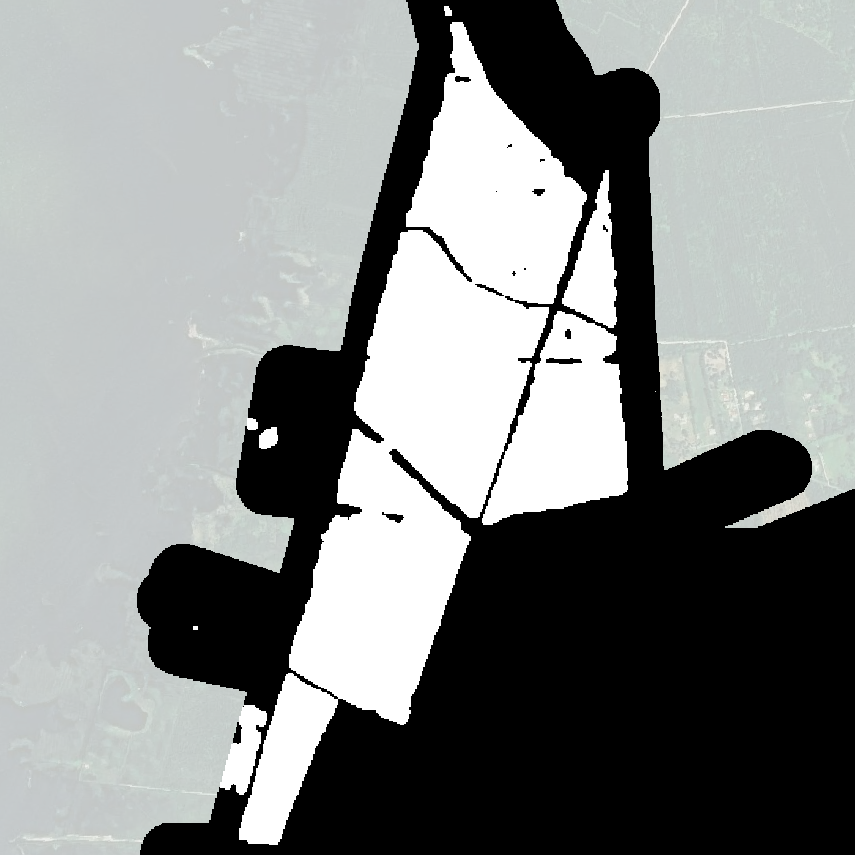}\caption{Ours 5~m}\end{subfigure} &
\begin{subfigure}{.067\textwidth}\includegraphics[width=\linewidth]{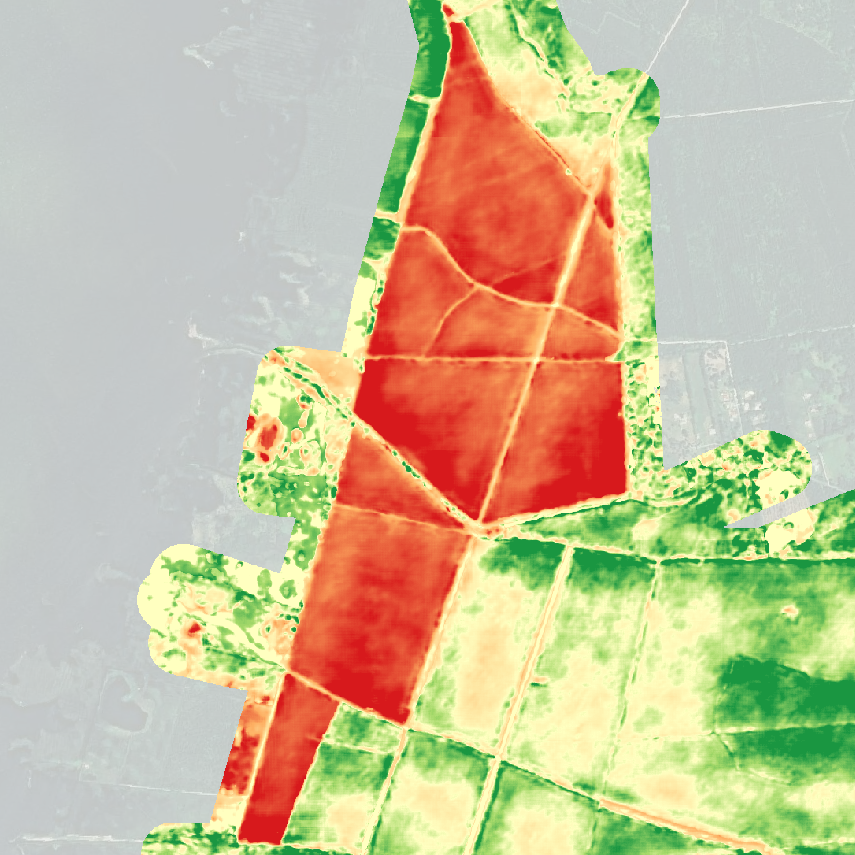}\caption{Ours 2.5~m}\end{subfigure} &
\begin{subfigure}{.067\textwidth}\includegraphics[width=\linewidth]{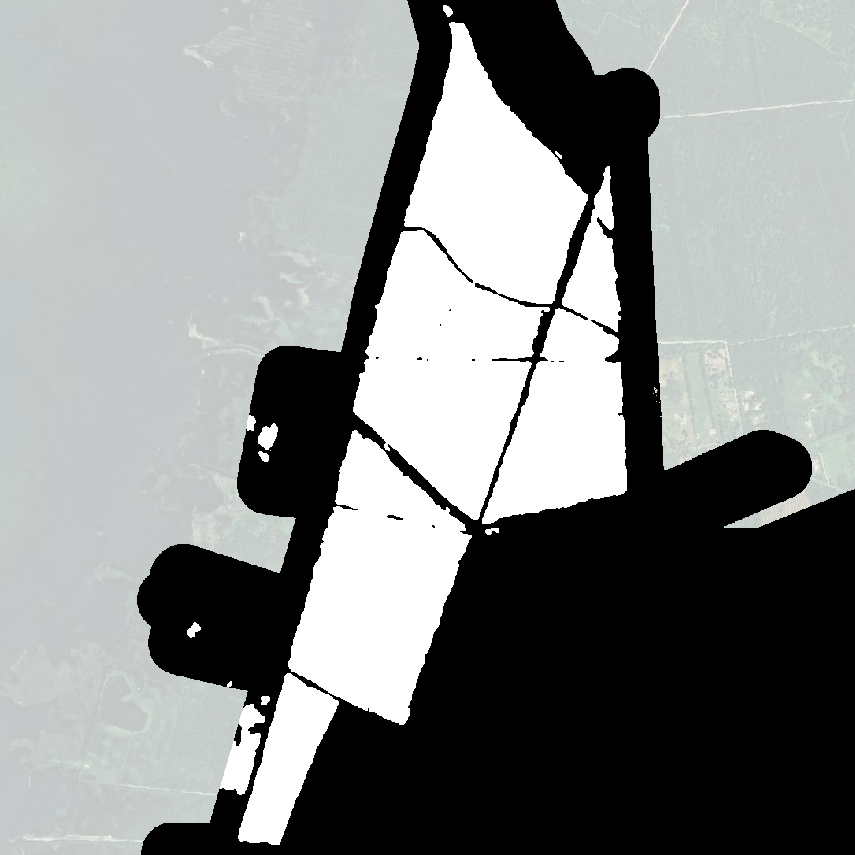}\caption{Ours 2.5~m}\end{subfigure} &
\begin{subfigure}{.067\textwidth}\includegraphics[width=\linewidth]{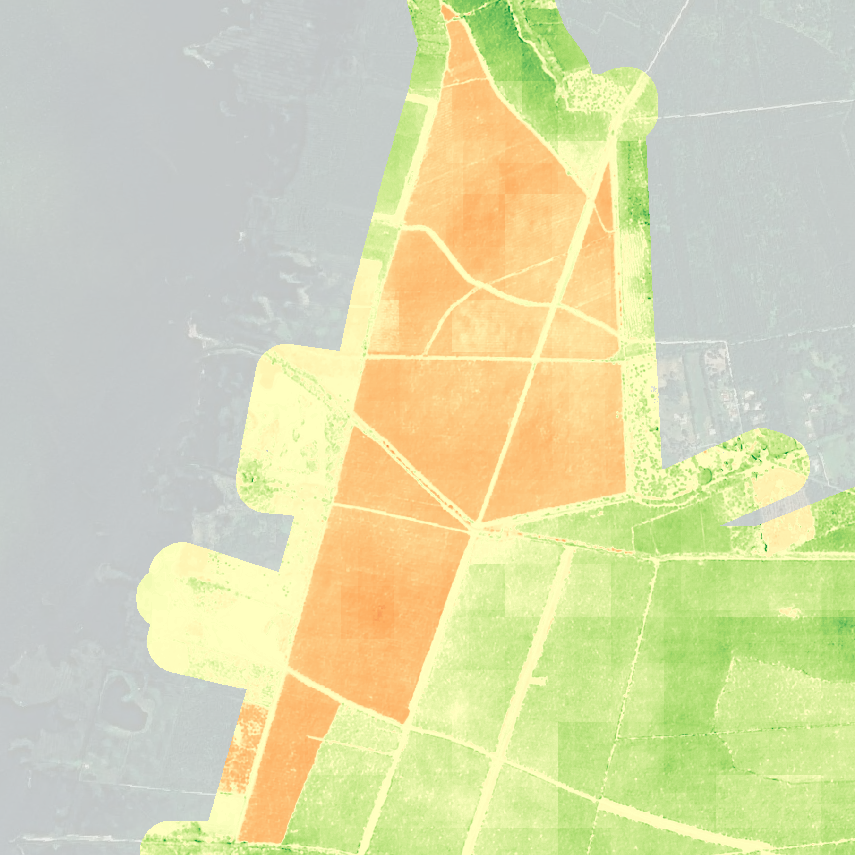}\caption{Open-Canopy}\end{subfigure} &
\begin{subfigure}{.067\textwidth}\includegraphics[width=\linewidth]{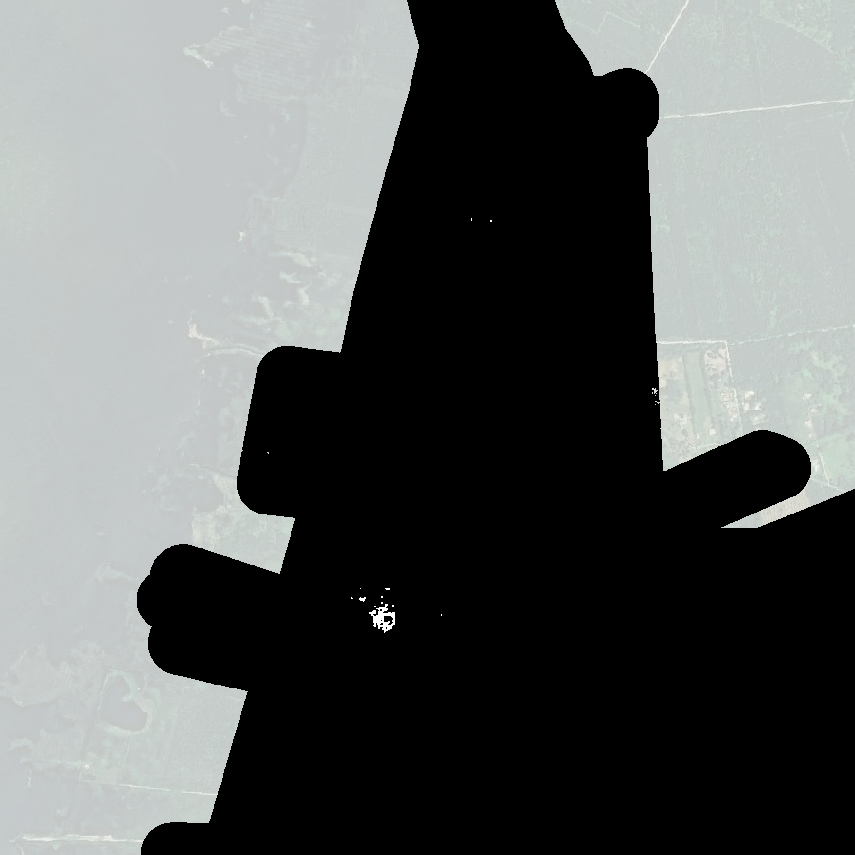}\caption{Open-Canopy}\end{subfigure} &
\begin{subfigure}{.067\textwidth}\includegraphics[width=\linewidth]{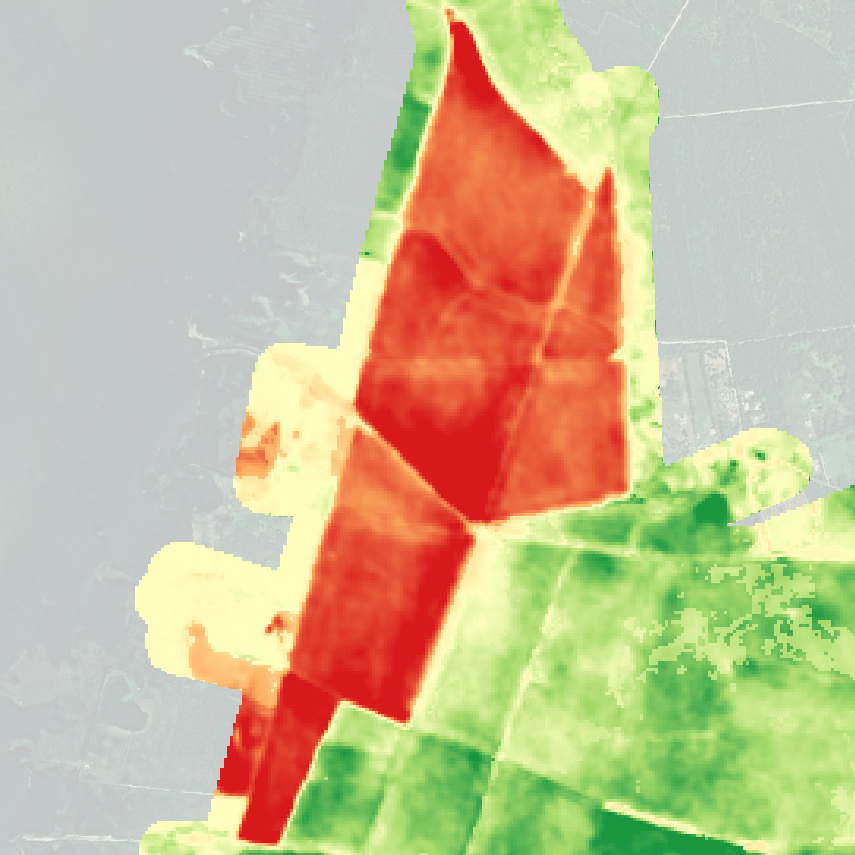}\caption{FORMS-T}\end{subfigure} &
\begin{subfigure}{.067\textwidth}\includegraphics[width=\linewidth]{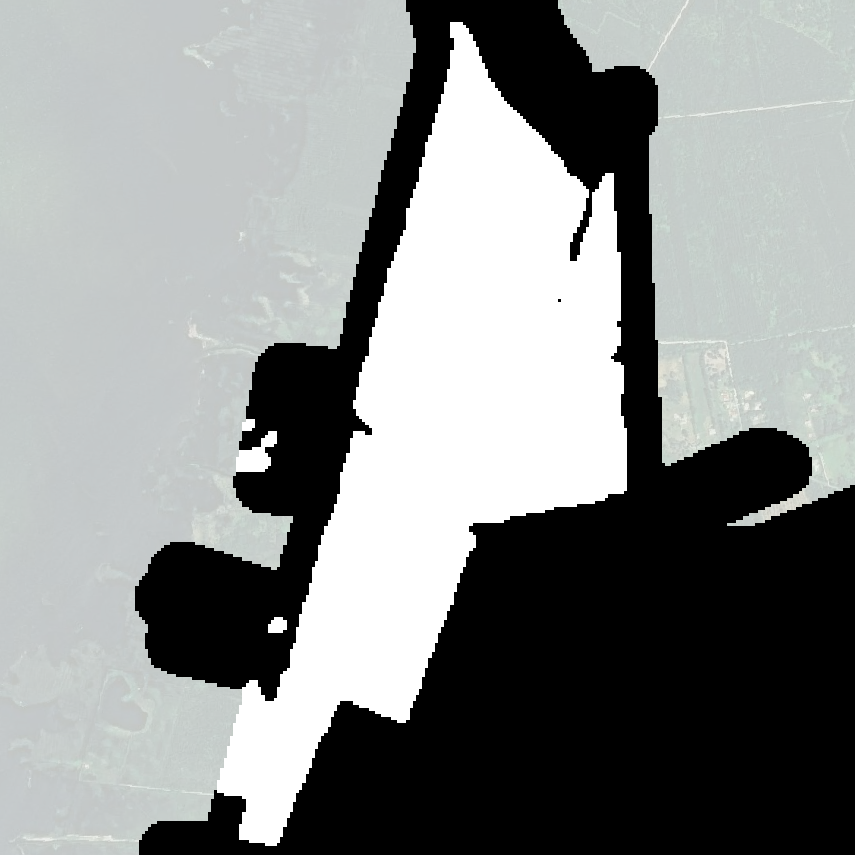}\caption{FORMS-T}\end{subfigure} &
\begin{subfigure}{.04\textwidth}\includegraphics[width=\linewidth]{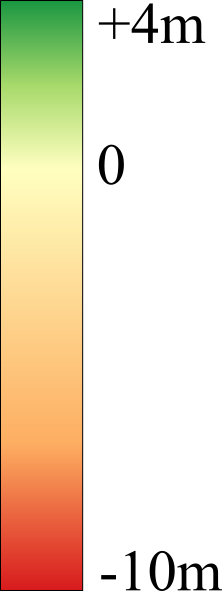}\end{subfigure} \\

\multicolumn{15}{c}{\textbf{Lajoux-Fresse}} \\[-2pt]
\setcounter{subfigure}{0}
\begin{subfigure}{.067\textwidth}\includegraphics[width=\linewidth]{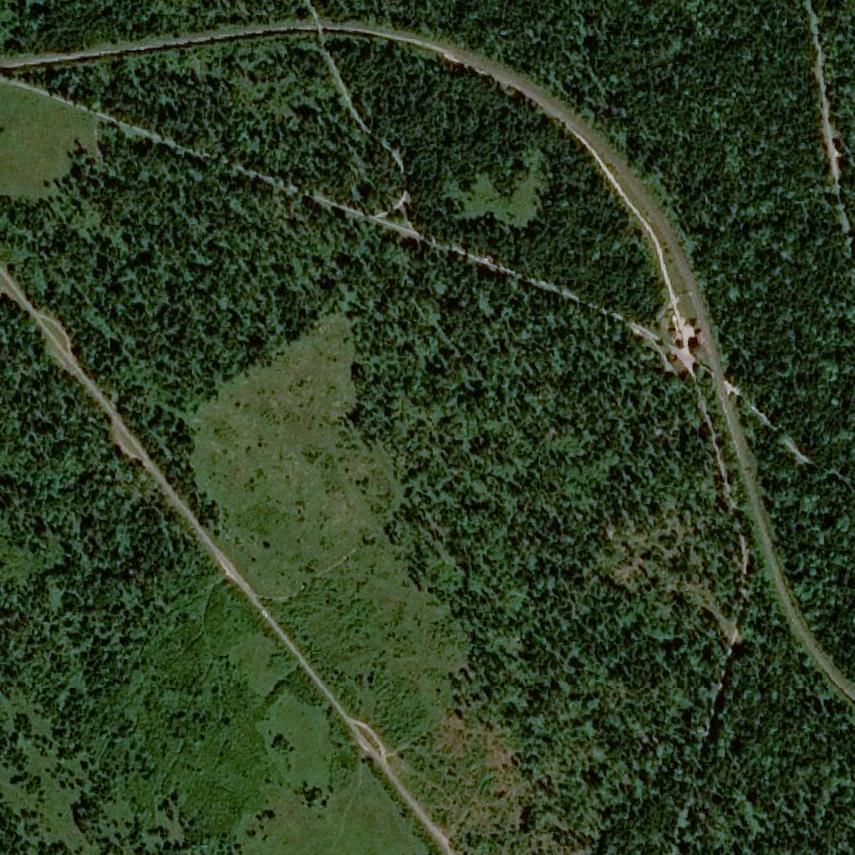}\caption{SPOT, 2019}\end{subfigure} &
\begin{subfigure}{.067\textwidth}\includegraphics[width=\linewidth]{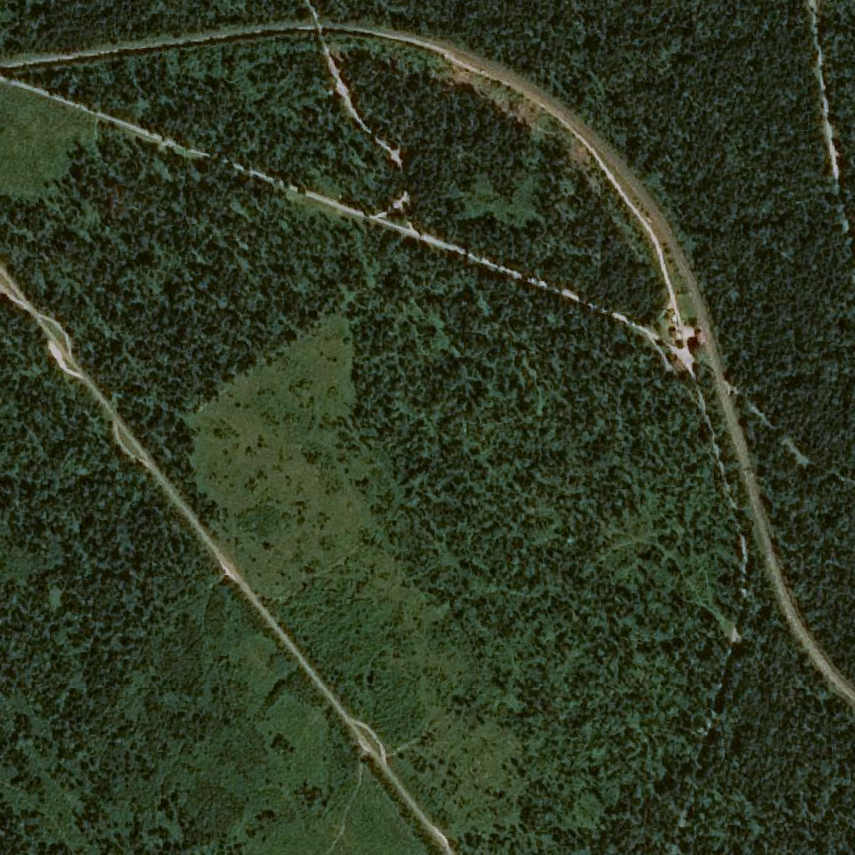}\caption{SPOT, 2022}\end{subfigure} &
\begin{subfigure}{.067\textwidth}\includegraphics[width=\linewidth]{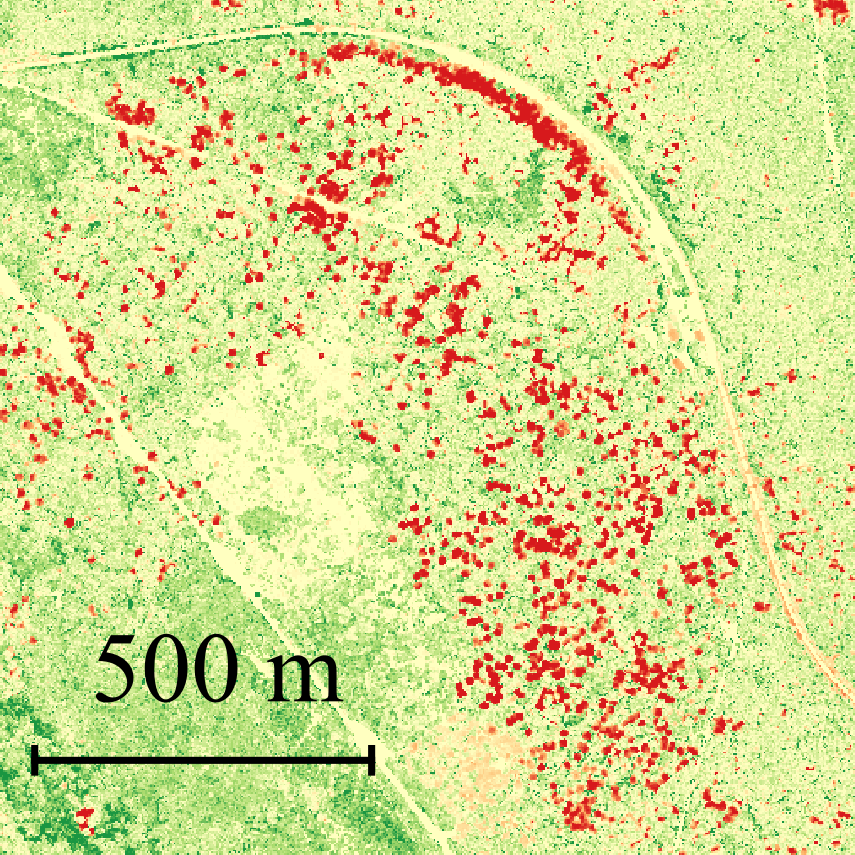}\caption{HDM ALS}\end{subfigure} &
\begin{subfigure}{.067\textwidth}\includegraphics[width=\linewidth]{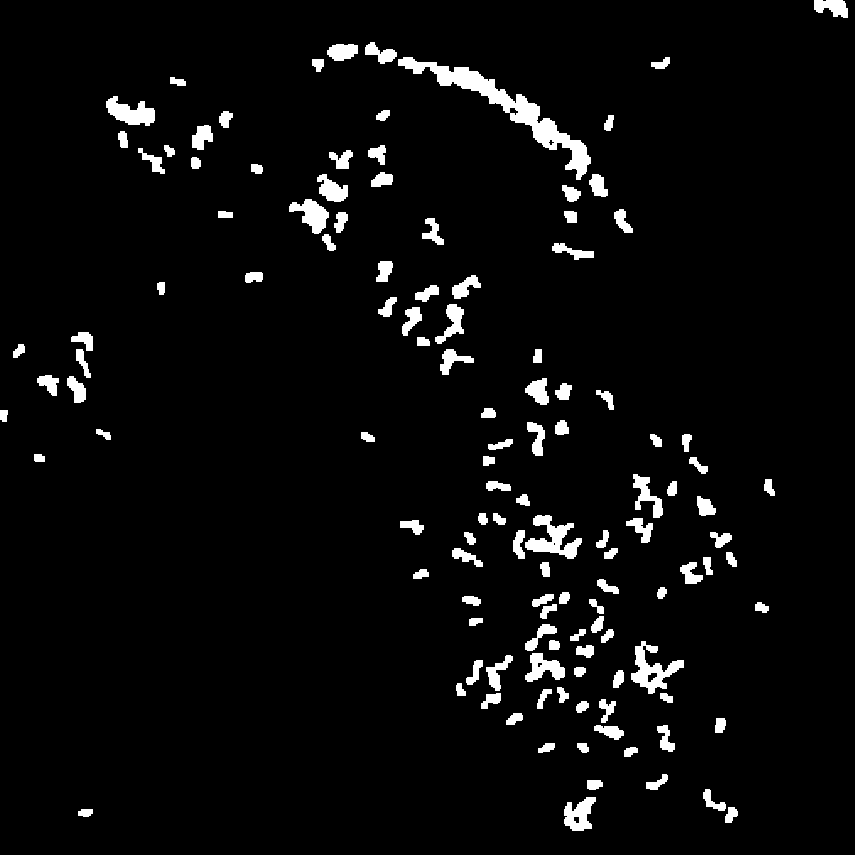}\caption{CM ALS}\end{subfigure} &
\begin{subfigure}{.067\textwidth}\includegraphics[width=\linewidth]{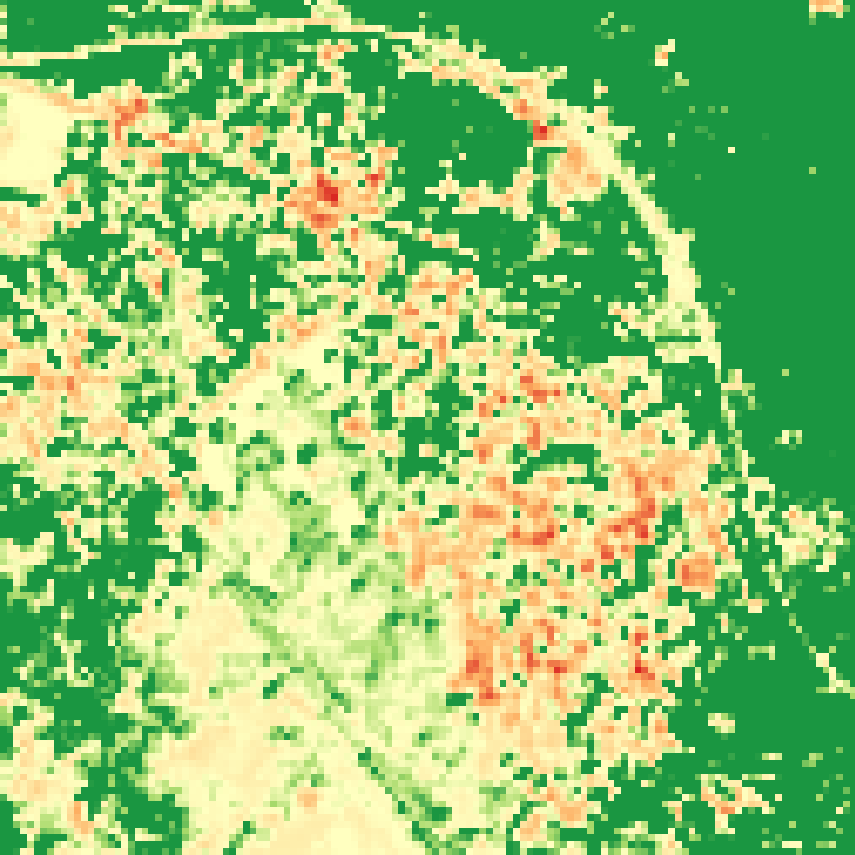}\caption{Ours 10~m}\end{subfigure} &
\begin{subfigure}{.067\textwidth}\includegraphics[width=\linewidth]{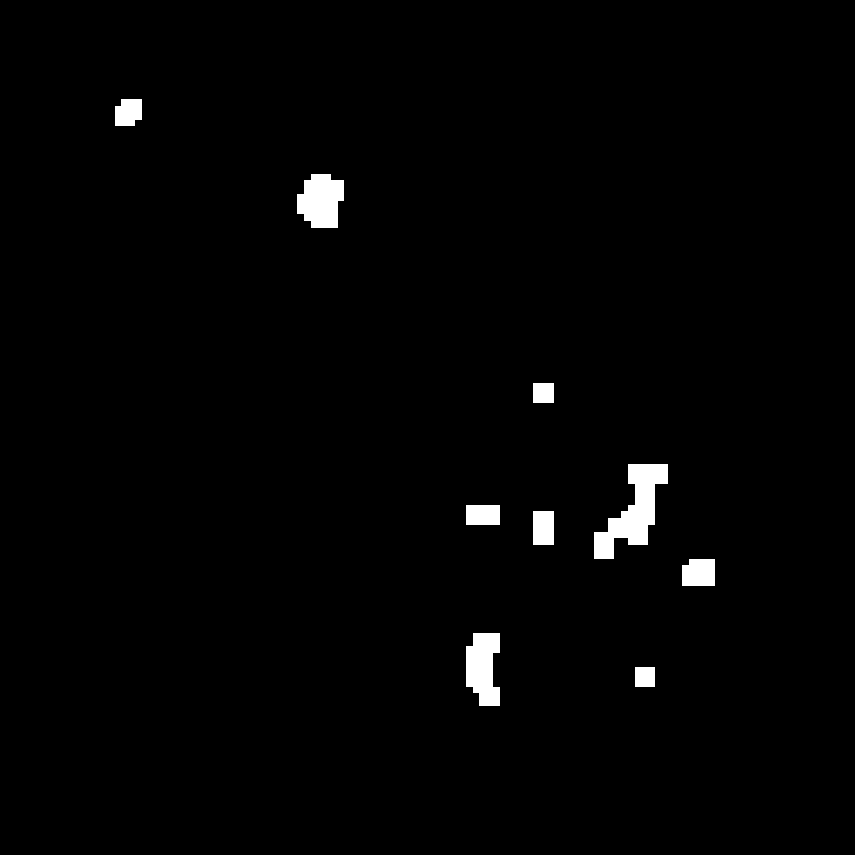}\caption{Ours 10~m}\end{subfigure} &
\begin{subfigure}{.067\textwidth}\includegraphics[width=\linewidth]{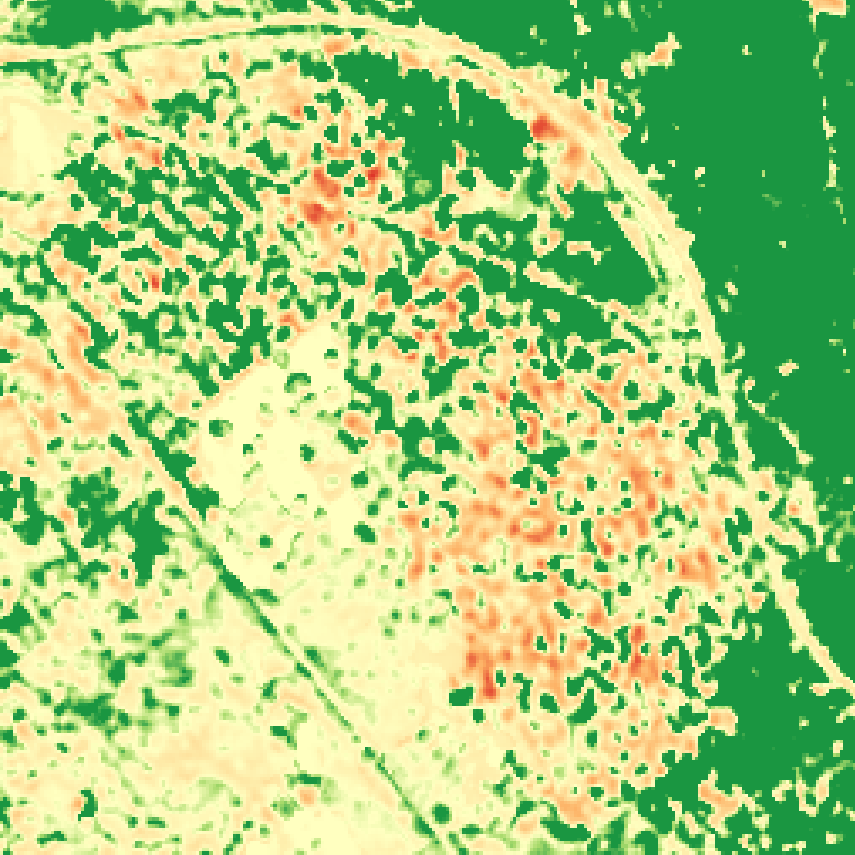}\caption{Ours 5~m}\end{subfigure} &
\begin{subfigure}{.067\textwidth}\includegraphics[width=\linewidth]{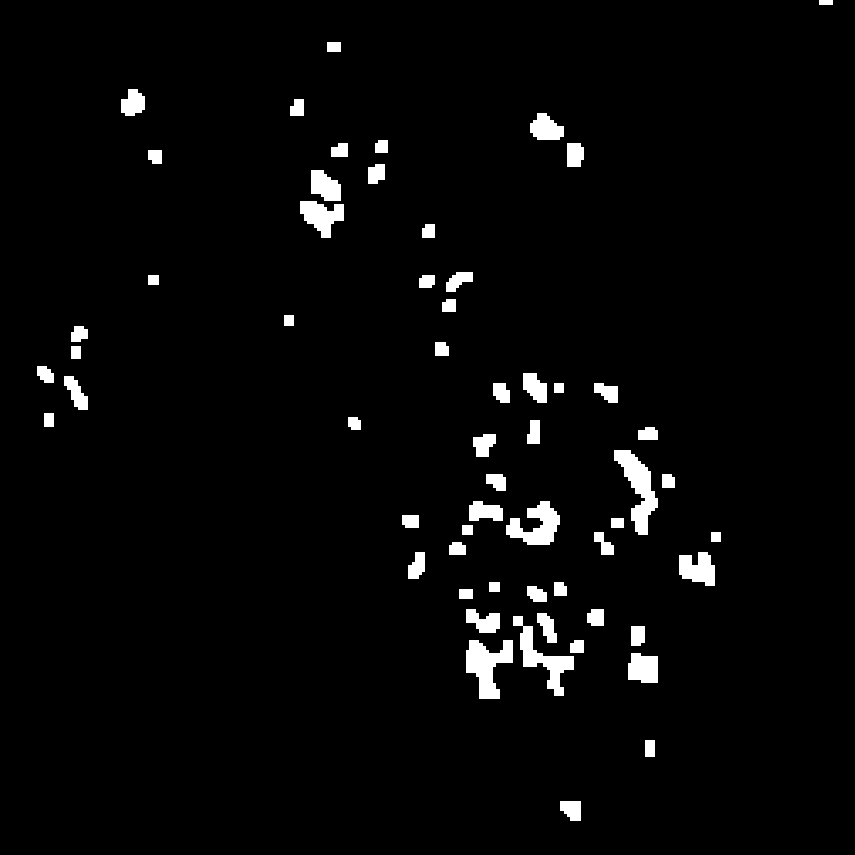}\caption{Ours 5~m}\end{subfigure} &
\begin{subfigure}{.067\textwidth}\includegraphics[width=\linewidth]{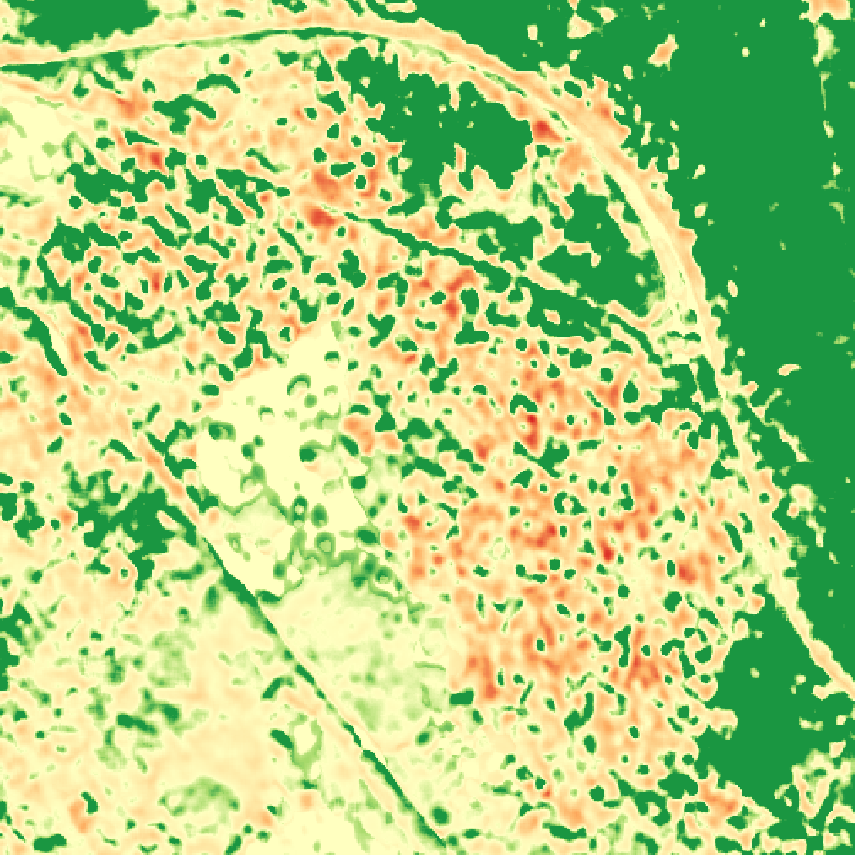}\caption{Ours 2.5~m}\end{subfigure} &
\begin{subfigure}{.067\textwidth}\includegraphics[width=\linewidth]{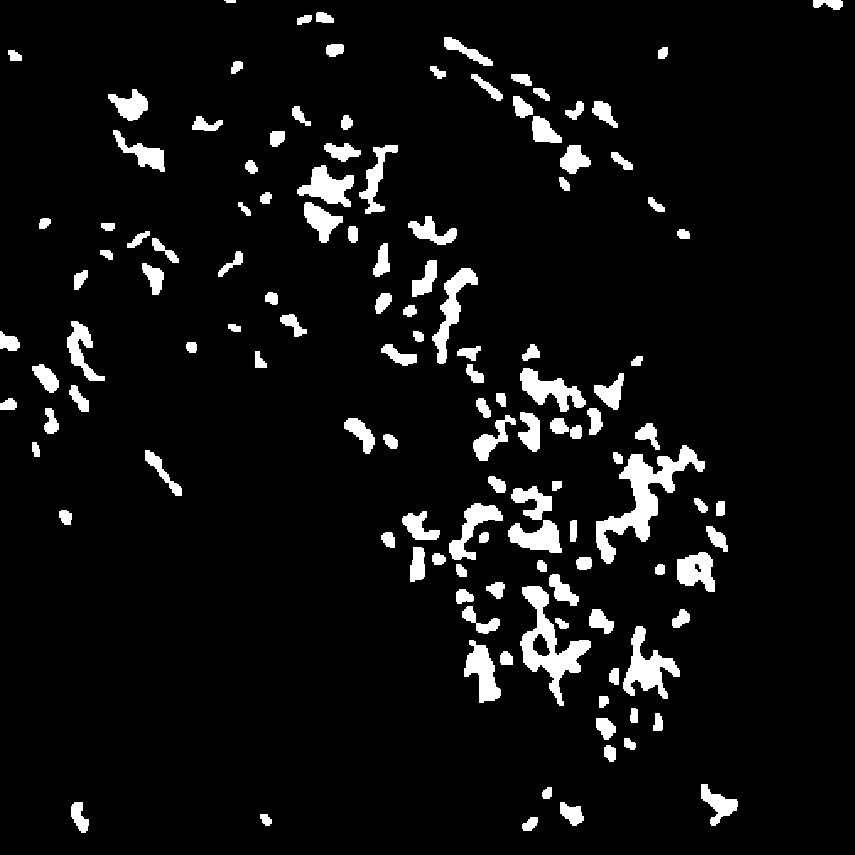}\caption{Ours 2.5~m}\end{subfigure} &
\begin{subfigure}{.067\textwidth}\includegraphics[width=\linewidth]{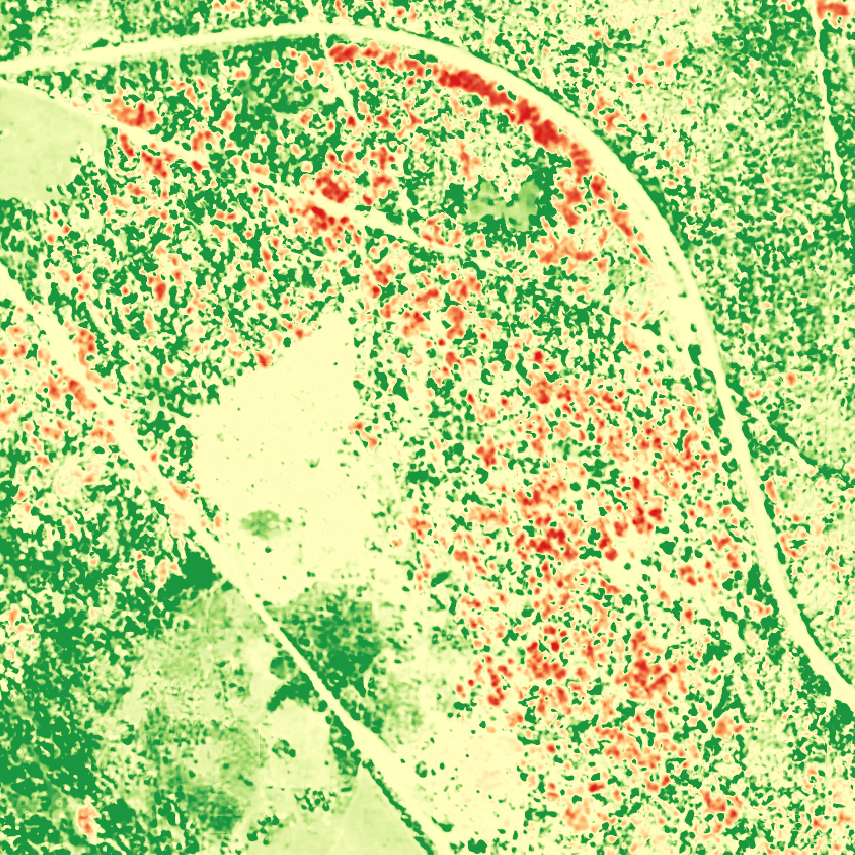}\caption{Open-Canopy}\end{subfigure} &
\begin{subfigure}{.067\textwidth}\includegraphics[width=\linewidth]{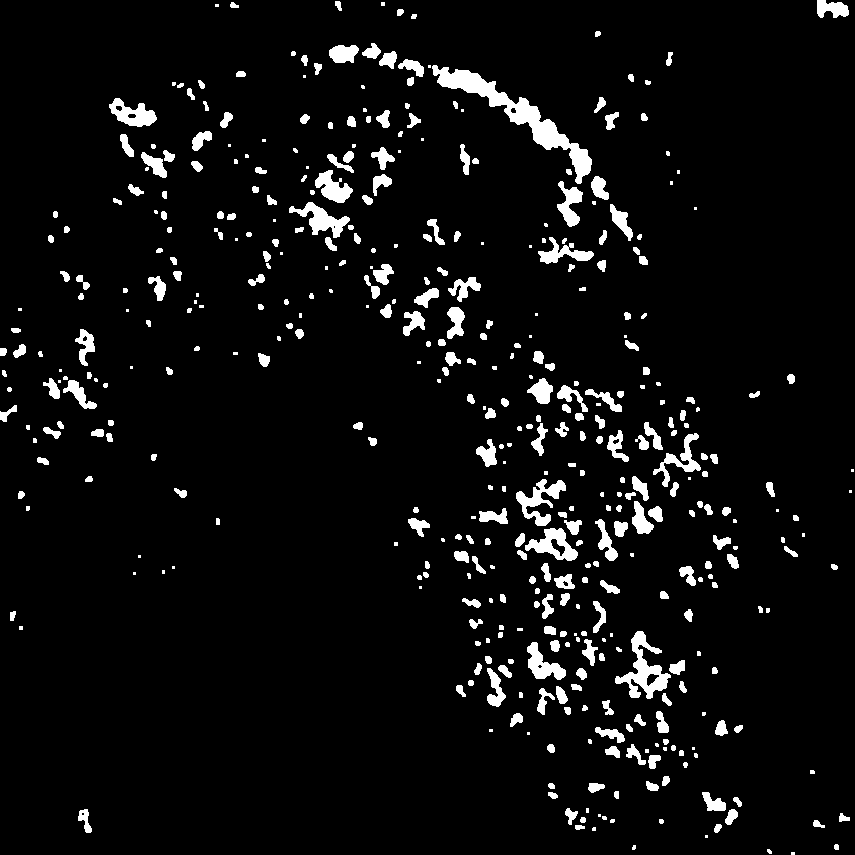}\caption{Open-Canopy}\end{subfigure} &
\begin{subfigure}{.067\textwidth}\includegraphics[width=\linewidth]{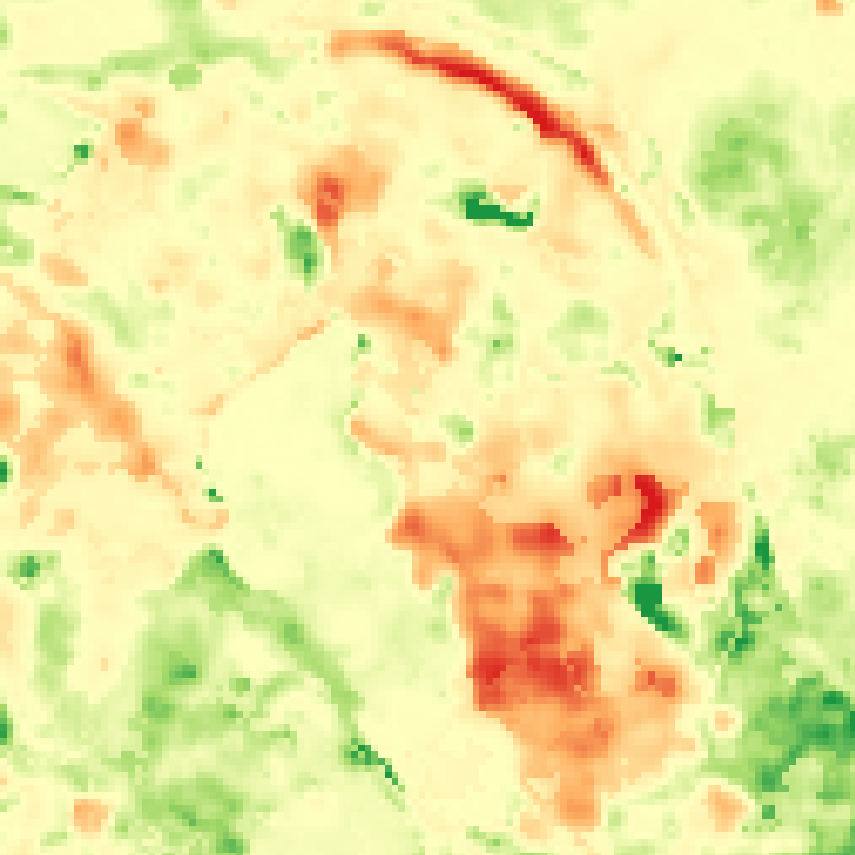}\caption{FORMS-T}\end{subfigure} &
\begin{subfigure}{.067\textwidth}\includegraphics[width=\linewidth]{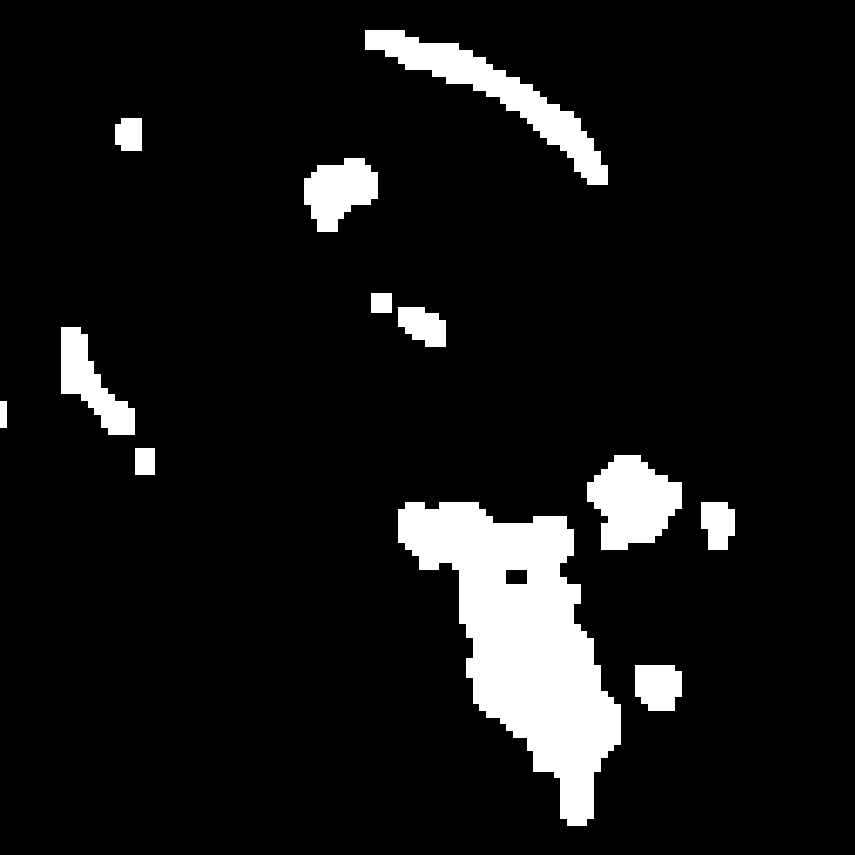}\caption{FORMS-T}\end{subfigure} &
\begin{subfigure}{.04\textwidth}\includegraphics[width=\linewidth]{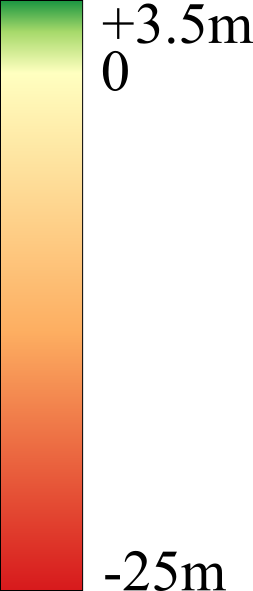}\end{subfigure} \\


\multicolumn{15}{c}{\textbf{Déodatie}} \\[-2pt]
\setcounter{subfigure}{0}
\begin{subfigure}{.067\textwidth}\includegraphics[width=\linewidth]{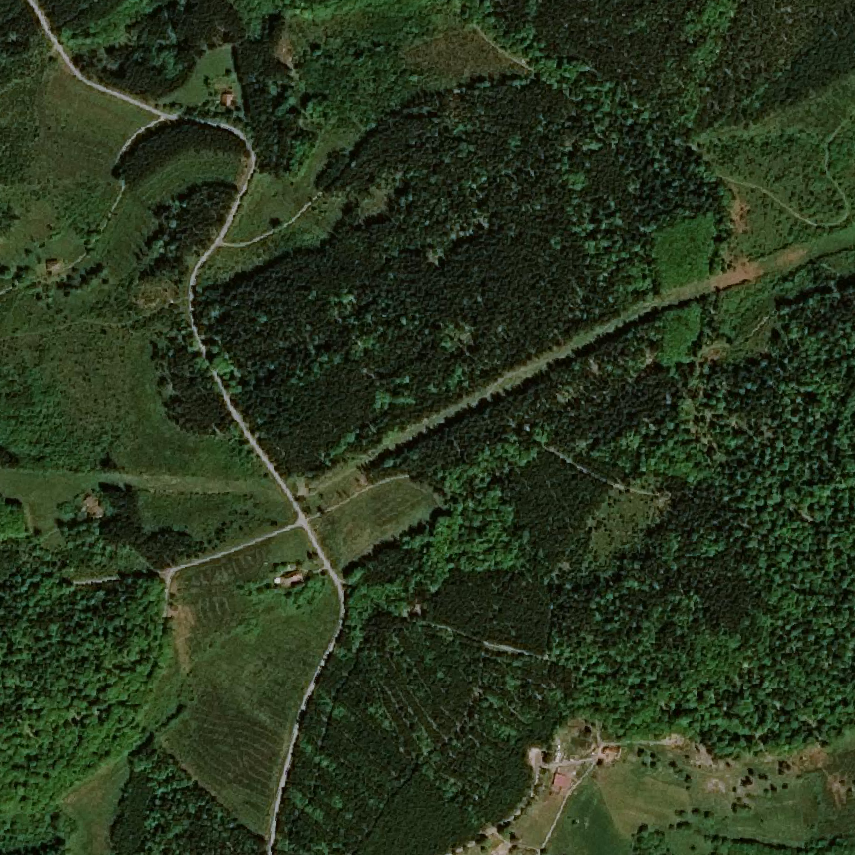}\caption{SPOT, 2018}\end{subfigure} &
\begin{subfigure}{.067\textwidth}\includegraphics[width=\linewidth]{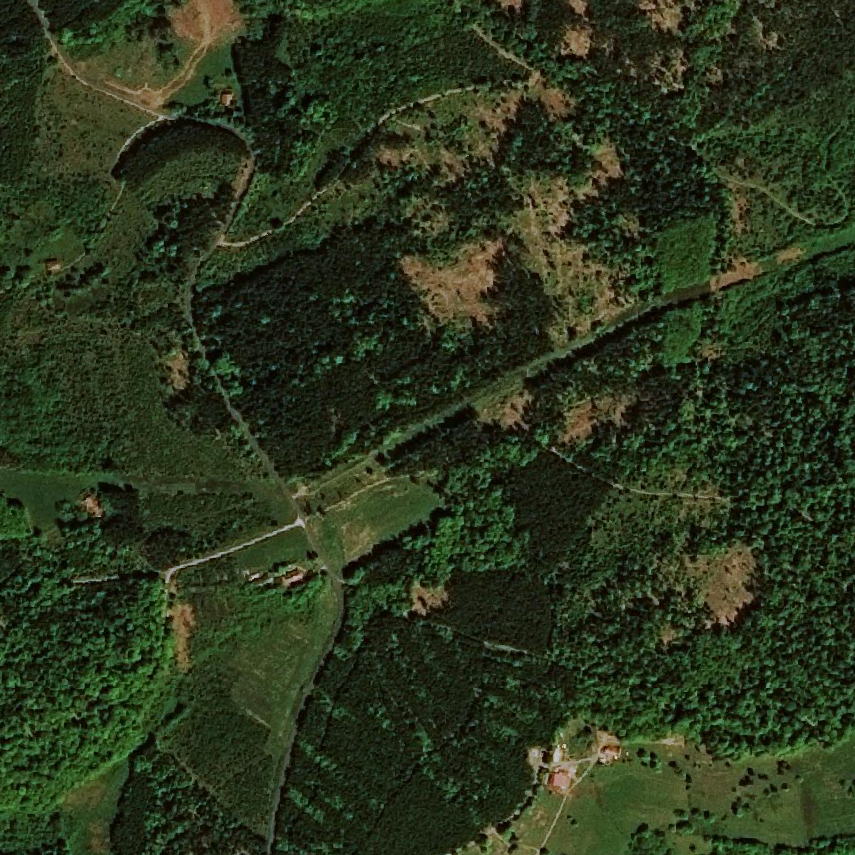}\caption{SPOT, 2022}\end{subfigure} &
\begin{subfigure}{.067\textwidth}\includegraphics[width=\linewidth]{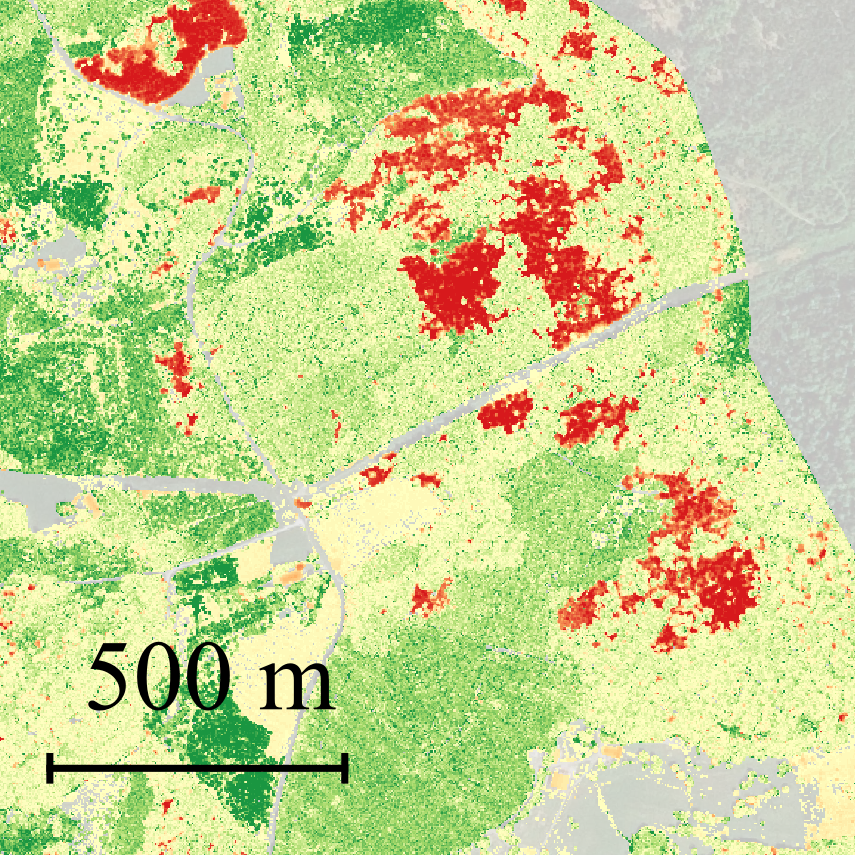}\caption{HDM ALS}\end{subfigure} &
\begin{subfigure}{.067\textwidth}\includegraphics[width=\linewidth]{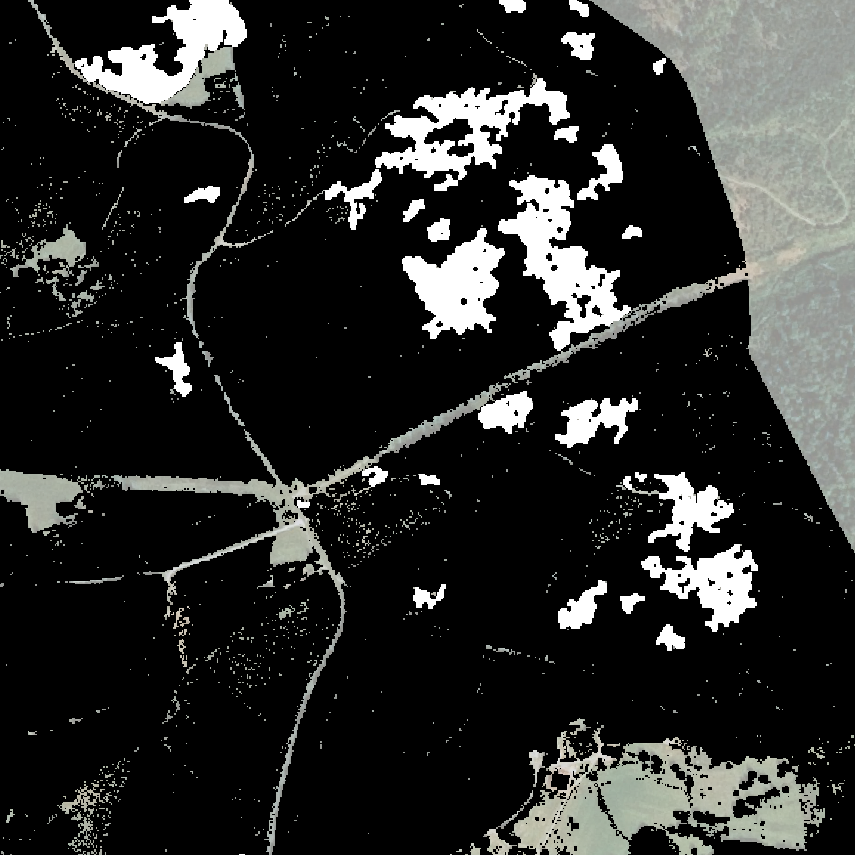}\caption{CM ALS}\end{subfigure} &
\begin{subfigure}{.067\textwidth}\includegraphics[width=\linewidth]{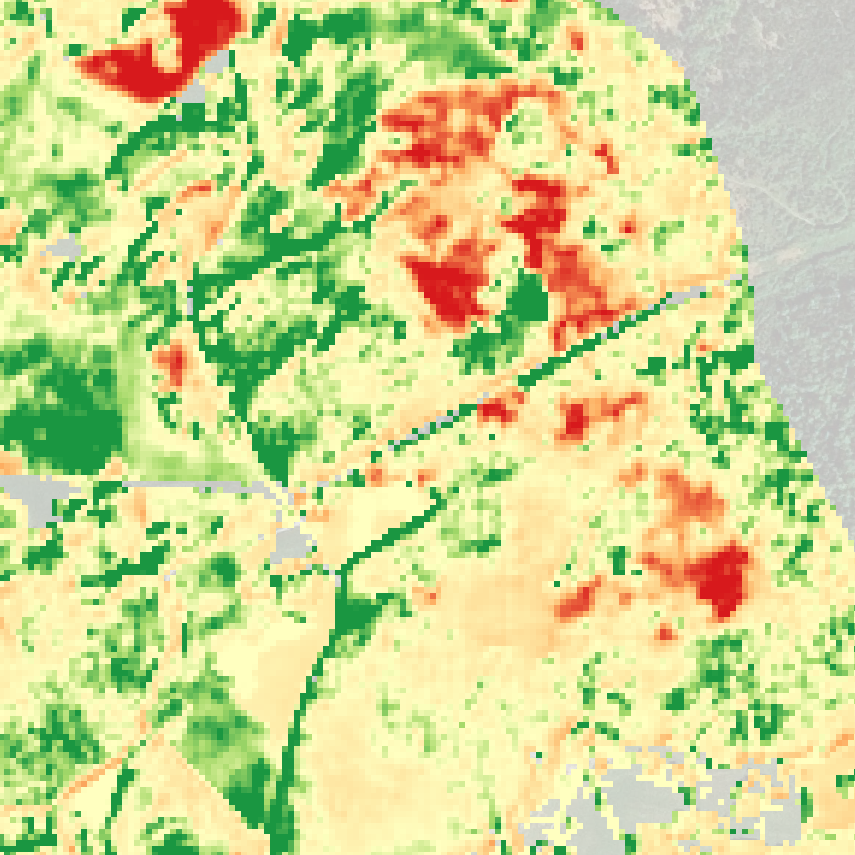}\caption{Ours 10~m}\end{subfigure} &
\begin{subfigure}{.067\textwidth}\includegraphics[width=\linewidth]{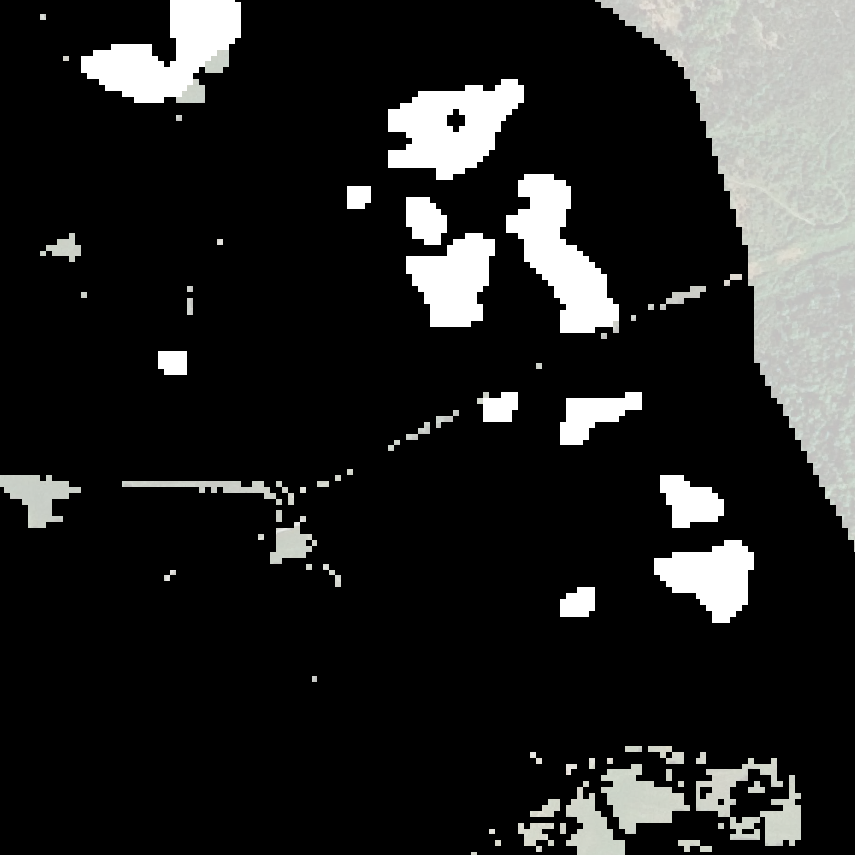}\caption{Ours 10~m}\end{subfigure} &
\begin{subfigure}{.067\textwidth}\includegraphics[width=\linewidth]{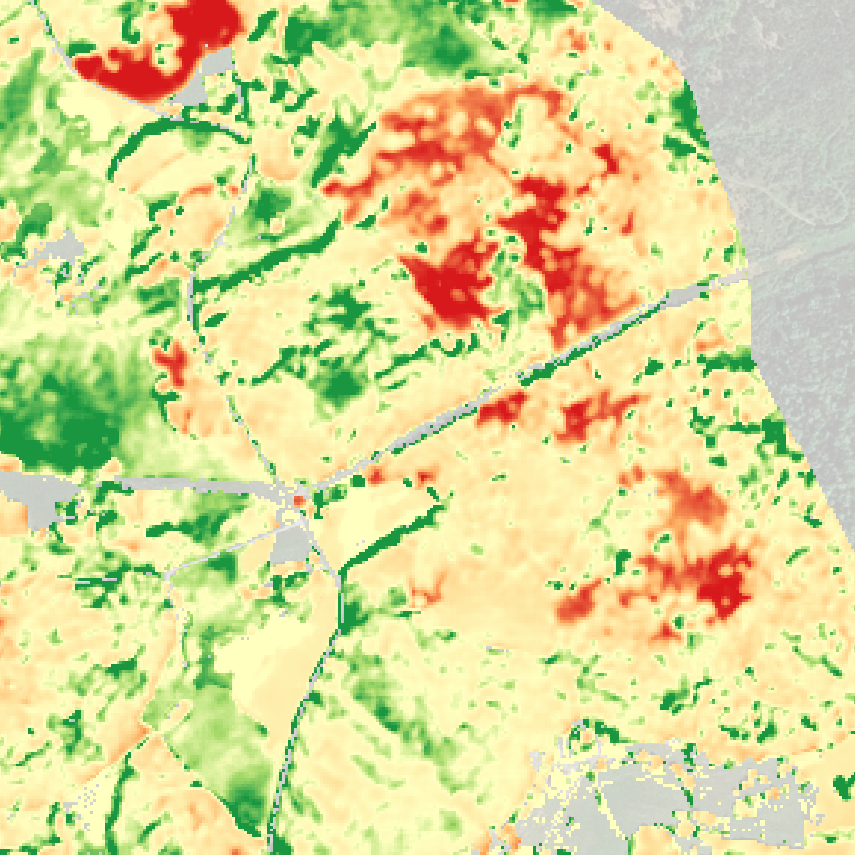}\caption{Ours 5~m}\end{subfigure} &
\begin{subfigure}{.067\textwidth}\includegraphics[width=\linewidth]{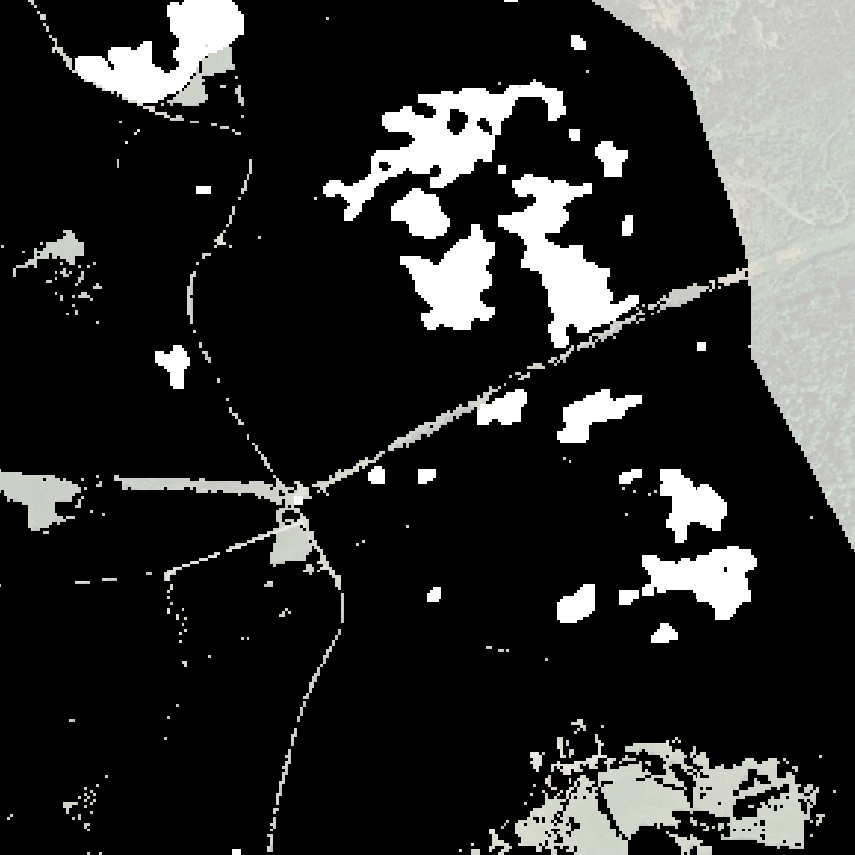}\caption{Ours 5~m}\end{subfigure} &
\begin{subfigure}{.067\textwidth}\includegraphics[width=\linewidth]{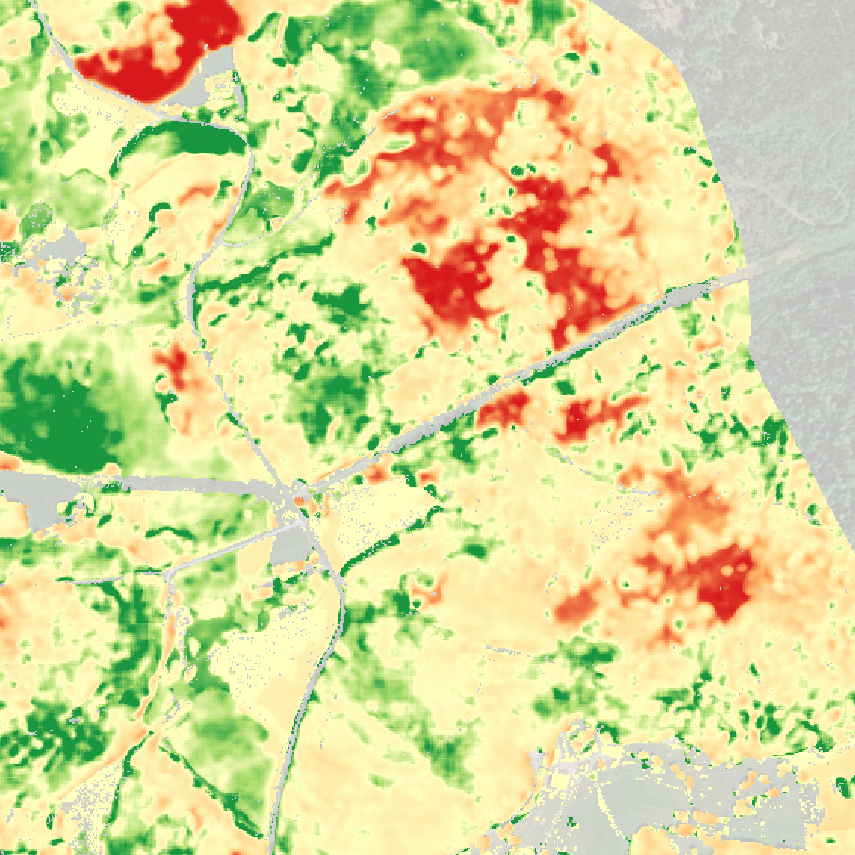}\caption{Ours 2.5~m}\end{subfigure} &
\begin{subfigure}{.067\textwidth}\includegraphics[width=\linewidth]{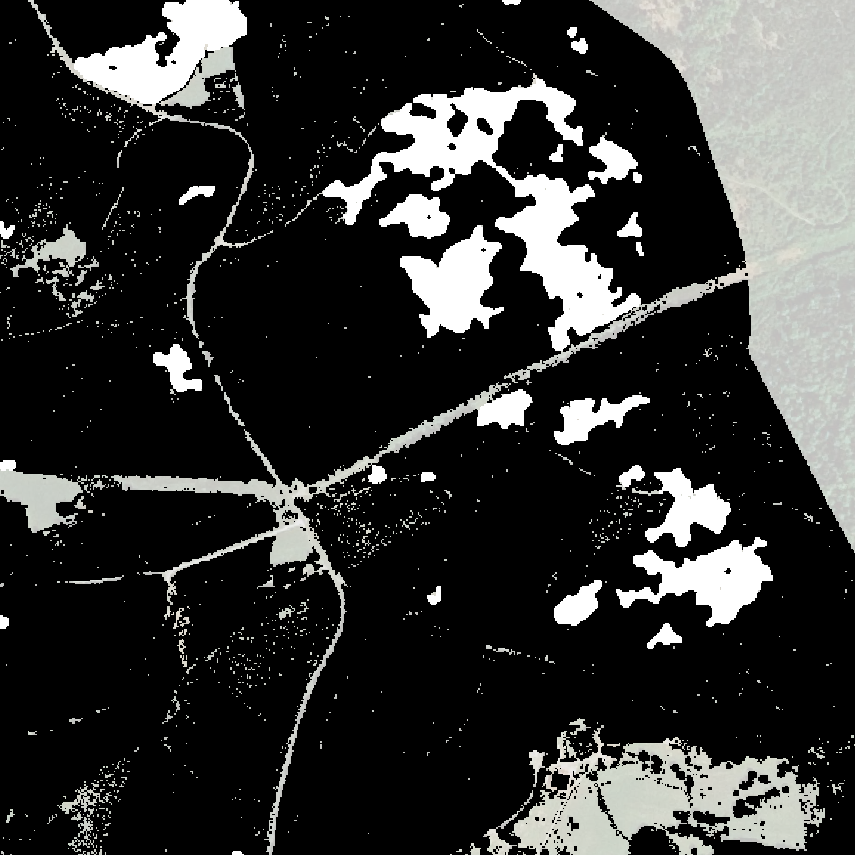}\caption{Ours 2.5~m}\end{subfigure} &
\begin{subfigure}{.067\textwidth}\includegraphics[width=\linewidth]{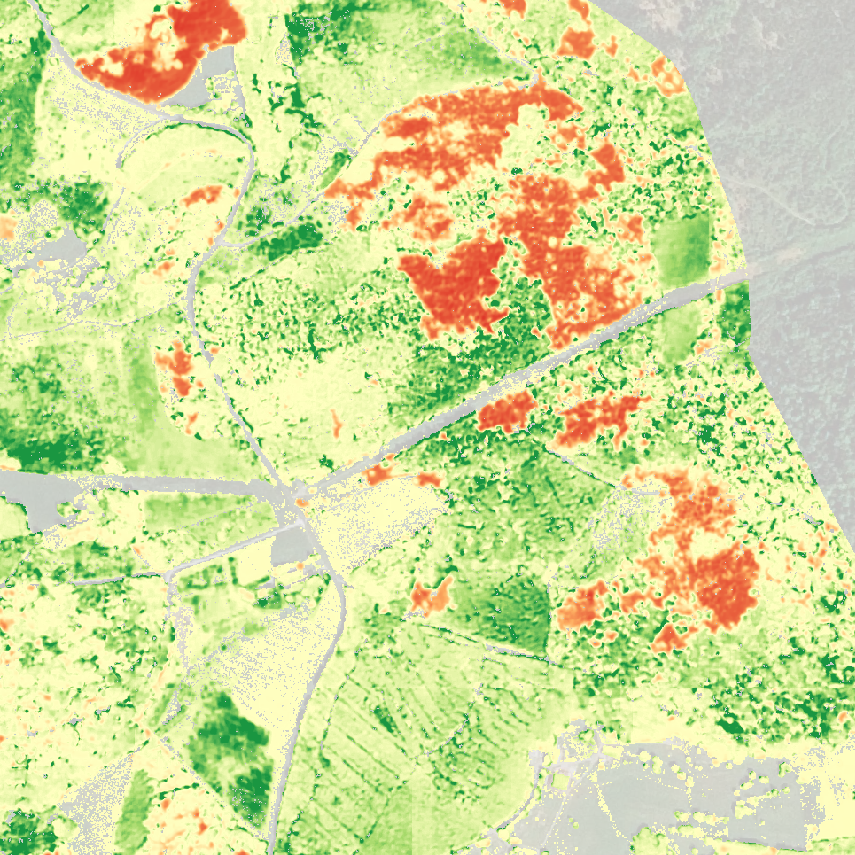}\caption{Open-Canopy}\end{subfigure} &
\begin{subfigure}{.067\textwidth}\includegraphics[width=\linewidth]{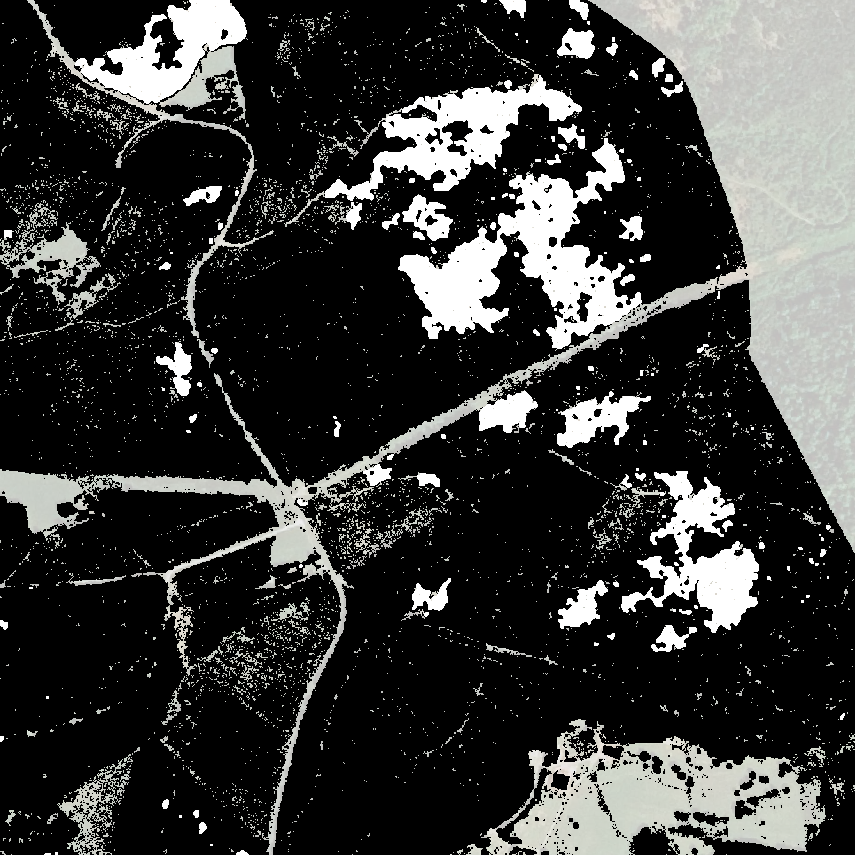}\caption{Open-Canopy}\end{subfigure} &
\begin{subfigure}{.067\textwidth}\includegraphics[width=\linewidth]{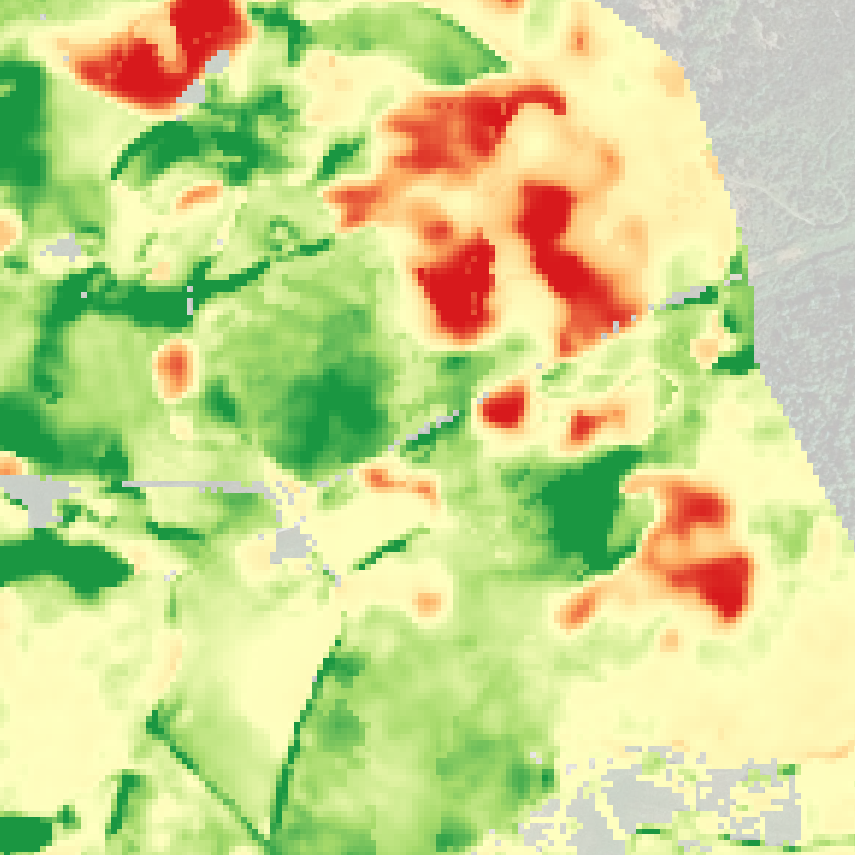}\caption{FORMS-T}\end{subfigure} &
\begin{subfigure}{.067\textwidth}\includegraphics[width=\linewidth]{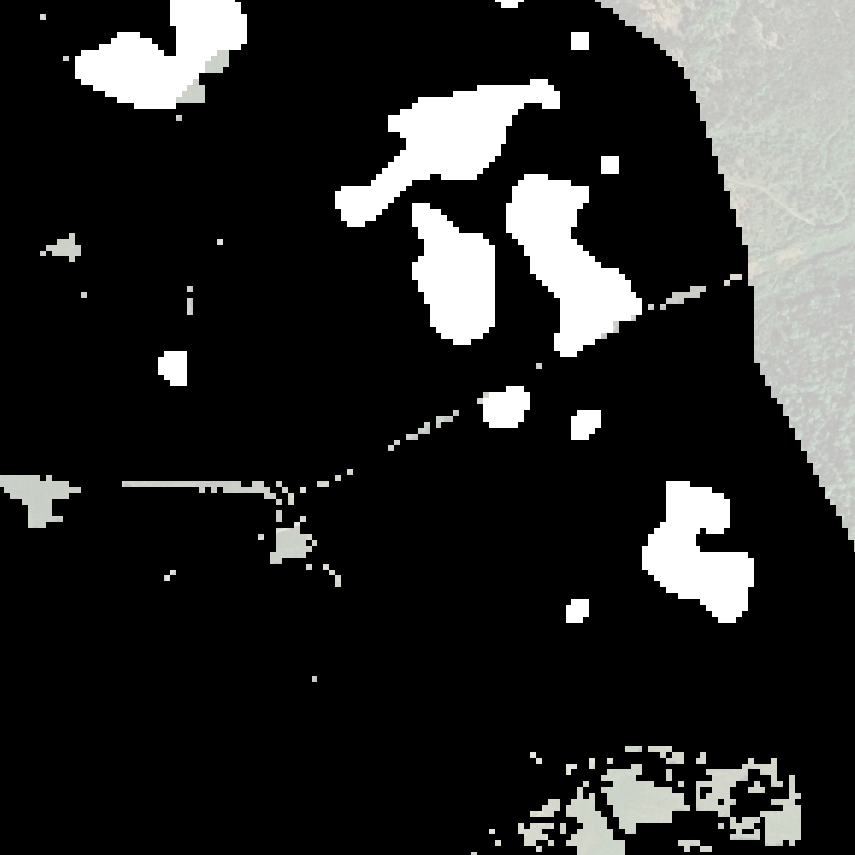}\caption{FORMS-T}\end{subfigure} &
\begin{subfigure}{.04\textwidth}\includegraphics[width=\linewidth]{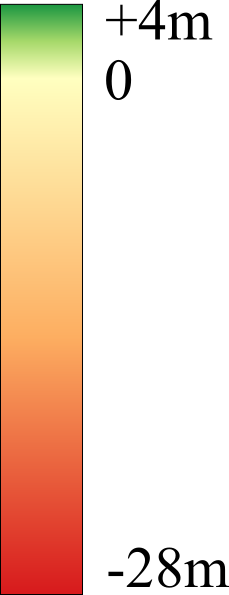}\end{subfigure} \\

\multicolumn{15}{c}{\textbf{Mouterhouse}} \\[-2pt]
\setcounter{subfigure}{0}
\begin{subfigure}{.067\textwidth}\includegraphics[width=\linewidth]{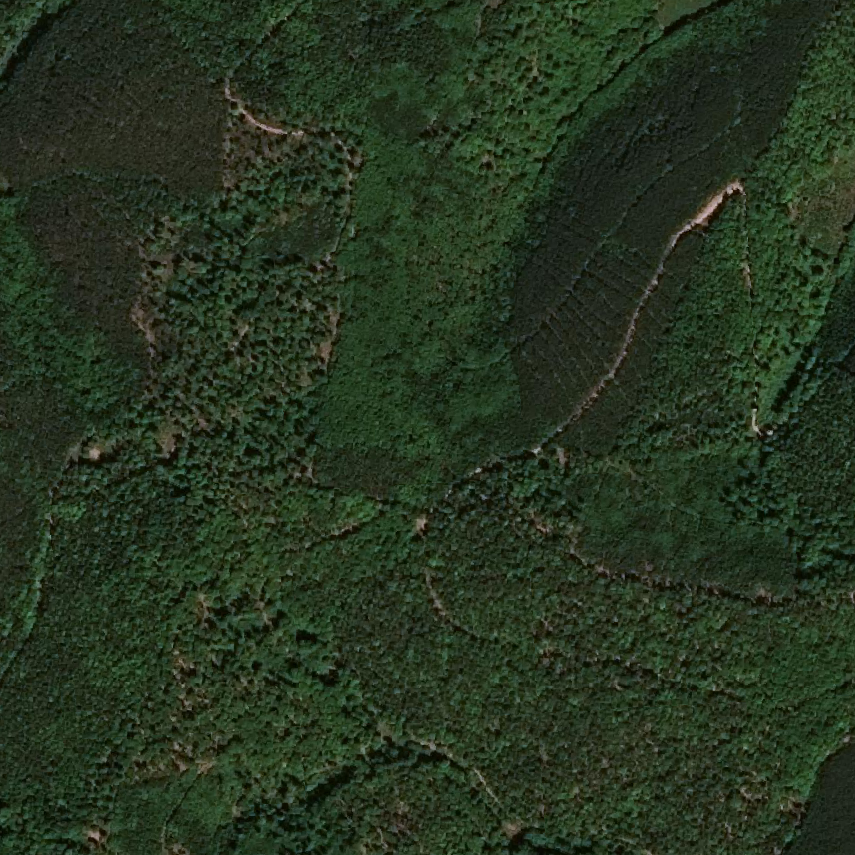}\caption{SPOT, 2019}\end{subfigure} &
\begin{subfigure}{.067\textwidth}\includegraphics[width=\linewidth]{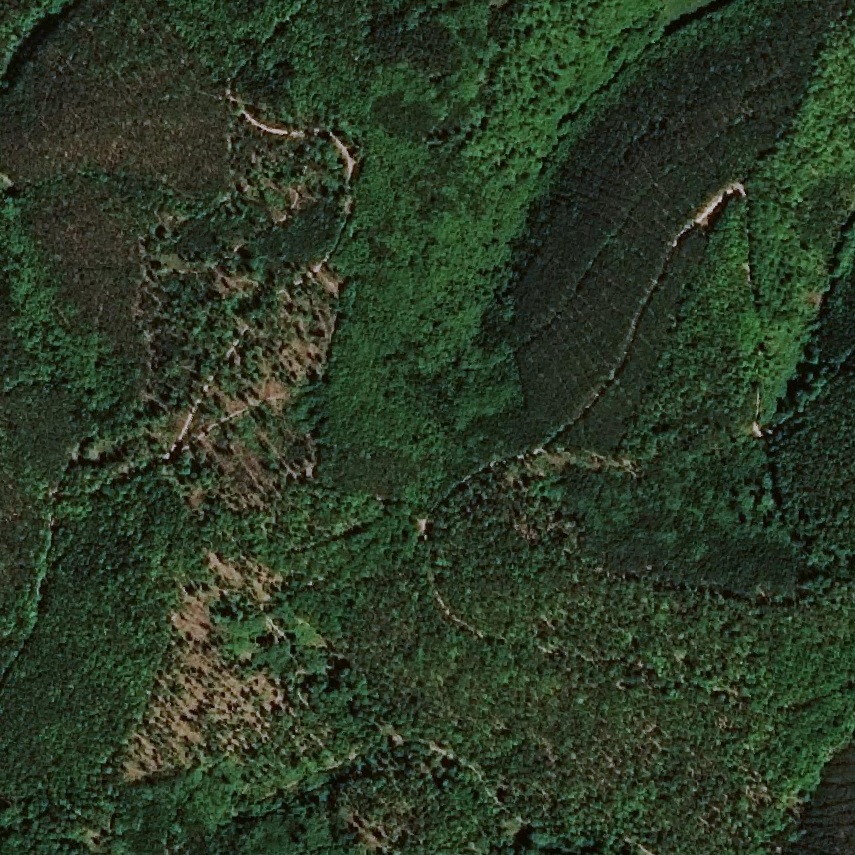}\caption{SPOT, 2022}\end{subfigure} &
\begin{subfigure}{.067\textwidth}\includegraphics[width=\linewidth]{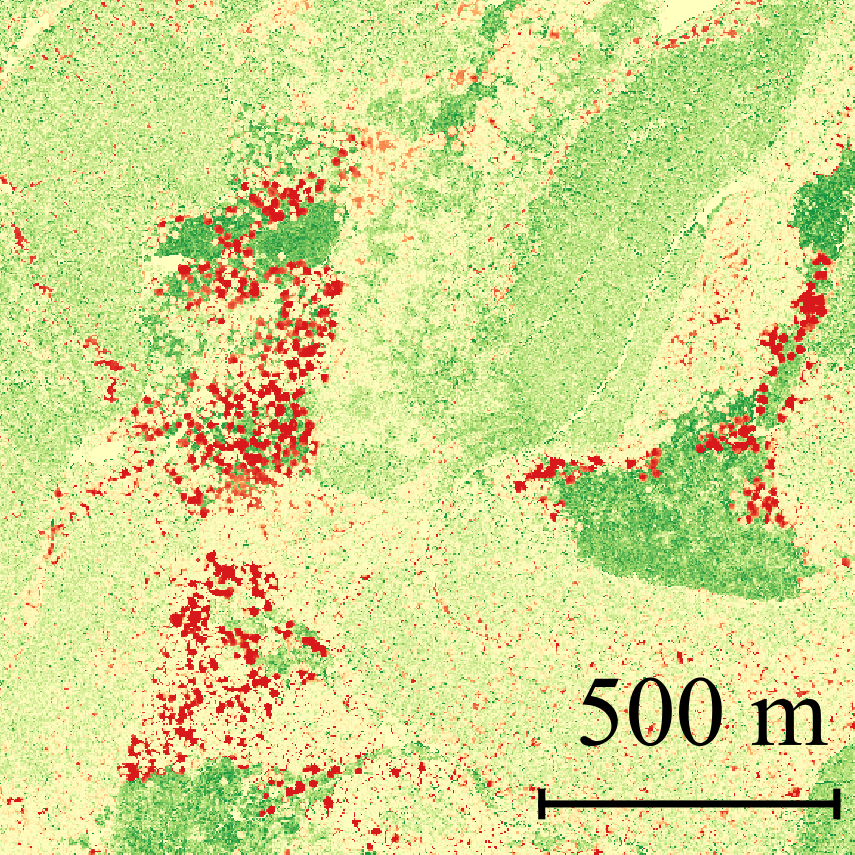}\caption{HDM ALS}\end{subfigure} &
\begin{subfigure}{.067\textwidth}\includegraphics[width=\linewidth]{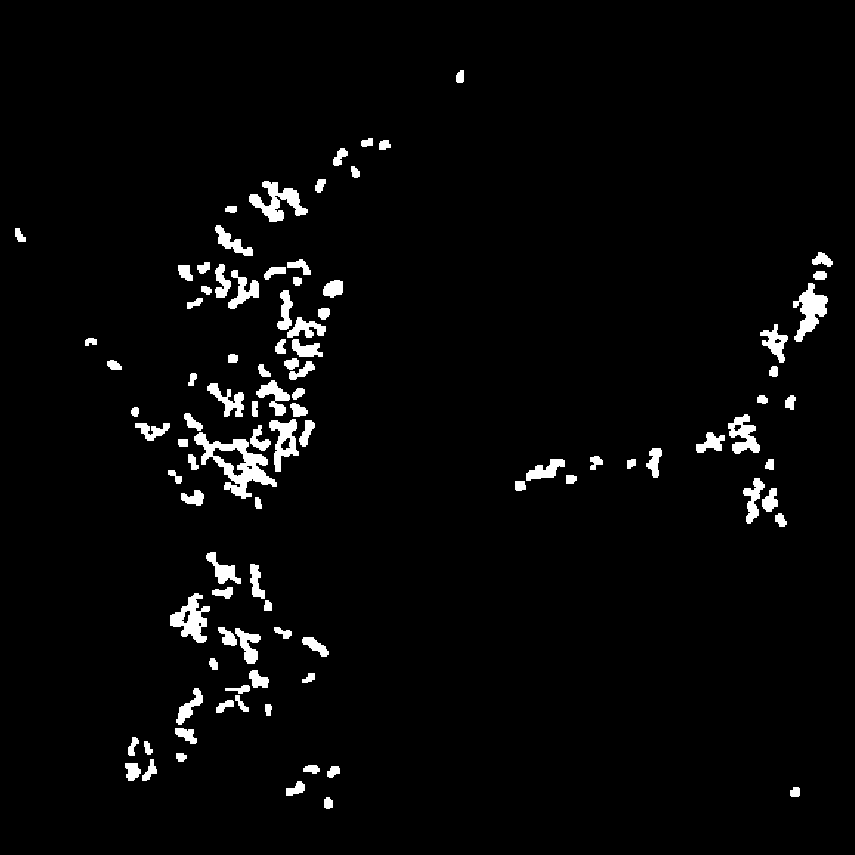}\caption{CM ALS}\end{subfigure} &
\begin{subfigure}{.067\textwidth}\includegraphics[width=\linewidth]{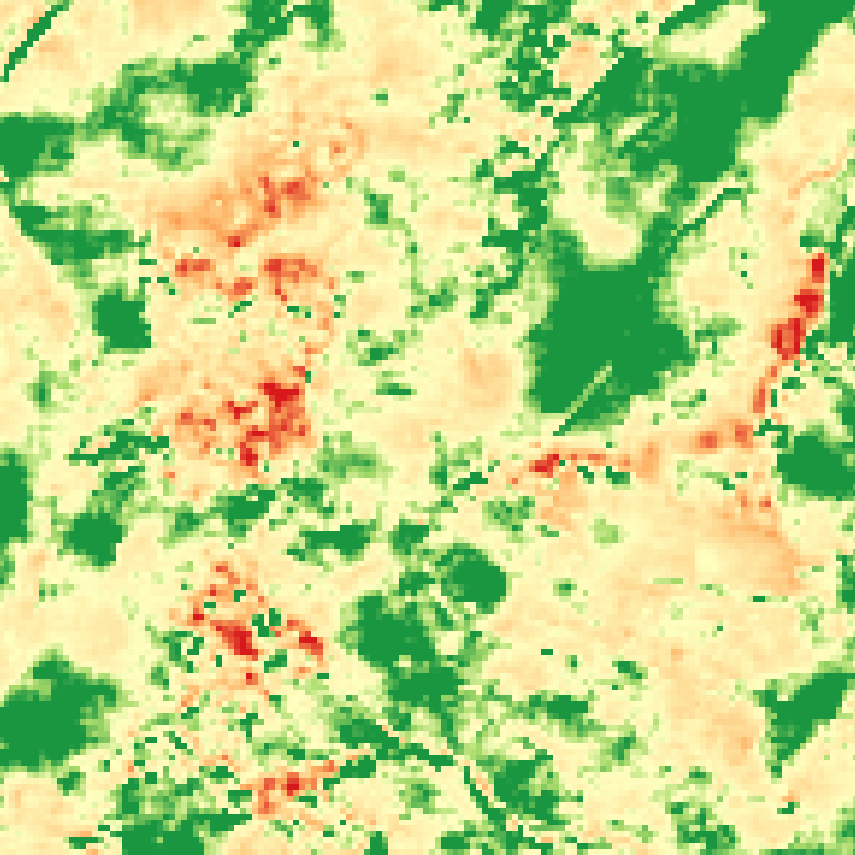}\caption{Ours 10~m}\end{subfigure} &
\begin{subfigure}{.067\textwidth}\includegraphics[width=\linewidth]{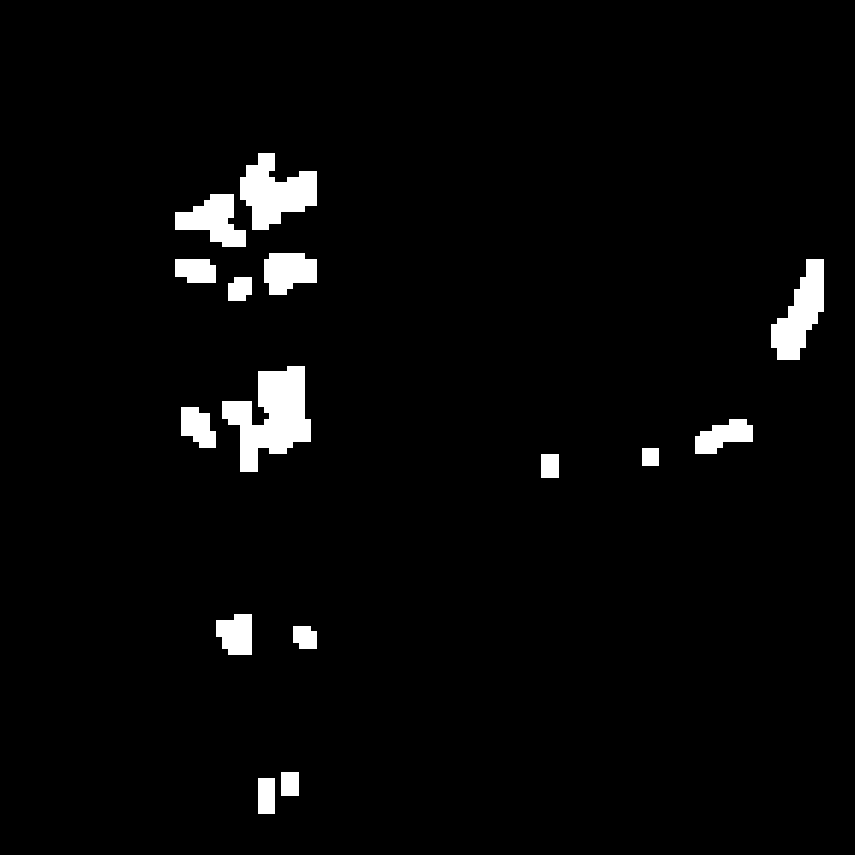}\caption{Ours 10~m}\end{subfigure} &
\begin{subfigure}{.067\textwidth}\includegraphics[width=\linewidth]{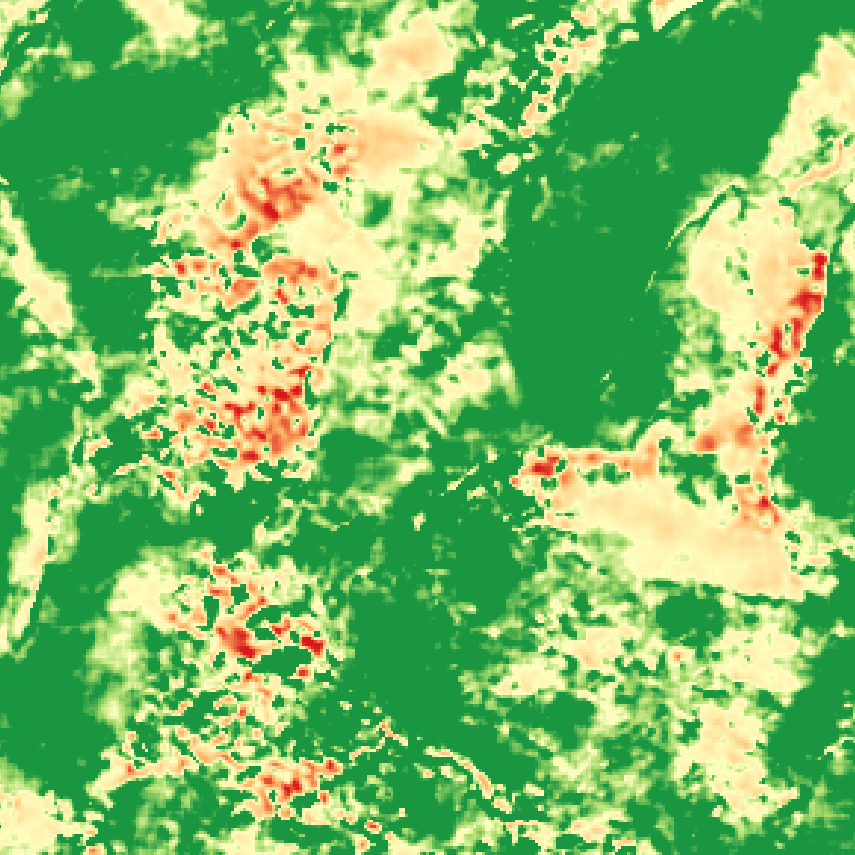}\caption{Ours 5~m}\end{subfigure} &
\begin{subfigure}{.067\textwidth}\includegraphics[width=\linewidth]{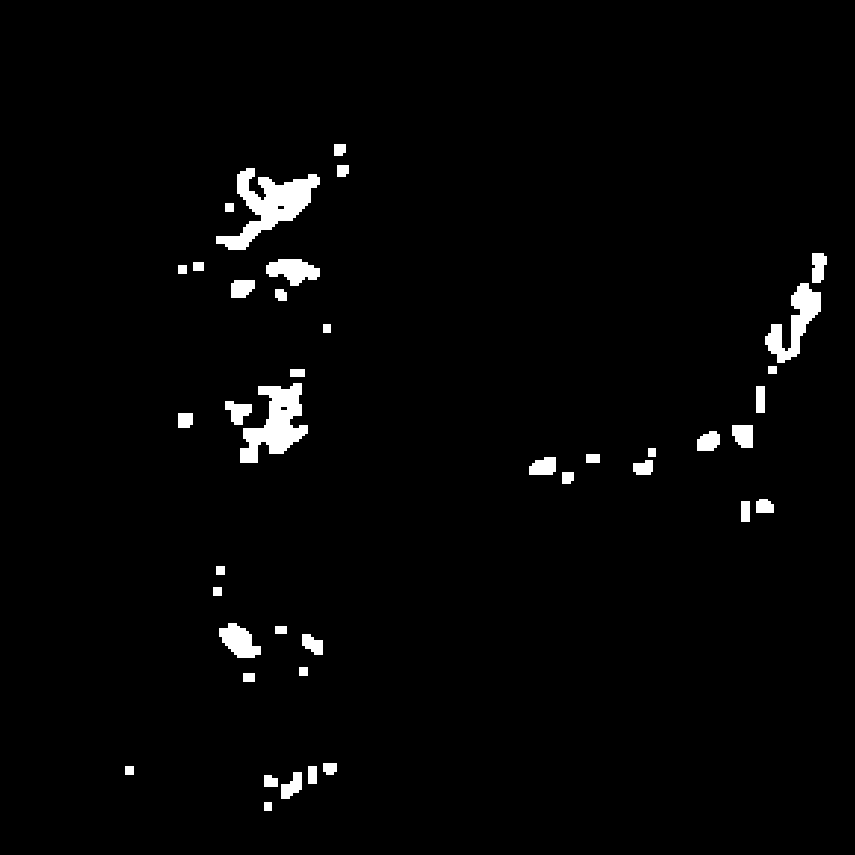}\caption{Ours 5~m}\end{subfigure} &
\begin{subfigure}{.067\textwidth}\includegraphics[width=\linewidth]{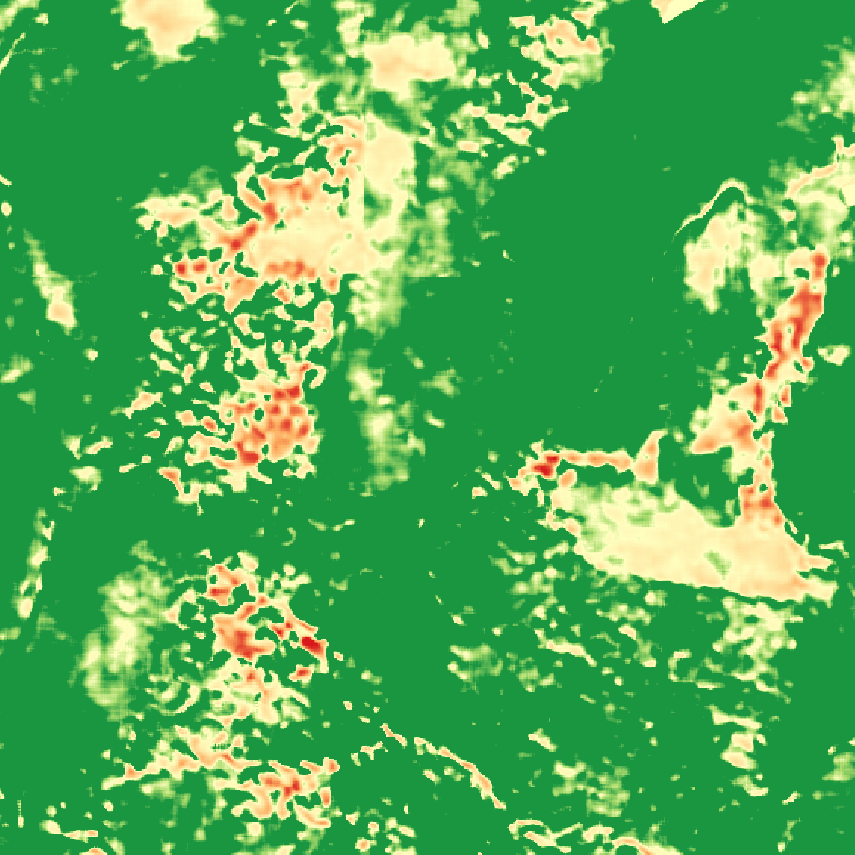}\caption{Ours 2.5~m}\end{subfigure} &
\begin{subfigure}{.067\textwidth}\includegraphics[width=\linewidth]{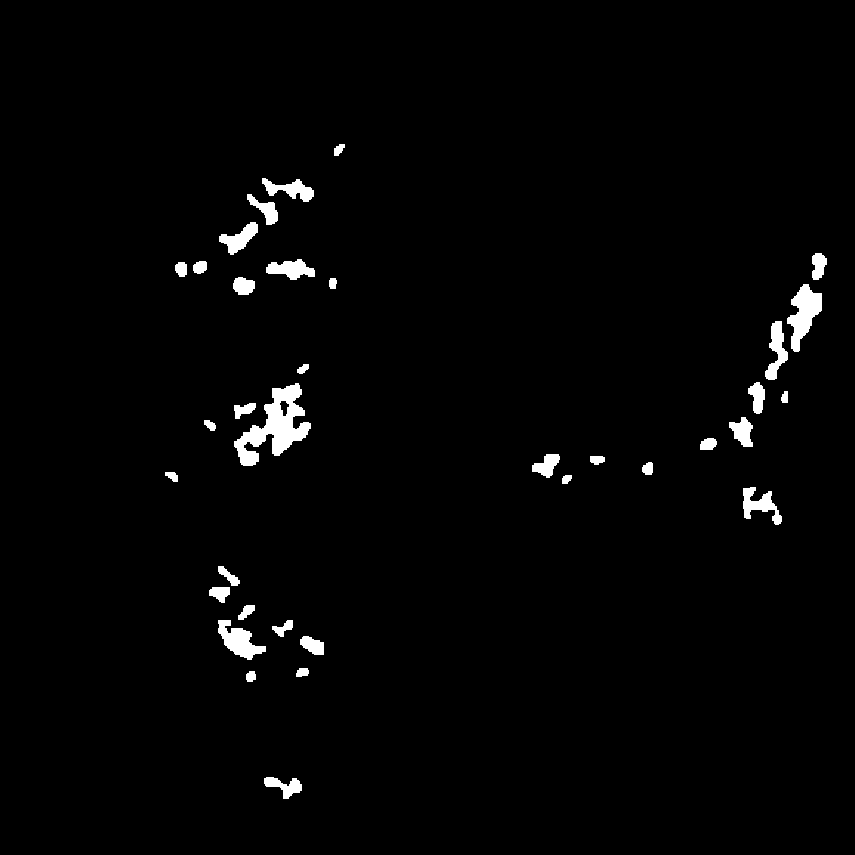}\caption{Ours 2.5~m}\end{subfigure} &
\begin{subfigure}{.067\textwidth}\includegraphics[width=\linewidth]{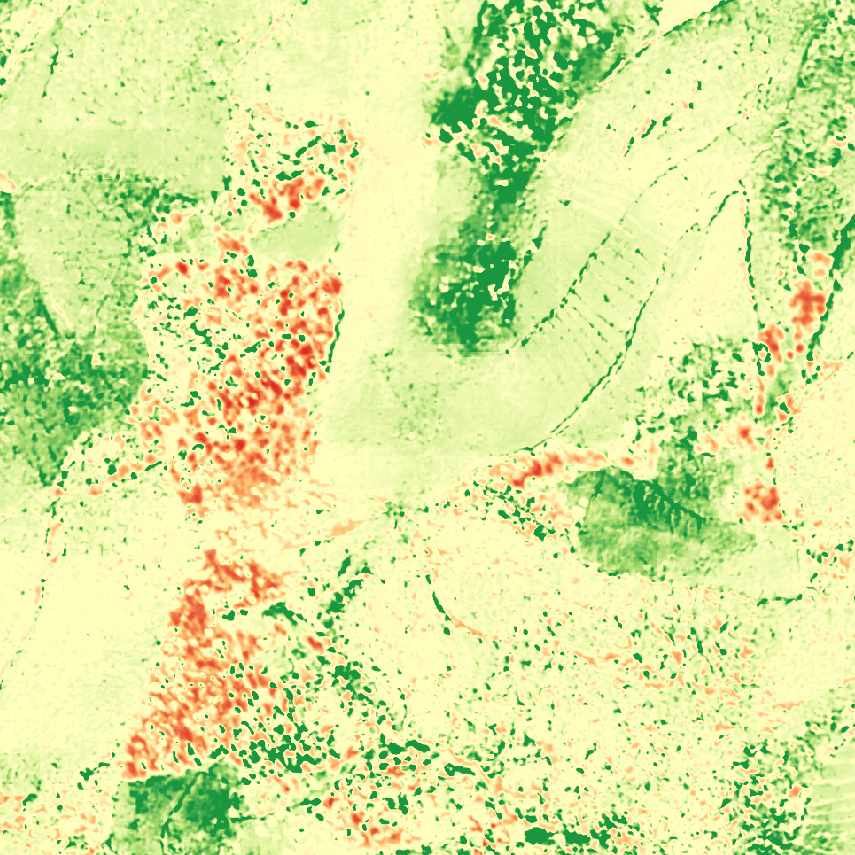}\caption{Open-Canopy}\end{subfigure} &
\begin{subfigure}{.067\textwidth}\includegraphics[width=\linewidth]{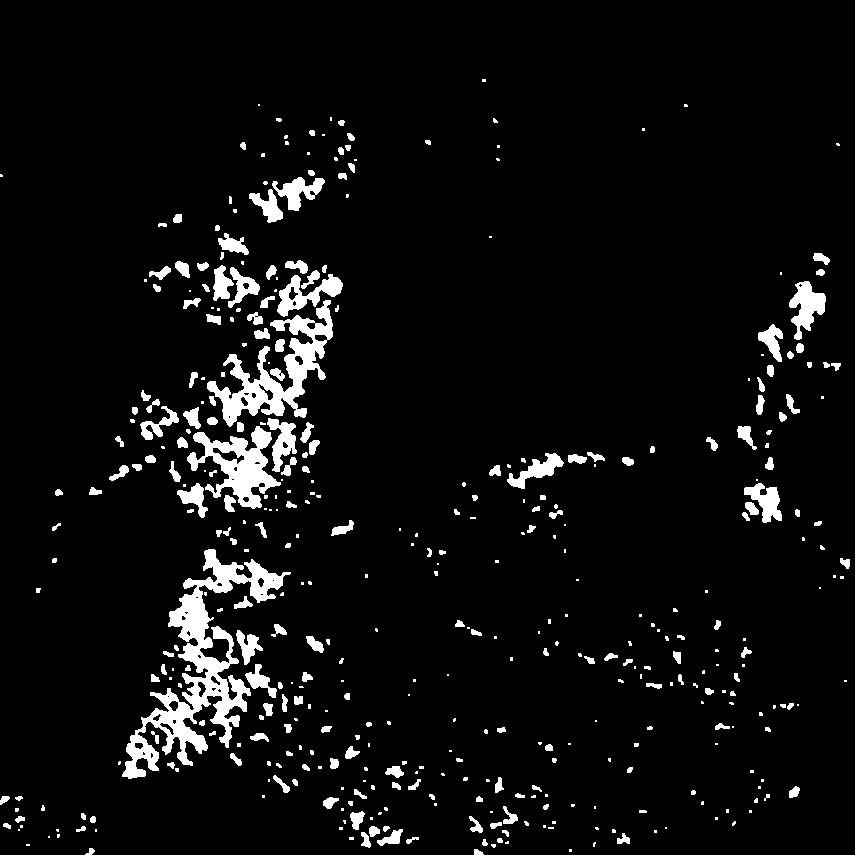}\caption{Open-Canopy}\end{subfigure} &
\begin{subfigure}{.067\textwidth}\includegraphics[width=\linewidth]{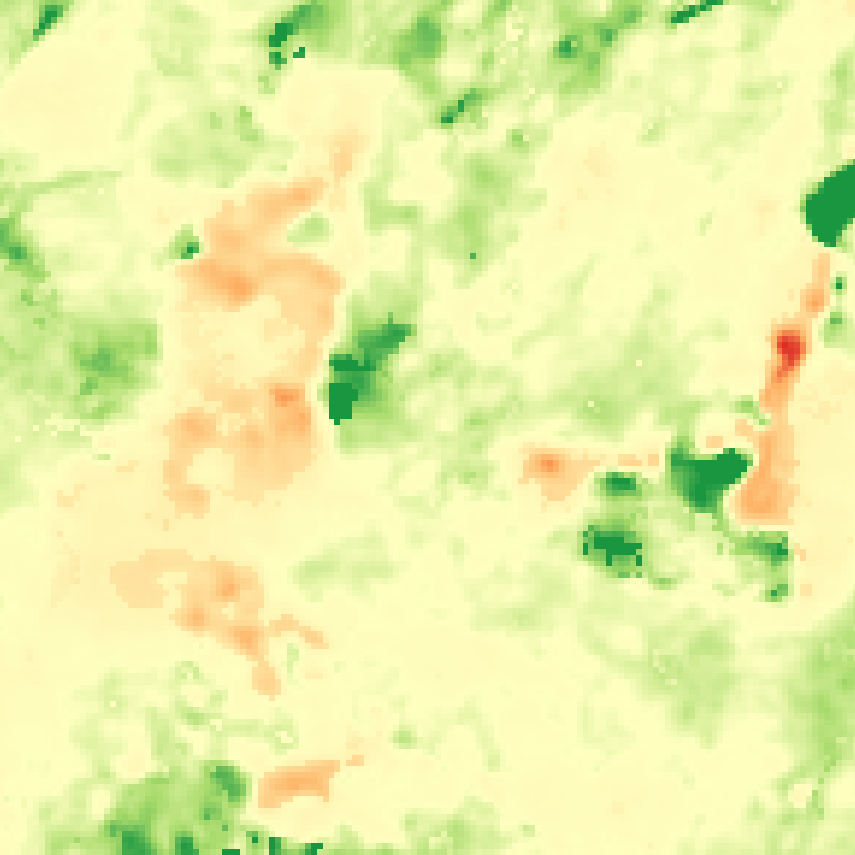}\caption{FORMS-T}\end{subfigure} &
\begin{subfigure}{.067\textwidth}\includegraphics[width=\linewidth]{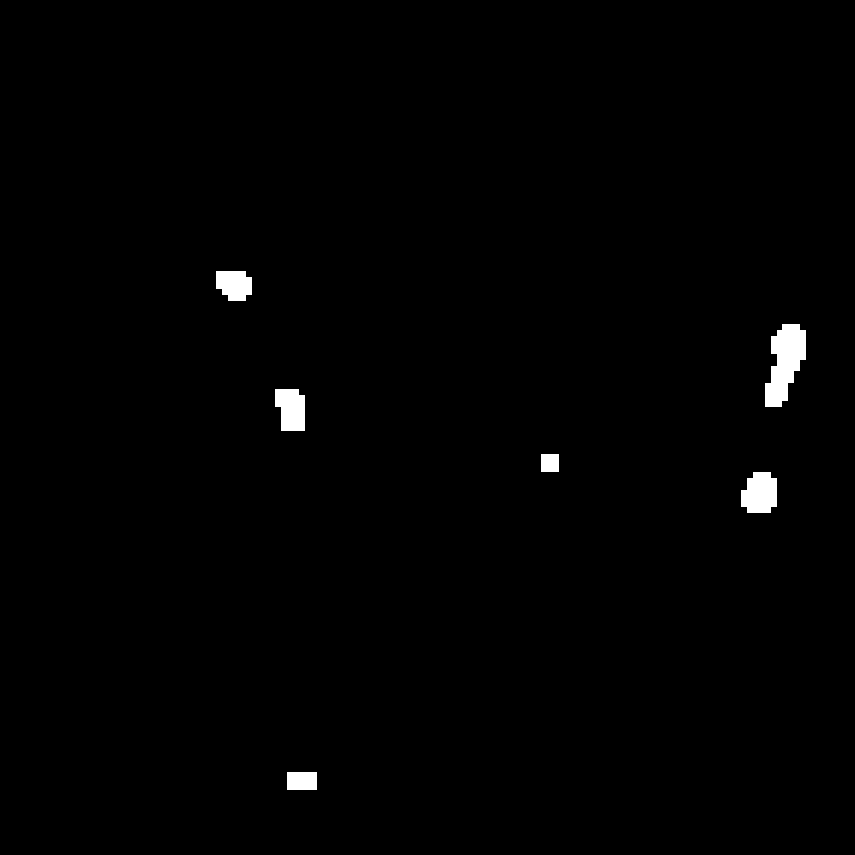}\caption{FORMS-T}\end{subfigure} &
\begin{subfigure}{.04\textwidth}\includegraphics[width=\linewidth]{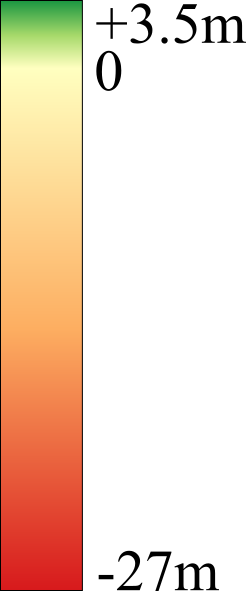}\end{subfigure} \\
\end{tabular*}
\end{adjustbox}

\caption{Change detection results across the four sites. For each row, the first two panels show SPOT-6/7 images for the change year pairs (a)-(b). Reference ALS-derived maps are shown as height difference maps (HDM) (c) and binary change maps (CM) (d) at 2.5~m resolution. Predictions from THREASURE-Net at three different resolutions are shown in pairs: HDM and CM for each resolution (e)-(f), (g)-(h), (i)-(j) for 10, 5 and 2.5~m respectively . Open-Canopy~\citep{11095116} is shown as HDM and CM (k)-(l), and FORMS-T~\citep{SCHWARTZ2025114959} as HDM and CM (m)-(n). The final panel in each row shows the corresponding legend.}
\label{fig:change_detection}


\end{figure}
\end{landscape}

The quantitative results are presented in Table~\ref{tab:change_metrics} and the qualitative results are presented in Figure~\ref{fig:change_detection}. 
For each site, we report \textbf{intersection over union (IoU)}, \textbf{precision}, \textbf{recall}, and \textbf{F1-score}. Let \(TP\) denote true positives, \(FP\) false positives, and \(FN\) false negatives. Then the metrics are defined as:

\[
\text{IoU} = \frac{TP}{TP + FP + FN}
\]

\[
\text{Precision} = \frac{TP}{TP + FP}
\]

\[
\text{Recall} = \frac{TP}{TP + FN}
\]

\[
\text{F1-score} = \frac{2 \cdot TP}{2 \cdot TP + FP + FN}
\]

As in the previous experiments, the quantitative performance of THREASURE-Net generally decreases as the spatial resolution becomes finer, although the visual results show an increased level of detail. This behavior can be explained by the fact that the reference ALS-derived height difference maps are resampled to match the resolution of the predictions. As a result, at coarser resolutions, the reference data contain fewer fine-scale details, which can artificially improve the quantitative metrics.

 \rev{Overall, THREASURE-Net is not systematically better or worse than the compared methods. It achieves competitive or state-of-the-art performance in terms of tree loss detection, particularly in Lajoux-Fresse and Carcans-Hourtins, where F1 scores reach 74.7\% and 83.1\%, respectively. In particular, all algorithms exhibit relatively poor performance at the Mouterhouse site. This may be related to the local topography, which consists of numerous small steep hills that can introduce additional errors in the height estimates. Nevertheless, at 10~m resolution, our method achieves performance comparable to that of FORMS-T.} In contrast, despite using VHR input data, Open-Canopy  did not show the best performance. The visual results indicate that this method tends to underestimate the magnitude of height changes.

\begin{table*}[htbp]
{\color{blue}
\centering
\caption{Tree loss detection performance for for different sites. We report the performance of THREASURE-Net at three different spatial resolutions, as well as Open-Canopy~\citep{11095116} and FORMS-T~\citep{SCHWARTZ2025114959}.}
\label{tab:change_metrics}
\begin{tabular}{l l c c c c c}
\hline
\textbf{Site (Years)} & \textbf{Model} & \textbf{Resolution} & \textbf{IoU (\%)} & \textbf{Recall (\%)} & \textbf{Precision (\%)} & \textbf{F1 (\%)} \\
\hline

\multirow{5}{*}{Carcans-Hourtins (2023-2020)}
 & \multirow{3}{*}{THREASURE-Net}   & 10 m  & 69.6          & 74.7          & 91.1          & 82.1 \\
 &                                  & 5 m   & \textbf{71.1} & 76.0          & 91.8          & \textbf{83.1} \\
 &                                  & 2.5 m & 65.1          & \textbf{75.7} & 82.3          & 78.9 \\
 \cline{2-7}
 & Open-Canopy                      & 1.5 m & 30.2          & 30.8          & \textbf{93.7} & 46.4 \\
 & FORMS-T                          & 10 m  & 68.7          & \textbf{77.1} & 86.3          & 81.5 \\
\hline

\multirow{5}{*}{Déodatie (2022-2018)}
 & \multirow{3}{*}{THREASURE-Net}   & 10 m  & 54.2          & 78.2          & 63.8          & 70.3 \\
 &                                  & 5 m   & 57.3          & 75.9          & 70.0          & 72.9 \\
 &                                  & 2.5 m & 52.7          & 69.7          & 68.3          & 69.0 \\
\cline{2-7}
 & Open-Canopy                      & 1.5 m & \textbf{67.6} & 79.7          & \textbf{81.6} & \textbf{80.6} \\
 & FORMS-T                          & 10 m  & 56.5          & \textbf{85.6} & 62.4          & 72.2 \\
\hline

\multirow{5}{*}{Lajoux-Fresse (2022-2019)}
 & \multirow{3}{*}{THREASURE-Net}   & 10 m  & \textbf{59.6} & 83.5          & 67.5 & \textbf{74.7} \\
 &                                  & 5 m   & 47.6          & 74.3          & 56.9          & 64.5 \\
 &                                  & 2.5 m & 43.8          & 73.0          & 52.3          & 61.0 \\
 \cline{2-7}
 & Open-Canopy                      & 1.5 m & 51.4          & 67.9          & \textbf{67.9}          & 67.9 \\
 & FORMS-T                          & 10 m  & 53.5          & \textbf{84.8} & 59.2          & 69.7 \\
\hline

\multirow{5}{*}{Mouterhouse (2022-2019)}
 & \multirow{3}{*}{THREASURE-Net}   & 10 m  & 31.7          & 45.3          & 51.2          & 48.1 \\
 &                                  & 5 m   & 23.5          & 31.9          & 47.0          & 38.0 \\
 &                                  & 2.5 m & 18.9          & 22.6          & \textbf{54.0} & 31.8 \\
 \cline{2-7}
 & Open-Canopy                      & 1.5 m & 20.7          & 26.5          & 48.3          & 34.2 \\
 & FORMS-T                          & 10 m  & \textbf{33.7} & \textbf{48.1} & 52.9 & \textbf{50.4} \\
\hline

\end{tabular}
}
\end{table*}

\bibliographystyle{elsarticle-harv} 
\bibliography{biblio}

@article{DBLP:journals/corr/abs-1802-08797,
  author       = {Yulun Zhang and
                  Yapeng Tian and
                  Yu Kong and
                  Bineng Zhong and
                  Yun Fu},
  title        = {Residual Dense Network for Image Super-Resolution},
  journal      = {CoRR},
  volume       = {abs/1802.08797},
  year         = {2018},
  url          = {http://arxiv.org/abs/1802.08797},
  eprinttype    = {arXiv},
  eprint       = {1802.08797},
  bibsource    = {dblp computer science bibliography, https://dblp.org}
}

@article{bouvet2018above,
  title={An above-ground biomass map of African savannahs and woodlands at 25 m resolution derived from ALOS PALSAR},
  author={Bouvet, Alexandre and Mermoz, St{\'e}phane and Le Toan, Thuy and Villard, Ludovic and Mathieu, Renaud and Naidoo, Laven and Asner, Gregory P},
  journal={Remote sensing of environment},
  volume={206},
  pages={156--173},
  year={2018},
  publisher={Elsevier}
}

@article{mermoz2024submonthly,
  title={Submonthly assessment of temperate forest Clear-Cuts in Mainland France},
  author={Mermoz, St{\'e}phane and Prieto, Juan Doblas and Planells, Milena and Morin, David and Koleck, Thierry and Mouret, Florian and Bouvet, Alexandre and Le Toan, Thuy and Sheeren, David and Hamrouni, Yousra and others},
  journal={IEEE Journal of Selected Topics in Applied Earth Observations and Remote Sensing},
  volume={17},
  pages={13743--13764},
  year={2024},
  publisher={IEEE}
}

@article{avitabile2016integrated,
  title={An integrated pan-tropical biomass map using multiple reference datasets},
  author={Avitabile, Valerio and Herold, Martin and Heuvelink, Gerard BM and Lewis, Simon L and Phillips, Oliver L and Asner, Gregory P and Armston, John and Ashton, Peter S and Banin, Lindsay and Bayol, Nicolas and others},
  journal={Global change biology},
  volume={22},
  number={4},
  pages={1406--1420},
  year={2016},
  publisher={Wiley Online Library}
}

@article{simard2011mapping,
  title={Mapping forest canopy height globally with spaceborne lidar},
  author={Simard, Marc and Pinto, Naiara and Fisher, Joshua B and Baccini, Alessandro},
  journal={Journal of Geophysical Research: Biogeosciences},
  volume={116},
  number={G4},
  year={2011},
  publisher={Wiley Online Library}
}

@article{morin2023estimation,
  title={Estimation of forest height and biomass from open-access multi-sensor satellite imagery and GEDI Lidar data: high-resolution maps of metropolitan France},
  author={Morin, David and Planells, Milena and Mermoz, St{\'e}phane and Mouret, Florian},
  journal={arXiv preprint arXiv:2310.14662},
  year={2023}
}

@article{DBLP:journals/corr/HeZRS15,
  author       = {Kaiming He and
                  Xiangyu Zhang and
                  Shaoqing Ren and
                  Jian Sun},
  title        = {Deep Residual Learning for Image Recognition},
  journal      = {CoRR},
  volume       = {abs/1512.03385},
  year         = {2015},
  url          = {http://arxiv.org/abs/1512.03385},
  eprinttype    = {arXiv},
  eprint       = {1512.03385},
  bibsource    = {dblp computer science bibliography, https://dblp.org}
}

@article{DBLP:journals/corr/abs-1809-00219,
  author       = {Xintao Wang and
                  Ke Yu and
                  Shixiang Wu and
                  Jinjin Gu and
                  Yihao Liu and
                  Chao Dong and
                  Chen Change Loy and
                  Yu Qiao and
                  Xiaoou Tang},
  title        = {{ESRGAN:} Enhanced Super-Resolution Generative Adversarial Networks},
  journal      = {CoRR},
  volume       = {abs/1809.00219},
  year         = {2018},
  url          = {http://arxiv.org/abs/1809.00219},
  eprinttype    = {arXiv},
  eprint       = {1809.00219},
  bibsource    = {dblp computer science bibliography, https://dblp.org}
}

@article{8310638,
  author={Nie, Dong and Trullo, Roger and Lian, Jun and Wang, Li and Petitjean, Caroline and Ruan, Su and Wang, Qian and Shen, Dinggang},
  journal={IEEE Transactions on Biomedical Engineering}, 
  title={Medical Image Synthesis with Deep Convolutional Adversarial Networks}, 
  year={2018},
  volume={65},
  number={12},
  pages={2720-2730},
  doi={10.1109/TBME.2018.2814538}}

@INPROCEEDINGS{shuffle,
  author={Shi, Wenzhe and Caballero, Jose and Huszár, Ferenc and Totz, Johannes and Aitken, Andrew P. and Bishop, Rob and Rueckert, Daniel and Wang, Zehan},
  booktitle={2016 IEEE Conference on Computer Vision and Pattern Recognition (CVPR)}, 
  title={Real-Time Single Image and Video Super-Resolution Using an Efficient Sub-Pixel Convolutional Neural Network}, 
  year={2016},
  volume={},
  number={},
  pages={1874-1883},
  doi={10.1109/CVPR.2016.207}}

@misc{aitken2017checkerboardartifactfreesubpixel,
      title={Checkerboard artifact free sub-pixel convolution: A note on sub-pixel convolution, resize convolution and convolution resize}, 
      author={Andrew Aitken and Christian Ledig and Lucas Theis and Jose Caballero and Zehan Wang and Wenzhe Shi},
      year={2017},
      eprint={1707.02937},
      archivePrefix={arXiv},
      primaryClass={cs.CV},
      url={https://arxiv.org/abs/1707.02937}, 
}

@article{CAO2024114241,
title = {A deep learning-based super-resolution method for building height estimation at 2.5 m spatial resolution in the Northern Hemisphere},
journal = {Remote Sensing of Environment},
volume = {310},
pages = {114241},
year = {2024},
issn = {0034-4257},
doi = {https://doi.org/10.1016/j.rse.2024.114241},
url = {https://www.sciencedirect.com/science/article/pii/S0034425724002591},
author = {Yinxia Cao and Qihao Weng},
}

@article{Lei2019SimultaneousSA,
  title={Simultaneous Super-Resolution and Segmentation for Remote Sensing Images},
  author={Sen Lei and Zhenwei Shi and Xi Wu and Bin Pan and Xia Xu and Hongxun Hao},
  journal={IGARSS 2019 - 2019 IEEE International Geoscience and Remote Sensing Symposium},
  year={2019},
  pages={3121-3124},
  url={https://api.semanticscholar.org/CorpusID:208037945}
}

@Article{rs13224547,
AUTHOR = {Abadal, Saüc and Salgueiro, Luis and Marcello, Javier and Vilaplana, Verónica},
TITLE = {A Dual Network for Super-Resolution and Semantic Segmentation of Sentinel-2 Imagery},
JOURNAL = {Remote Sensing},
VOLUME = {13},
YEAR = {2021},
NUMBER = {22},
ARTICLE-NUMBER = {4547},
URL = {https://www.mdpi.com/2072-4292/13/22/4547},
ISSN = {2072-4292},
DOI = {10.3390/rs13224547}
}

@article{DBLP:journals/corr/abs-2007-00586,
  author       = {Vivien Sainte Fare Garnot and
                  Lo{\"{\i}}c Landrieu},
  title        = {Lightweight Temporal Self-Attention for Classifying Satellite Image
                  Time Series},
  journal      = {CoRR},
  volume       = {abs/2007.00586},
  year         = {2020},
  url          = {https://arxiv.org/abs/2007.00586},
  eprinttype    = {arXiv},
  eprint       = {2007.00586},
  bibsource    = {dblp computer science bibliography, https://dblp.org}
}

@article{DBLP:journals/corr/abs-1811-10166,
  author       = {Charlotte Pelletier and
                  Geoffrey I. Webb and
                  Fran{\c{c}}ois Petitjean},
  title        = {Temporal Convolutional Neural Network for the Classification of Satellite
                  Image Time Series},
  journal      = {CoRR},
  volume       = {abs/1811.10166},
  year         = {2018},
  url          = {http://arxiv.org/abs/1811.10166},
  eprinttype    = {arXiv},
  eprint       = {1811.10166},
  bibsource    = {dblp computer science bibliography, https://dblp.org}
}

@article{COOPS2021112477,
title = {Modelling lidar-derived estimates of forest attributes over space and time: A review of approaches and future trends},
journal = {Remote Sensing of Environment},
volume = {260},
pages = {112477},
year = {2021},
issn = {0034-4257},
doi = {https://doi.org/10.1016/j.rse.2021.112477},
url = {https://www.sciencedirect.com/science/article/pii/S0034425721001954},
author = {Nicholas C. Coops and Piotr Tompalski and Tristan R.H. Goodbody and Martin Queinnec and Joan E. Luther and Douglas K. Bolton and Joanne C. White and Michael A. Wulder and Oliver R. {van Lier} and Txomin Hermosilla}
}

@article{SCHWARTZ2025114959,
title = {Retrieving yearly forest growth from satellite data: A deep learning based approach},
journal = {Remote Sensing of Environment},
volume = {330},
pages = {114959},
year = {2025},
issn = {0034-4257},
doi = {https://doi.org/10.1016/j.rse.2025.114959},
url = {https://www.sciencedirect.com/science/article/pii/S0034425725003633},
author = {Martin Schwartz and Philippe Ciais and Ewan Sean and Aurélien {de Truchis} and Cédric Vega and Nikola Besic and Ibrahim Fayad and Jean-Pierre Wigneron and Sarah Brood and Agnès Pelissier-Tanon and Jan Pauls and Gabriel Belouze and Yidi Xu}
}

@INPROCEEDINGS{11095116,
  author={Fogel, Fajwel and Perron, Yohann and Besic, Nikola and Saint-André, Laurent and Pellissier-Tanon, Agnès and Schwartz, Martin and Boudras, Thomas and Fayad, Ibrahim and d’Aspremont, Alexandre and Landrieu, Loic and Ciais, Philippe},
  booktitle={2025 IEEE/CVF Conference on Computer Vision and Pattern Recognition (CVPR)}, 
  title={Open-Canopy: Towards Very High Resolution Forest Monitoring}, 
  year={2025},
  volume={},
  number={},
  pages={1395-1406},
  doi={10.1109/CVPR52734.2025.00138}}

@inproceedings{warmrestart,
author = {Loshchilov, Ilya and Hutter, Frank},
year = {2016},
month = {08},
title = {SGDR: Stochastic Gradient Descent with Warm Restarts},
doi = {10.48550/arXiv.1608.03983}
}

@article{lang2023high,
  title={A high-resolution canopy height model of the Earth},
  author={Lang, Nico and Jetz, Walter and Schindler, Konrad and Wegner, Jan Dirk},
  journal={Nature Ecology \& Evolution},
  volume={7},
  number={11},
  pages={1778--1789},
  year={2023},
  publisher={Nature Publishing Group UK London}
}

@article{Pinus,
author = {Chevalier, Romain and Catapano, Anita and Pommier, Regis and Montemurro, Marco},
year = {2024},
month = {04},
pages = {},
title = {A review on properties and variability of Pinus Pinaster Ait. ssp. Atlantica existing in the Landes of Gascogne},
volume = {70},
journal = {Journal of Wood Science},
doi = {10.1186/s10086-024-02127-3}
}

@article{DUBAYAH2020100002,
title = {The Global Ecosystem Dynamics Investigation: High-resolution laser ranging of the Earth’s forests and topography},
journal = {Science of Remote Sensing},
volume = {1},
pages = {100002},
year = {2020},
issn = {2666-0172},
doi = {https://doi.org/10.1016/j.srs.2020.100002},
url = {https://www.sciencedirect.com/science/article/pii/S2666017220300018},
author = {Ralph Dubayah and James Bryan Blair and Scott Goetz and Lola Fatoyinbo and Matthew Hansen and Sean Healey and Michelle Hofton and George Hurtt and James Kellner and Scott Luthcke and John Armston and Hao Tang and Laura Duncanson and Steven Hancock and Patrick Jantz and Suzanne Marselis and Paul L. Patterson and Wenlu Qi and Carlos Silva}
}

@inproceedings{BreizhSR,
author = {Okabayashi, Aimi and Audebert, Nicolas and Donike, Simon and Pelletier, Charlotte},
year = {2024},
month = {06},
pages = {502-511},
title = {Cross-sensor super-resolution of irregularly sampled Sentinel-2 time series},
doi = {10.1109/CVPRW63382.2024.00055}
}

@article{DBLP:journals/corr/abs-2002-06460,
  author       = {Michel Deudon and
                  Alfredo Kalaitzis and
                  Israel Goytom and
                  Md Rifat Arefin and
                  Zhichao Lin and
                  Kris Sankaran and
                  Vincent Michalski and
                  Samira Ebrahimi Kahou and
                  Julien Cornebise and
                  Yoshua Bengio},
  title        = {HighRes-net: Recursive Fusion for Multi-Frame Super-Resolution of
                  Satellite Imagery},
  journal      = {CoRR},
  volume       = {abs/2002.06460},
  year         = {2020},
  url          = {https://arxiv.org/abs/2002.06460},
  eprinttype    = {arXiv},
  eprint       = {2002.06460},
  bibsource    = {dblp computer science bibliography, https://dblp.org}
}

@article{DBLP:journals/corr/abs-2406-10225,
  author       = {Zhaoxu Luo and
                  Bowen Song and
                  Liyue Shen},
  title        = {SatDiffMoE: {A} Mixture of Estimation Method for Satellite Image Super-resolution
                  with Latent Diffusion Models},
  journal      = {CoRR},
  volume       = {abs/2406.10225},
  year         = {2024},
  url          = {https://doi.org/10.48550/arXiv.2406.10225},
  doi          = {10.48550/ARXIV.2406.10225},
  eprinttype    = {arXiv},
  eprint       = {2406.10225},
  bibsource    = {dblp computer science bibliography, https://dblp.org}
}

@ARTICLE{11010858,
  author={Michel, Julien and Kalinicheva, Ekaterina and Inglada, Jordi},
  journal={IEEE Transactions on Geoscience and Remote Sensing}, 
  title={Revisiting Remote Sensing Cross-Sensor Single Image Super-Resolution: The Overlooked Impact of Geometric and Radiometric Distortion}, 
  year={2025},
  volume={63},
  number={},
  pages={1-22},
  doi={10.1109/TGRS.2025.3572548}}

@article{cresson2022sr4rs,
  title={SR4RS: A Tool for Super Resolution of Remote Sensing Images},
  author={Cresson, R{\'e}mi},
  journal={Journal of Open Research Software},
  volume={10},
  number={1},
  year={2022},
  publisher={Ubiquity Press}
}

@book{FAO2025,
  title        = {Global Forest Resources Assessment 2025},
  author       = {{FAO}},
  year         = {2025},
  publisher    = {Food and Agriculture Organization of the United Nations},
  address      = {Rome},
  url          = {https://openknowledge.fao.org/server/api/core/bitstreams/12322cae-5b20-4be2-927a-72a86fd319e9/content}
}

@article{AHMED201589,
title = {Characterizing stand-level forest canopy cover and height using Landsat time series, samples of airborne LiDAR, and the Random Forest algorithm},
journal = {ISPRS Journal of Photogrammetry and Remote Sensing},
volume = {101},
pages = {89-101},
year = {2015},
issn = {0924-2716},
doi = {https://doi.org/10.1016/j.isprsjprs.2014.11.007},
url = {https://www.sciencedirect.com/science/article/pii/S0924271614002755},
author = {Oumer S. Ahmed and Steven E. Franklin and Michael A. Wulder and Joanne C. White}
}

@article{WANG201624,
title = {A combined GLAS and MODIS estimation of the global distribution of mean forest canopy height},
journal = {Remote Sensing of Environment},
volume = {174},
pages = {24-43},
year = {2016},
issn = {0034-4257},
doi = {https://doi.org/10.1016/j.rse.2015.12.005},
url = {https://www.sciencedirect.com/science/article/pii/S0034425715302261},
author = {Yuanyuan Wang and Guicai Li and Jianhua Ding and Zhaodi Guo and Shihao Tang and Cheng Wang and Qingni Huang and Ronggao Liu and Jing M. Chen}
}

@article{rs_forest,
author = {Besic, Nikola and Picard, Nicolas and Vega, Cedric and Bontemps, Jean-Daniel and Hertzog, Lionel and Renaud, Jean-Pierre and Fogel, Fajwel and Schwartz, Martin and Pellissier-Tanon, Agnès and Destouet, Gabriel and Mortier, Frédéric and Planells-Rodriguez, Milena and Ciais, Philippe},
year = {2025},
month = {01},
pages = {337-359},
title = {Remote-sensing-based forest canopy height mapping: some models are useful, but might they provide us with even more insights when combined?},
volume = {18},
journal = {Geoscientific Model Development},
doi = {10.5194/gmd-18-337-2025}
}

@ARTICLE{9044873,
  author={Wang, Zhihao and Chen, Jian and Hoi, Steven C. H.},
  journal={IEEE Transactions on Pattern Analysis and Machine Intelligence}, 
  title={Deep Learning for Image Super-Resolution: A Survey}, 
  year={2021},
  volume={43},
  number={10},
  pages={3365-3387},
  doi={10.1109/TPAMI.2020.2982166}}

@Article{rs14215423,
AUTHOR = {Wang, Xuan and Yi, Jinglei and Guo, Jian and Song, Yongchao and Lyu, Jun and Xu, Jindong and Yan, Weiqing and Zhao, Jindong and Cai, Qing and Min, Haigen},
TITLE = {A Review of Image Super-Resolution Approaches Based on Deep Learning and Applications in Remote Sensing},
JOURNAL = {Remote Sensing},
VOLUME = {14},
YEAR = {2022},
NUMBER = {21},
ARTICLE-NUMBER = {5423},
URL = {https://www.mdpi.com/2072-4292/14/21/5423},
ISSN = {2072-4292},
DOI = {10.3390/rs14215423}
}

@article{rs_esrgan,
author = {Salgueiro Romero, Luis and Marcello, J. and Vilaplana, Verónica},
year = {2022},
month = {11},
pages = {5862},
title = {SEG-ESRGAN: A Multi-Task Network for Super-Resolution and Semantic Segmentation of Remote Sensing Images},
volume = {14},
journal = {Remote Sensing},
doi = {10.3390/rs14225862}
}

@article{Liu2024,
author = {Liu, Denghui and Zhong, Lin and Wu, Haiyang and Li, Songyang and Li, Yida},
year = {2025},
month = {01},
pages = {},
title = {Remote sensing image Super-resolution reconstruction by fusing multi-scale receptive fields and hybrid transformer},
volume = {15},
journal = {Scientific Reports},
doi = {10.1038/s41598-025-86446-5}
}

@article{Muller2020,
author = {Müller, Markus and Ekhtiari, Nikoo and Almeida, R. and Rieke, Christoph},
year = {2020},
month = {08},
pages = {33-40},
title = {SUPER-RESOLUTION OF MULTISPECTRAL SATELLITE IMAGES USING CONVOLUTIONAL NEURAL NETWORKS},
volume = {V-1-2020},
journal = {ISPRS Annals of Photogrammetry, Remote Sensing and Spatial Information Sciences},
doi = {10.5194/isprs-annals-V-1-2020-33-2020}
}

@Article{essd-15-4927-2023,
AUTHOR = {Schwartz, M. and Ciais, P. and De Truchis, A. and Chave, J. and Ottl\'e, C. and Vega, C. and Wigneron, J.-P. and Nicolas, M. and Jouaber, S. and Liu, S. and Brandt, M. and Fayad, I.},
TITLE = {FORMS: Forest Multiple Source height, wood volume, and biomass maps in
France at 10 to 30\,m resolution based on Sentinel-1, Sentinel-2, and Global Ecosystem Dynamics Investigation (GEDI) data with a deep learning approach},
JOURNAL = {Earth System Science Data},
VOLUME = {15},
YEAR = {2023},
NUMBER = {11},
PAGES = {4927--4945},
URL = {https://essd.copernicus.org/articles/15/4927/2023/},
DOI = {10.5194/essd-15-4927-2023}
}

@article{BECKER2023269,
title = {Country-wide retrieval of forest structure from optical and SAR satellite imagery with deep ensembles},
journal = {ISPRS Journal of Photogrammetry and Remote Sensing},
volume = {195},
pages = {269-286},
year = {2023},
issn = {0924-2716},
doi = {https://doi.org/10.1016/j.isprsjprs.2022.11.011},
url = {https://www.sciencedirect.com/science/article/pii/S0924271622003045},
author = {Alexander Becker and Stefania Russo and Stefano Puliti and Nico Lang and Konrad Schindler and Jan Dirk Wegner}
}

@article{
doi10.1126_sciadv.adh4097,
author = {Siyu Liu  and Martin Brandt  and Thomas Nord-Larsen  and Jerome Chave  and Florian Reiner  and Nico Lang  and Xiaoye Tong  and Philippe Ciais  and Christian Igel  and Adrian Pascual  and Juan Guerra-Hernandez  and Sizhuo Li  and Maurice Mugabowindekwe  and Sassan Saatchi  and Yuemin Yue  and Zhengchao Chen  and Rasmus Fensholt },
title = {The overlooked contribution of trees outside forests to tree cover and woody biomass across Europe},
journal = {Science Advances},
volume = {9},
number = {37},
pages = {eadh4097},
year = {2023},
doi = {10.1126/sciadv.adh4097},
URL = {https://www.science.org/doi/abs/10.1126/sciadv.adh4097},
eprint = {https://www.science.org/doi/pdf/10.1126/sciadv.adh4097}}

@article{TOLAN2024113888,
title = {Very high resolution canopy height maps from RGB imagery using self-supervised vision transformer and convolutional decoder trained on aerial lidar},
journal = {Remote Sensing of Environment},
volume = {300},
pages = {113888},
year = {2024},
issn = {0034-4257},
doi = {https://doi.org/10.1016/j.rse.2023.113888},
url = {https://www.sciencedirect.com/science/article/pii/S003442572300439X},
author = {Jamie Tolan and Hung-I Yang and Benjamin Nosarzewski and Guillaume Couairon and Huy V. Vo and John Brandt and Justine Spore and Sayantan Majumdar and Daniel Haziza and Janaki Vamaraju and Theo Moutakanni and Piotr Bojanowski and Tracy Johns and Brian White and Tobias Tiecke and Camille Couprie}
}

@report{Kakoulaki2021LiDAR,
  author       = {Kakoulaki, Georgia and Martinez, Alberto and Florio, Pietro},
  title        = {Non-commercial Light Detection and Ranging (LiDAR) data in Europe},
  institution  = {Publications Office of the European Union},
  year         = {2021},
  address      = {Luxembourg},
  number       = {EUR 30817 EN},
  doi          = {10.2760/212427},
  isbn         = {978-92-76-41150-5},
  note         = {JRC126223}
}

@article{otsu1979threshold,
  title={A threshold selection method from gray-level histograms},
  author={Otsu, Nobuyuki},
  journal={IEEE Transactions on Systems, Man, and Cybernetics},
  volume={9},
  number={1},
  pages={62--66},
  year={1979},
  publisher={IEEE}
}

@InProceedings{otsu_change,
author="Kalinicheva, Ekaterina
and Sublime, J{\'e}r{\'e}mie
and Trocan, Maria",
editor="Tetko, Igor V.
and K{\r{u}}rkov{\'a}, V{\v{e}}ra
and Karpov, Pavel
and Theis, Fabian",
title="Change Detection in Satellite Images Using Reconstruction Errors of Joint Autoencoders",
booktitle="Artificial Neural Networks and Machine Learning -- ICANN 2019: Image Processing",
year="2019",
publisher="Springer International Publishing",
address="Cham",
pages="637--648",
isbn="978-3-030-30508-6"
}

@article{mann1945mann,
  title={Non-parametric test against trend},
  author={Mann, Henry B.},
  journal={Econometrica},
  volume={13},
  number={3},
  pages={245--259},
  year={1945},
  publisher={The Econometric Society},
  url={http://www.jstor.org/stable/1907187}
}

@Article{rs16213992,
AUTHOR = {Zhang, Boya and Gann, Daniel and Wdowinski, Shimon and Lin, Chaohao and Hestir, Erin and Lamb-Wotton, Lukas and Ishtiaq, Khandker S. and Smith, Kaleb and Li, Yuepeng},
TITLE = {Space-Based Mapping of Pre- and Post-Hurricane Mangrove Canopy Heights Using Machine Learning with Multi-Sensor Observations},
JOURNAL = {Remote Sensing},
VOLUME = {16},
YEAR = {2024},
NUMBER = {21},
ARTICLE-NUMBER = {3992},
URL = {https://www.mdpi.com/2072-4292/16/21/3992},
ISSN = {2072-4292},
DOI = {10.3390/rs16213992}
}

@Article{rs17091536,
AUTHOR = {Wang, Chao and Song, Conghe and Schroeder, Todd A. and Woodcock, Curtis E. and Pavelsky, Tamlin M. and Han, Qianqian and Yao, Fangfang},
TITLE = {Interpretable Multi-Sensor Fusion of Optical and SAR Data for GEDI-Based Canopy Height Mapping in Southeastern North Carolina},
JOURNAL = {Remote Sensing},
VOLUME = {17},
YEAR = {2025},
NUMBER = {9},
ARTICLE-NUMBER = {1536},
URL = {https://www.mdpi.com/2072-4292/17/9/1536},
ISSN = {2072-4292},
DOI = {10.3390/rs17091536}
}

@article{CHEN2025104814,
title = {Multimodal deep learning enables forest height mapping from patchy spaceborne LiDAR using SAR and passive optical satellite data},
journal = {International Journal of Applied Earth Observation and Geoinformation},
volume = {143},
pages = {104814},
year = {2025},
issn = {1569-8432},
doi = {https://doi.org/10.1016/j.jag.2025.104814},
url = {https://www.sciencedirect.com/science/article/pii/S1569843225004613},
author = {Man Chen and Wenquan Dong and Hao Yu and Iain H. Woodhouse and Casey M. Ryan and Haoyu Liu and Selena Georgiou and Edward T.A. Mitchard}
}

\section*{List of Figure Captions}

\textbf{Figure 1.} Distribution of train, test and validation patches. Buffers of 1 km are set around test patches to ensure reliable evaluation of model results.

\vspace{0.6em}

\textbf{Figure 2.} THREASURE-Net model architecture.

\vspace{0.6em}

\textbf{Figure 3.} Sub-pixel convolution. Example of factor 2 upsampling. A LR image with C feature channels and H × W spatial dimensions is passed through a 2D convolution layer to obtain $2^2 \cdot C$ feature channels, where each group of $2^2$ consecutive channels forms a sub-pixel group that is reorganized by pixel shuffle operation into an SR image with C channels and 2H × 2W spatial dimensions.

\vspace{0.6em}

\textbf{Figure 4.} Error distribution across canopy-height bins (1.5–5, 5–10, 10–15, 15–20, 20–25, 25–30, >30 m) for our models at different output resolutions in box-and-whisker plots. A boxplot shows the median and interquartile range (IQR), with whiskers extending to values within 1.5×IQR. Gray bars indicate the proportion of samples in each bin relative to the dataset.

\vspace{0.6em}

\textbf{Figure 5.} Prediction–target scatterplots for our 10 m, 5 m, and 2.5 m models for tree pixels. An equal number of points ($\approx$96 million) were randomly sampled from each model. The identity line is shown in white. Predictions are reported without clipping.

\vspace{0.6em}

\textbf{Figure 6.} 640 × 640 m tree height prediction patches centred on 44.893°N and 0.997°W. (a) Sentinel-2 image from 7 July 2023. (b) VHR Bing image. (c–e) Model predictions at 10 m, 5 m, and 2.5 m resolutions. (f–g) Upsampled predictions using bicubic interpolation. (h) Reference LiDAR-derived height map at 2.5 m resolution. Zoomed-in views highlight fine-scale differences.

\vspace{0.6em}

\textbf{Figure 7.} Normalized log-frequency attenuation profile (FAP) for HR reference patch and SR results from THREASURE-Net and bicubic interpolation at 5 m and 2.5 m resolutions for the patch in Figure 6. Spatial frequencies are normalized as $f = f_n / f_N$, where $f_N$ is the maximum representable spatial frequency.

\vspace{0.6em}

\textbf{Figure 8.} FAP profile for the target image and SR image produced by the model at 5 m resolution, including variants trained with MAE loss only and pwMAE only (without wGDL loss).

\vspace{0.6em}

\textbf{Figure 9.} Tree height prediction of 580 ha area centered on 45.127°N and 0.901°W. Comparison of a) independent reference ALS-derived heights and our model at three resolutions – b) 10 m, c) 5 m, and d) 2.5 m – with concurrent approaches: d) FORMST-T (Schwartz et al., 2025), e) Global Canopy Height Model (Lang et al., 2023), and f) Open-Canopy (Fogel et al., 2025). Zoomed-in views (top left) highlight differences in fine details among predictions on the eastern side of the study area.

\vspace{0.6em}

\textbf{Figure B.10.} Sylvo-ecoregions of France, illustrating the ecological diversity of forest types across the country.

\vspace{0.6em}

\textbf{Figure C.11.} Tree height dynamics for Mann–Kendall high trend class at 2.5 m, 5 m, and 10 m resolutions. Median curves with interquartile ranges are shown.

\vspace{0.6em}

\textbf{Figure D.12.} Change detection results across four sites. Each row shows SPOT-6/7 image pairs, ALS-derived reference maps (height difference and binary change maps), predictions from THREASURE-Net at three resolutions, and comparisons with Open-Canopy (Fogel et al., 2025) and FORMS-T (Schwartz et al., 2025). Zoomed views highlight spatial detail differences.

\end{document}